\documentclass{article}

\usepackage[margin=1in]{geometry}
\usepackage[utf8]{inputenc}
\usepackage[T1]{fontenc}
\usepackage{microtype}
\usepackage{hyperref}
\usepackage{url}
\makeatletter\g@addto@macro{\UrlBreaks}{\do\-}\makeatother
\usepackage{booktabs}
\usepackage{amsfonts}
\usepackage{amsmath}
\usepackage{amssymb}
\usepackage{mathtools}
\usepackage{nicefrac}
\usepackage{xcolor}
\usepackage{graphicx}
\usepackage{longtable}
\usepackage{float}
\usepackage{array}
\usepackage{algorithm}
\usepackage{algpseudocode}
\usepackage{rotating}
\usepackage[capitalize,noabbrev]{cleveref}
\usepackage{enumitem}

\usepackage{natbib}
\newcommand{\figonewidth}{\textwidth}

\newcommand{\codeavailability}{Code: \href{https://github.com/aims-foundations/evarium}{\nolinkurl{github.com/aims-foundations/evarium}}. Website: \href{https://aimslab.stanford.edu/evaluation-ecosystem}{\nolinkurl{aimslab.stanford.edu/evaluation-ecosystem}}.}

\newcommand{\showcomments}{0}
\newcommand{\authcomment}[3]{%
\ifnum\showcomments=1
  {\textcolor{#2}{\textbf{[#1: #3]}}}%
\fi}

\title{The AI Evaluation Ecosystem}
\date{}

\author{%
  Yash Dave$^{1,*}$ \quad\quad
  Sang T. Truong$^{1,*}$ \quad\quad
  Serena Wang$^{2,\dagger}$ \quad\quad
  Sanmi Koyejo$^{1,\dagger}$\\
  \\
  \small{$^{1}$Stanford University \quad $^{2}$University of British Columbia} \\
  \small{$^{*}$Equal contribution \quad $^{\dagger}$Equal advising} \\
  \\
  \small{Correspondence: \texttt{yashdave@stanford.edu}}
}

\begin{document}

\maketitle

\begin{abstract}
    AI evaluation shapes the decisions of model providers, users, funders, and regulators. We argue that designing valid benchmarks requires contextualizing design choices in the dynamics of this ecosystem of actors. We develop a simulation architecture that combines rule-based market dynamics with LLM-driven strategic actors, building on advances in Generative Agent-Based Modeling (GABM). We model benchmarks, consumer needs, and provider capabilities as vectors over a six-dimensional capability space (reasoning, coding, knowledge, safety, communication, agentic), with structural information partitions across actors. As a case study, we apply this stylized simulation to explore benchmark holdout design. We find that moving from public benchmarks to private holdout benchmarks shrinks the gap between benchmark scores and user satisfaction on most benchmarks but widens it on a few, depending on where holdout weights shift scoring credit. We stress-test our findings at both the instrument and case-study level, drawing on the V\&V framework of \citet{sargent2013verification} and GABM-specific evidence criteria. Beyond holdout design, our simulation is a hypothesis-generating sandbox for studying how evaluator and policy choices, in turn, reshape the ecosystem.
\end{abstract}

\section{Introduction}
\label{sec:intro}

Benchmarks have been among the most influential coordination devices in modern AI research. The lasting ones do more than measure: each defines what counts as progress and proposes a roadmap for follow-on research, as MMLU, HELM, and Terminal-bench each did~\citep{hendrycks2020measuring,liang2023holistic,merrill2026terminalbench}. Yet methodology for benchmark design still treats benchmarks as one-shot instruments, evaluating them on properties (calibration, validity, contamination resistance) that hold the rest of the ecosystem fixed. Treating evaluation as a closed-loop system instead changes what counts as an evaluation question: not only ``is this metric well-calibrated?'' but ``what does the ecosystem do with this benchmark once it exists?''\footnote{\codeavailability}

\begin{figure}
    \centering
    \providecommand{\figonewidth}{0.85\textwidth}
    \includegraphics[width=\figonewidth]{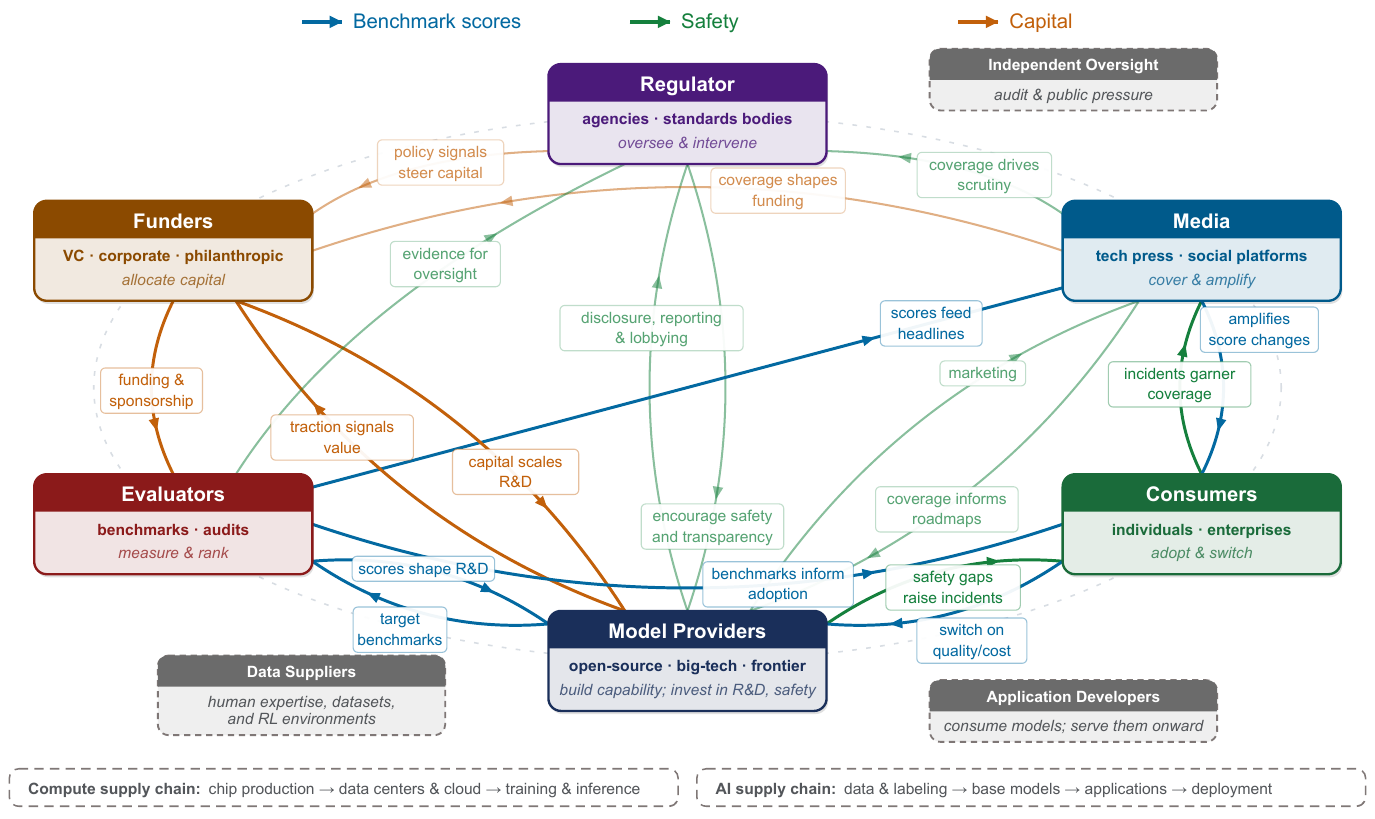}
    \caption{The AI evaluation ecosystem: six modeled actor types and the feedback loops between them, along three pathways. Fainter arrows are softer institutional channels. Grey dashed boxes mark actors and adjacent supply chains~\citep{hopkins2025ai} that the simulation leaves out.}
    \label{fig:ecosystem}
\end{figure}

Once a benchmark gains traction, its influence runs in both directions (Figure~\ref{fig:ecosystem}). The same scores that signal progress also steer research directions and inform model selection~\citep{bommasani2021opportunities,liang2023holistic}; once optimized as targets, they distort what they were meant to track~\citep{thomas2020problem}. The same pull plausibly extends to where funders direct capital, what companies red-team, and which failure modes regulators attend to. Adjacent high-stakes industries (clinical trials and aviation) suggest the gap matters: mature governance of such feedback-rich measurement systems, through mandatory disclosure, independent audit, and incident reporting, stabilizes information environments that private incentives do not (Appendix~\ref{app:related-extended}). AI evaluation has no comparable structured account of these joint dynamics; without one, policy proposals and institutional designs risk addressing symptoms rather than causes.

The questions an ecosystem perspective raises are known to be difficult to study: for example, how holdout choices reshape provider incentives; when evaluator independence affects market structure; or which feedback channels stabilize or amplify gaming. Formal game-theoretic models become intractable as actor heterogeneity, information asymmetries, and the number of feedback channels grow. These questions are also difficult to study empirically due to their counterfactual nature, and experimental data is not always available.

Recent advances in generative agent-based modeling~\citep{park2023generative,vezhnevets2023concordia} present a new opportunity to advance our understanding of ecosystem dynamics: actors reason over textual state (incident reports, competitor moves, regulatory news) the way real strategy teams and regulatory staff do, and a single design lever can be swapped while the rest holds fixed. 
The methodology to run such simulations has emerged only recently, with validation still an open problem~\citep{manning2024automated,ghaffarzadegan2024generative,larooij2025validation,zhou2026pimmur}; the phenomena this perspective anticipates, contamination~\citep{sainz2023nlp,balloccu2024leak} and leaderboard manipulation~\citep{singh2025leaderboard}, are now documented; and evaluation results now carry consequences on both sides of the market: the EU AI Act obliges providers of general-purpose models with systemic risk to evaluate them under standardised protocols~\citep{euaiact2024}, while frontier developers have tied their own deployment decisions to internal capability thresholds~\citep{anthropic2023rsp,openai2023preparedness,deepmind2024fsf}. 
Thus, in this work, we develop a framework to analyze the AI evaluation ecosystem. We operationalize this framework through a GABM to study one \textbf{design lever}, a choice the evaluator alone controls: whether a benchmark's items and their dimension weighting are published or withheld, which we refer to as \textbf{holdout design}. The simulation serves as a hypothesis generator rather than a predictor. It surfaces candidate dynamics, together with the conditions they depend on, for empirical work and policy analysis to test.
Our contributions can be summarized as follows:
\begin{itemize}[left=0pt, itemsep=1pt, topsep=0pt]
    \item We describe \textbf{an ecosystem-level framework for AI evaluation} that includes providers, evaluators, consumers, regulators, funders, and media coupled through feedback loops across development, market, and regulatory dynamics; distinct from supply- and value-chain framings (\S\ref{sec:system}). 
    \item To study this, we propose a \textbf{hybrid generative agent-based model (GABM).} LLM-driven actors reason over observable state with structurally enforced public/private/ground-truth information partitions; a heuristic mode isolates structural from behavioral dynamics (\S\ref{sec:simulation}).
    \item We apply this model in \textbf{a concrete simulation study on holdout design,} showing that in the context of the ecosystem, private-holdout benchmarks do not always reduce the gap between the benchmark score and user satisfaction (\S\ref{sec:results}).
    \item We \textbf{stress-test our simulation methodology} at the instrument level (is the simulation a reasonable laboratory?) and the case-study level (does the holdout-design finding hold up?), drawing on the V\&V framework of \citet{sargent2013verification}, the principles of \citet{zhou2026pimmur} for LLM multi-agent simulations, and the evidence hierarchy \citet{vezhnevets2023concordia} adapt for GABM (\S\ref{sec:validation}, Appendix~\ref{app:validation}).
\end{itemize}

\section{Related Work}
\label{sec:related}

Recent work on AI supply chains traces how foundation models, data, and services flow through networks of specialized actors~\citep{hopkins2025ai,cobbe2023supply,widder2023dislocated,bommasani2023ecosystem}. Evaluation sits orthogonal to these framings: it is neither a product flowing downstream (the supply-chain view) nor one of the value-creating activities into which a firm is disaggregated (the value-chain view of \citet{porter1985competitive}), but a shared measurement infrastructure on which providers self-assess, consumers choose, funders allocate, and regulators intervene. Our work studies that layer as a distinct ecosystem, building on three adjacent lines of research: AI evaluation and benchmarking, governance of information systems (treated in Appendix~\ref{app:related-extended}), and generative agent-based modeling with LLM actors.

\noindent \textbf{AI evaluation and benchmarking.}
Frameworks such as HELM~\citep{liang2023holistic} set standards for coverage and standardized, multi-metric measurement, and the Foundation Model report~\citep{bommasani2021opportunities} calls for valid evaluation across many criteria; preference evaluations such as Arena (formerly LMArena and Chatbot Arena)~\citep{zheng2023judging, chiang2024chatbot} complement them with dynamic rankings. A parallel literature documents evaluation pathologies: metric gaming~\citep{thomas2020problem,manheim2019categorizing} (kin to Campbell's Law and the Lucas critique~\citep{campbell1979assessing,lucas1976critique}), contamination~\citep{sainz2023nlp,balloccu2024leak,magar2022data}, critiques of construct validity~\citep{jacobs2021measurement,raji2021ai}, and meta-scientific analyses of how benchmark adoption concentrates on a few canonical datasets~\citep{dehghani2021benchmark,koch2021reduced}. These works address failure modes at the individual-benchmark level; we model evaluation within an ecosystem in which competitive dynamics, capital allocation, and consumer behavior produce failures even when individual benchmarks are well designed.

\noindent \textbf{Generative agent-based modeling with LLMs.}
Generative agent-based models (GABMs) have developed along two axes. Methodologically, work including \citet{aher2023using}, \citet{park2023generative}, \citet{gao2023s3}, \citet{manning2024automated}, and \citet{cross2025validating} has built increasingly rigorous methods for designing and validating generative-agent simulations, and \citet{ghaffarzadegan2024generative} add robustness and prompt-sensitivity checks. In parallel, \citet{larooij2025validation} argue that LLMs' black-box structure, cultural biases, and stochastic outputs make validation harder, and \citet{wu2025boundary} document variance underrepresentation in LLM actors. Our simulation restricts LLM reasoning to strategic planning (investment allocation, regulatory choice, funding allocation) while grounding market mechanics, scoring, and state transitions in deterministic rules: a hybrid that preserves LLM expressiveness without sacrificing the causal transparency that ABMs require. Extended discussion of this design choice is in Appendix~\ref{app:related-extended}.
\section{Mapping the AI Evaluation Ecosystem}
\label{sec:system}
Evaluation does not occur in isolation: it unfolds through the interactions of actors with different goals, information access, and decision rights, whose responses to evaluation outcomes reshape what future evaluations measure. Figure~\ref{fig:ecosystem} illustrates the six actor types and their primary information flows.

This section sets out the frame: who the actors are, what each of them observes, and which channels carry evaluation results between them. Section~\ref{sec:simulation} turns the frame into an instrument, a simulation in which every channel is explicit and can be altered one at a time, and Section~\ref{sec:results} puts the instrument to work on a single case study, the design of benchmark holdouts.

\noindent \textbf{Model Providers.} Model providers (e.g., OpenAI, Google, Anthropic, open-source consortia) develop and deploy AI systems, observing internal signals (user data, internal red-team results, commercial metrics) and external signals (public benchmarks, competitor performance, media coverage, policy requirements). They release models and adjust training strategies in response to benchmark performance.

\noindent \textbf{Evaluation Providers.} Evaluation providers (academic groups, independent labs, third-party efforts such as Stanford CRFM's HELM~\citep{liang2023holistic}, ARC Evals~\citep{shevlane2023model}, and third-party auditors~\citep{raji2022outsider}) design benchmarks, publish scores, and retire outdated evaluations. They may face conflicts of interest when paid by, or selling services to, the providers they evaluate~\citep{costanzachock2022who,raji2022outsider}. Third-party evaluators have since proposed minimum operating conditions for independent evaluation, covering model access, contingent compensation and recusal~\citep{aievaluatorforum2025aef1}.

\noindent \textbf{Consumers.} Consumers range from individuals to organizations facing integration costs and regulatory exposure. Benchmark scores reach them through two channels: a direct leaderboard signal, weighted by how much a consumer trusts published scores relative to their own experience (a modeling choice motivated by practitioner interviews in~\citet{hardy2025benchmarks}), and indirect exposure through news coverage of score movements and incidents. Scores serve as proxies for capabilities consumers cannot directly assess, a credence good~\citep{dulleck2006credence}, and adoption decisions drive provider revenue and funder attention.

\noindent \textbf{Regulators.} Regulators observe evaluation results through public leaderboards, technical reports, and audits, and can mandate compliance evaluations or enforce disclosure~\citep{euaiact2024}, require developer reporting as US policy did until 2025~\citep{whitehouse2023eo14110}, or issue voluntary frameworks~\citep{nist2023}. They cannot observe provider capabilities or consumer satisfaction directly, acting instead on public signals, incident reports, and media coverage.

\noindent \textbf{Funders.} Funders (venture capital, corporate investors, government agencies, foundations) allocate capital based on benchmark scores, market adoption, media sentiment, and incident history. Their decisions are performative~\citep{mackenzie2006engine,perdomo2020performative}: capital flows toward high-scorers, reinforcing the competitive value of benchmark optimization.

\noindent \textbf{Media.} Media outlets observe scores, market shifts, incidents, and regulatory actions. Selective coverage creates an attention economy that amplifies or attenuates signals other actors receive.

\section{Generative Agent-Based Model of the Evaluation Ecosystem}
\label{sec:simulation}

To study how the feedback dynamics described in Section~\ref{sec:system} produce emergent failures, we developed a GABM modeling six providers, an evaluator, a media actor, a regulator, four funder types, and 51 consumer segments, with actor decisions implemented in both LLM-driven and heuristic modes. The unit of modeling is the institution, not a human: LLM agents play organizations (a provider's strategy team, an investment committee, a regulatory body) making monthly allocation decisions, while the one mass-human population, consumers, is rule-based throughout. Following RecSim's view of simulators as controlled environments that need not be faithful to live systems~\citep{ie2019recsim}, we use the simulation as a hypothesis generator: the results surface candidate dynamics worth investigating empirically, rather than proving that specific failures will occur in real systems. Several structural features described below are deliberate modeling choices: the six-dimensional capability ontology, hand-calibrated consumer need weights, the linear scoring model, the form of holdout-weight perturbation, and a dedicated safety lever. The simulator tests mechanism plausibility under this stylized structure: outcomes such as the score--satisfaction gap are partly defined by it, and the experiments show which patterns emerge predictably from interactions on top.

\subsection{Benchmark Model and Structural Misalignment}
\label{sec:benchmarks}

Each provider's capabilities are represented by a six-dimensional \textbf{capability vector} $\mathbf{c}_p$ across reasoning, coding, knowledge, safety, communication, and agentic capabilities. Each benchmark $b$ has public task categories but \textit{hidden} per-category dimension loadings $\mathbf{w}_b$ known only to the simulation. At round $t$, a noisy measurement is drawn and the published score is the running max: $\hat{s}_{p,b,t} = \max (\hat{s}_{p,b,t-1}, \; \tilde{s}_{p,b,t})$, where $\tilde{s}_{p,b,t} \sim \mathcal{N}\! (\mathbf{c}_p^\top \mathbf{w}_b,\; \sigma_b^2/n_b)$ and $n_b$ is the number of scored items on benchmark $b$. The monotonic non-decreasing constraint models public reporting behavior. No actor, including LLM-driven agents, can access the true dimension weights.

The simulation begins with four benchmarks at round~0, anchored to real evaluations (MMLU, HumanEval, TruthfulQA, MT-Bench), and introduces nine more in a preset order over the 40-round horizon, for 13 active benchmarks total (Appendix~\ref{app:benchmark-pool}; the dynamic-mode draws from an extended 22-benchmark pool).

The ecosystem is \textit{structurally misaligned}: averaged over the 13-benchmark suite, benchmark weights exceed population-average consumer need weights on reasoning and coding, and they fall below them on communication early in a run and on safety late in a run, as the organizational share of consumers grows (per-dimension arithmetic in Appendix~\ref{app:benchmark-pool}). The suite is modeled on the 2023 benchmark landscape, whose newer benchmarks target coding and advanced reasoning~\citep{stanfordaiindex2024}. A provider that follows benchmark scores therefore invests in a different mix of dimensions than consumers need. The experimental claim is that under stable ecosystem-wide investment asymmetries (e.g., consistent over-investment in reasoning relative to safety or knowledge), the resulting per-benchmark gaps become structured and predictable rather than random across conditions.

\subsection{Provider Investment and Emergent Gaming}
\label{sec:gaming}

Each round, providers allocate their training budget across three levers: \textbf{R\&D} directs capability growth toward dimensions the provider believes benchmarks and consumers value; \textbf{Safety} improves the safety dimension and reduces incident probability; \textbf{Product} quality-gates the consumer-need signal feeding R\&D targeting and raises effective switching costs. Allocations are applied as a 3-round rolling average to model execution lag between strategic decision and operational effect. Full mechanics in Appendix~\ref{app:investment}.

\noindent \textbf{Belief update.}
Providers maintain per-benchmark beliefs about hidden weights ($\hat{\mathbf{w}}_{p,b}$), initialized from a noisy reading of $\mathbf{w}_b$ ($\sigma{=}0.05$ per dimension, clipped and renormalized) that stands for what a benchmark's stated scope reveals, and updated each round from score prediction errors (learning rate $0.10$--$0.20$ by provider profile). The exact vector remains unobservable throughout. Belief updating is rule-based in both LLM and heuristic modes and reflects the epistemic state of the provider.

\noindent \textbf{Satisfaction and the gap.}
Published scores shape which provider a segment adopts, but not how satisfied the segment is with it: a segment's satisfaction is computed from the provider's true capabilities, weighted by the dimensions that segment needs. The headline quantity is the per-benchmark \textbf{score--satisfaction gap} $g$: a benchmark's published score minus the \textit{matched satisfaction} the same provider delivers, an average over consumer segments weighted by segment size and by how closely each segment's needs align with that benchmark's dimension weights $\mathbf{w}_b$, so that $g$ compares a benchmark's score against the consumers whose needs it most closely matches. Consumer switching adds incident and cost terms to satisfaction (Appendix~\ref{app:consumers}); $g$ uses the capability term alone. A positive gap means the benchmark promises more than consumers get; a negative gap means it promises less.

\noindent \textbf{Emergent gaming.}
The score--satisfaction gap arises as a measurement artifact when two structural conditions are simultaneously met: (1) benchmark dimension weights diverge from consumer need weights, and (2) providers discover, through accumulated score prediction errors, that benchmark-focused investment produces faster score gains than consumer-focused investment. This operationalizes Goodhart's Law~\citep{manheim2019categorizing,thomas2020problem} as a structural consequence of information architecture: the gap is a property of the system's design, not a modeled strategy.

\subsection{Ecosystem Actors and Feedback Channels}
\label{sec:actors}

The simulation operationalizes the actor relationships from Section~\ref{sec:system} as concrete feedback channels. Information access is structurally enforced across three tiers: \textit{public} state (scores, leaderboard, market shares) visible to all; \textit{private} state (beliefs, strategies) visible only to the owning agent; and \textit{ground truth} (capability vectors, benchmark dimension weights, true consumer satisfaction) held by the simulation and inaccessible to any actor prompt.

Figure~\ref{fig:simflow} (Appendix~\ref{app:round-protocol}) summarizes the per-actor mechanics and their round-order dependencies; only design choices that are load-bearing for later results are called out here. The \textbf{heterogeneous consumer market} comprises 51 segments (17 use-case profiles, each crossed with the three behavioral archetypes of its consumer type, individual or organizational; archetypes vary in leaderboard trust, switching cost, and cost sensitivity), with \emph{experience-based} satisfaction that reflects capability actually delivered to a segment rather than the published score, making the score--satisfaction gap structural rather than a modeled strategy (Appendix~\ref{app:consumers}).

\textbf{Incidents} are drawn from a four-level severity distribution (minor 50\%, moderate 31\%, major 12\%, critical 7\%) and propagate simultaneously to all downstream actors (Appendix~\ref{app:incidents}). \textbf{Media} influence runs through attention, not direct satisfaction (Appendix~\ref{app:media}): negative coverage raises the fraction of a provider's users who enter exploration. The \textbf{regulator} applies a graduated escalation ladder with cooldowns between interventions (Appendix~\ref{app:regulator}). Four \textbf{funder} types (VC, corporate, government, foundation) allocate capital from fixed pools into the provider budget via Eq.~\ref{eq:budget}, closing the performativity loop of Section~\ref{sec:system}; VC funders do not fund open-source providers (Appendix~\ref{app:funders}).
One \textbf{model provider} (OpenCore) is an open-weight lab that differs from the closed providers in its structure: VC funders do not fund it, it has a cost advantage that represents self-hosting savings, its safety capability floor is slightly lower, and its open weights move the other providers' benchmark beliefs toward the public weights each round. These differences are patterned on differences between open-weight and closed labs (Appendix~\ref{app:open-source}).

\subsection{Round Protocol and Experimental Setup}
\label{sec:protocol}

Figure~\ref{fig:simflow} in Appendix~\ref{app:round-protocol} shows the order of events in a round. In Phases~1 to~4, providers plan and update capabilities, the evaluator scores all providers and publishes the leaderboard, providers update beliefs from score prediction errors, incidents are generated, and the media covers the round. In Phase~5, consumers, the regulator and funders observe the round's public state and respond in that order. Funding and regulatory effects reach providers at the start of the next round. Full pseudocode is in Appendix~\ref{app:round-protocol}.
The split between the LLM and heuristic modes is methodologically intentional: heuristic mode supplies the statistical structure (large-$N$ distributional claims about direction and ordering), while LLM mode supplies strategic reasoning traces (Appendix~\ref{app:llm-results}) that allow qualitative auditing of \emph{why} actors take the actions they do. The lower seed count limits distributional inference in LLM mode; we therefore use LLM runs primarily for qualitative reasoning traces and cross-checking directional patterns.

\noindent \textbf{Round-based protocol and sequential phasing.}
A round maps to the cadences we model (budget cycles, benchmark releases, intervention windows, funding rounds) and keeps LLM decision points interpretable. Within a round, actors resolve sequentially so later actors observe earlier outcomes, mirroring real-world information flow: consumers respond to the month's scores, regulators to its incidents, funders to its leaderboard. Providers, the regulator and funders are LLM-driven; the evaluator is optionally LLM in dynamic mode; the media actor and the consumer market are rule-based throughout. A fully heuristic mode serves as a non-LLM baseline, separating structural dynamics from LLM-specific behavior. Prompts are framed in the natural language of each role (strategy teams, investment committees, regulatory staff) without apparatus vocabulary. Hyperparameters are consolidated in Appendix~\ref{app:parameters}.

\begin{figure}
  \centering
  \includegraphics[width=0.95\textwidth]{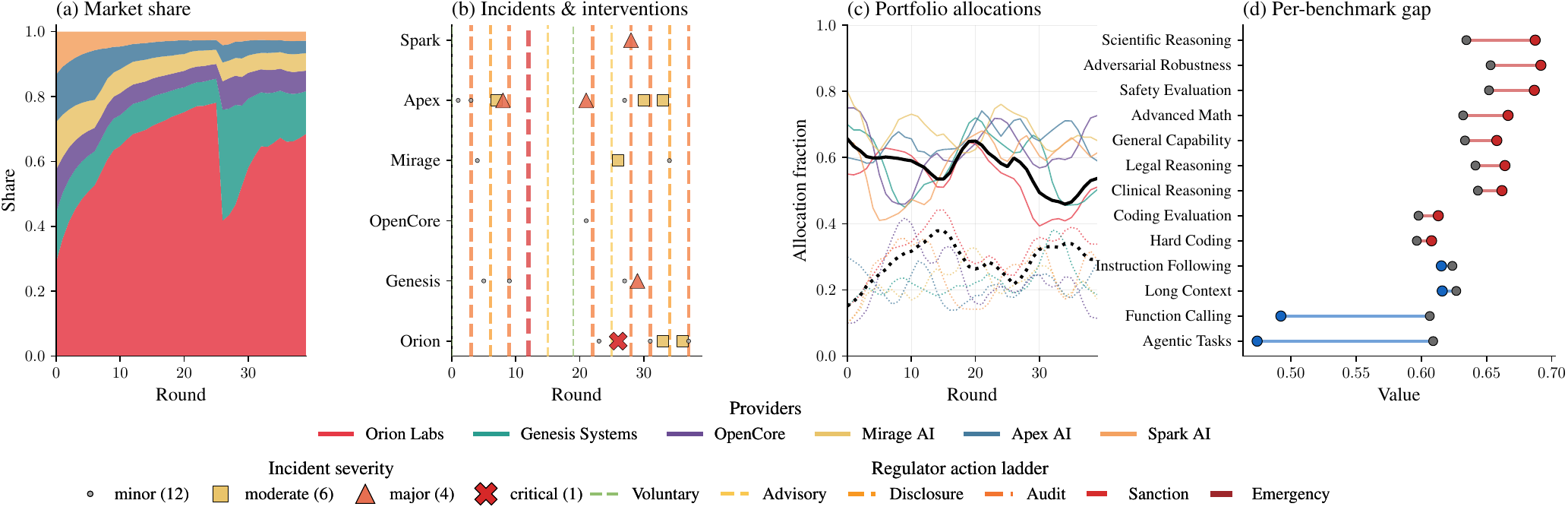}
  \caption{One 40-round LLM-driven run, calibrated \texttt{baseline}, Claude~Opus~4.6, seed~$43$ (\texttt{core\_privacy/llm/claude-opus-4-6/baseline/seed\_43}). (a) Stacked market share. (b) Per-provider incident markers (shape = severity) and regulator interventions (vertical dashed lines, color = action on the escalation ladder). (c) Portfolio allocations: R\&D (solid) and safety (dotted) per provider, share-weighted mean in bold. (d) End-of-run per-benchmark score (colored) vs matched consumer satisfaction (gray); red = benchmark overpromises, blue = underpromises.}
  \label{fig:single_run_loop}
\end{figure}

\noindent \textbf{Example run.}
Figure~\ref{fig:single_run_loop} shows one baseline run. The leader accumulates market share through round~$25$. A critical incident then costs it $36$pp of share in a single round, the regulator commissions an audit of it five rounds later, and it recovers only to ${\sim}68\%$ by round~$40$ (panels~a,~b). Safety allocations span ${\sim}10$--$44\%$ across providers, and the ranking among them changes repeatedly over the run (panel~c). Published score and matched consumer satisfaction diverge non-uniformly across the 13-benchmark suite (panel~d).

\section{Results: Case Study of Benchmark Holdout Design}
\label{sec:results}

This section uses the simulation as a hypothesis generator: varying one structural property of the evaluator at a time, we trace how evaluation practice propagates through provider investment, consumer choice, and capital allocation. The patterns below are candidate dynamics, not claims about real-world markets. Evidence is organized into two layers. The \textit{structural} layer draws on the heuristic baseline ($N{=}50$ seeds per condition) and supports distributional claims. The \textit{behavioral} layer draws on LLM-driven runs ($N{=}10$ Sonnet seeds per condition, matched seeds for paired comparisons; cross-vendor robustness in Appendix~\ref{app:model-robust}). The two layers cross-check each other. Agreement in direction across modes suggests that a finding does not depend on LLM-specific prompt framing or on a particular heuristic rule. Both modes share the same scoring layer, so the check bears on the planner's contribution to capability and leaves the scoring mechanism itself untested. Published score and matched consumer satisfaction do not agree uniformly across benchmarks (Figure~\ref{fig:single_run_loop}, panel~d); a single pooled score--satisfaction gap hides that the gap moves differently on different benchmarks. The analyses below ask what determines each benchmark's direction, and what an evaluator's privacy choice does to it.

\subsection{Holdout design: private holdouts shrink the gap unevenly across benchmarks}
\label{sec:privacy-ladder}

The evaluator's first lever is what it publishes, an information-design problem~\citep{kamenica2011bayesian}. We vary holdout design across a five-condition ladder over the $13$-benchmark active set:
\texttt{public\_only}, \texttt{baseline} (an $8/3/2$
public/partial/private mix reflecting the coexistence of public, semi-private, and fully-private evaluation regimes in $2024$--$2025$~\citep{scale2024seal,phan2025humanity,white2024livebench}, building on the public/private split ARC introduced~\citep{chollet2019measure}),
\texttt{private\_dominant}, \texttt{private\_only} (all benchmarks
holdout-only with perturbed weights, target cosine $0.85$), and \texttt{iid\_holdout}
(same weights as public, isolating reporting lag from weight asymmetry). The cosine knob aggregates contamination, training-on-the-test-task, and adversarial-construction sources of public-vs-holdout asymmetry; mechanism details in Appendix~\ref{app:private-benchmark-reporting}. The cleanest privacy contrast is
between the two $100\%$-endpoint conditions; the calibrated \texttt{baseline}
pools privacy with which benchmarks happen to carry the \texttt{partial} label.

\begin{figure}[!htb]
  \centering
  \includegraphics[width=\textwidth]{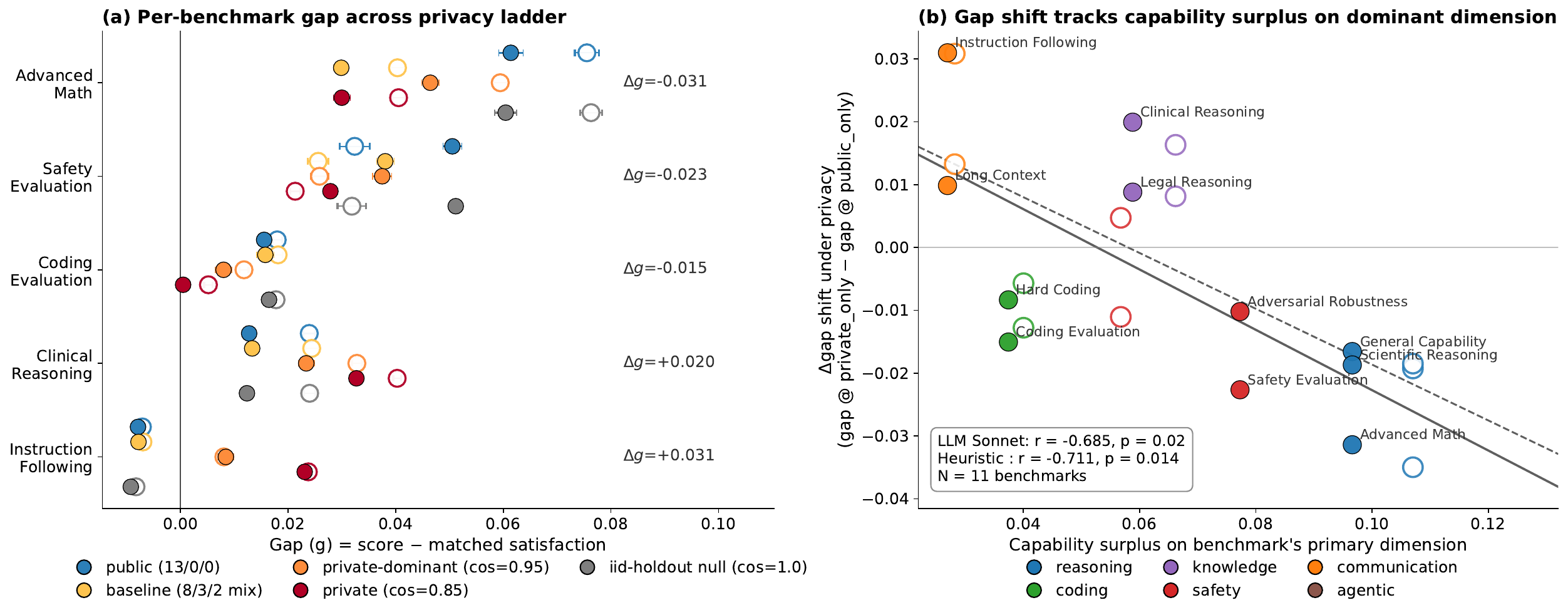}
  \caption{\textbf{(a)} Per-benchmark gap $g$ (published score $-$ matched
  consumer satisfaction, where matched satisfaction is ground-truth capability
  scored against consumer need-weights and averaged over segments, each
  weighted by its size and by its alignment with the benchmark's dimension
  weights) across the five
  privacy conditions, for five representative benchmarks (one per primary
  capability dimension; full $13$-benchmark version in
  Appendix~\ref{app:privacy-ladder-ext}). Foreground: LLM Sonnet~4.6
  ($N{=}10$) with $95\%$ CIs. Background: heuristic mode ($N{=}50$, all five
  conditions). $\Delta g$ per
  row $=$ gap@\texttt{private\_only} $-$ gap@\texttt{public\_only}.
  \textbf{(b)} $\Delta g$ vs.\ population capability surplus on each
  benchmark's primary dimension (capability on the highest-weighted dimension
  minus the cross-dimension mean, computed from \texttt{public\_only} seeds),
  over $11$ benchmarks; Agentic Tasks and Function Calling are excluded because
  the agentic dimension starts far below the other five (initial capability
  $0.08$ to $0.12$ against $0.25$ to $0.55$, Table~\ref{tab:providers}), which
  makes their surplus values extreme. The exclusion is conservative:
  Appendix~\ref{app:privacy-ladder-ext} reports the fit on all $13$, where the
  slope tightens.}
  \label{fig:privacy_ladder}
\end{figure}

The mechanism is weight asymmetry between public and holdout dimension weights (Figure~\ref{fig:privacy_ladder}b). Under \texttt{private\_only}, scoring weights are perturbed from the true weights (target cosine $0.85$), leaking credit from the dominant dimension onto adjacent ones. When credit moves from a stronger dimension to weaker ones, score drops; when it moves from a weaker dimension to stronger ones, score rises. The lower the population's capability surplus on the dominant dimension, the more $\Delta g$ rises ($r{=}{-}0.69$ LLM, ${-}0.71$ heuristic, over the $11$ benchmarks of panel~b). Matched satisfaction (true weights) is unaffected, so the gap moves.

Privacy does not uniformly shrink the gap. From \texttt{public\_only} to \texttt{private\_only}, the absolute gap shrinks on $10$ of the $13$ benchmarks and widens on three (Instruction Following, Clinical Reasoning, Legal Reasoning) in LLM mode, and shrinks on $9$ of $13$ in heuristic mode, which also widens Adversarial Robustness, the lone cross-mode disagreement, where both shifts are small (Figure~\ref{fig:privacy_ladder}a labels $\Delta g$ for the five representative rows; all $13$ in Appendix~\ref{app:privacy-ladder-ext}). The widening is stable across seeds: the three widen in $10$ of $10$ LLM seeds and $96$--$100\%$ of heuristic seeds, while under the \texttt{iid\_holdout} null (identical weights to \texttt{public\_only} but private reporting) every benchmark widens in $40$--$70\%$ of seeds (Table~\ref{tab:holdout-widening}). Along the perturbation axis (target cosine $1.0$, $0.95$, $0.85$), each benchmark's gap moves monotonically in its own direction: the pattern is consistent with weight asymmetry rather than reporting lag or observation noise.

\noindent \textbf{Worked example: Clinical Reasoning.} Clinical Reasoning loads primarily on \emph{knowledge} (weight $0.65$). The suite places more of its weight on reasoning (${\approx}0.26$) than on knowledge (${\approx}0.16$; Appendix~\ref{app:benchmark-pool}) and R\&D growth tracks those weights, so providers' knowledge sits below their reasoning. Under \texttt{private\_only}, holdout perturbation leaks credit from knowledge onto \emph{reasoning}, where providers are strong; score rises while matched satisfaction (capabilities unchanged) does not. A benchmark nominally about medical reasoning ends up rewarding reasoning-strong models: the ``accidental credit'' direction. The finding is conditional: a benchmark widens when its holdout weights shift credit toward stronger dimensions. We do not test whether any real benchmark has that property; it is a separate measurement question, which \citet{hardy2026cartography} approach through factor analysis of leaderboard data. In the simulation, the hand-authored holdout directions and the suite determine which benchmarks widen. In a suite whose round-0 set includes a domain-expertise benchmark, Clinical and Legal Reasoning never widen in heuristic seeds, while Instruction Following widens in every seed of every suite tested (Appendix~\ref{app:privacy-ladder-ext}).

\noindent \textbf{Consequence for evaluator design.}
Privacy is not a uniform tool for mitigating benchmark-gaming. It redistributes scoring credit across dimensions, toward or away from where the R\&D architecture already concentrates capability; the benchmark suite shapes where that is. Changing the suite's composition changes which benchmarks widen under privacy, as the worked example shows for Clinical and Legal Reasoning, though not for Instruction Following. Under a non-flat capability ontology (\S\ref{sec:limitations}) there is no neutral benchmark suite: composition choices shape where perturbation-induced shifts land, and heterogeneous consumer needs shape which of those shifts widen the gap. In this light, a benchmark's utility is not an intrinsic property; it is conditional on the ecosystem it enters.

\subsection{Validation}
\label{sec:validation}

Validation covers two tasks: \textbf{instrument validation} (whether the simulation is a reasonable laboratory) and \textbf{case-study validation} (whether the holdout-design finding survives perturbation). The case-study evidence is a stack of mechanism-isolation checks: cross-mode sign agreement on $12$ of $13$ benchmarks, the \texttt{iid\_holdout} null, cross-vendor replication on Opus~4.6 and GPT-5.5 ($N{=}3$ seeds each), a within-benchmark rescoring, and a fitted null recovering $97\%$ of $\Delta g$ variance. These checks are detailed in Table~\ref{tab:validation}, the prose below, and Appendix~\ref{app:validation}. We organize this within the V\&V activities of \citet{sargent2013verification} and the PIMMUR audit of \citet{zhou2026pimmur} (Appendix~\ref{app:pimmur}); in the evidence hierarchy \citet{vezhnevets2023concordia} adapt for GABM we sit on the lower rungs below its gold standard of prediction on new real-world data (theory consistency, plus the model-comparison and robustness practices it recommends), since no real-world counterfactual ecosystem is available.

\begin{table}[!htb]
\centering
\caption{Validation evidence by task. The instrument column records design guarantees and calibration provenance; case-study validation stress-tests the holdout-design $\Delta g$ result.}
\label{tab:validation}
\footnotesize
\setlength{\tabcolsep}{4pt}
\renewcommand{\arraystretch}{1.1}
\begin{tabular}{p{1.5cm}p{5cm}p{6cm}}
\toprule
 & \textbf{Instrument checks (design \& calibration)} & \textbf{Case study validation (holdout design)} \\
\midrule
\textbf{Approach} & Sargent V\&V activities; each structural assumption rated calibrated, anchored, or stipulated. & Mechanism-isolation tests on the per-benchmark $\Delta g$ result. \\
\addlinespace
\textbf{Evidence} & Three-tier visibility enforced structurally (Appendix~\ref{app:architecture}); parameter grounding ($K{=}3$ calibrated, cosine targets $0.85/0.95$ anchored, need-weights anchored in sector-level evidence; Appendices~\ref{app:parameters} and~\ref{app:assumption-status}). & Cross-mode sign agreement on 12 of 13 benchmarks; cross-model replication on Opus 4.6 and GPT-5.5 ($N{=}3$ seeds each; Appendix~\ref{app:model-robust}); \texttt{iid\_holdout} channel isolation; within-benchmark rescoring (Appendix~\ref{app:rescoring}); fitted null recovers $97\%$ of $\Delta g$ variance (Appendix~\ref{app:null-model}). \\
\bottomrule
\end{tabular}
\end{table}

\noindent Heuristic planners cannot see the \texttt{public}/\texttt{partial}/\texttt{private} label, yet the two modes agree in sign on $12$ of $13$ benchmarks, so LLM-specific reasoning is unlikely to drive the result. Both modes share the scoring layer, which the rescoring below tests directly. The \texttt{iid\_holdout} null (cosine $=1.0$) isolates weight asymmetry from reporting lag and observation noise.

\noindent \textbf{What the simulation adds.} The holdout result has a simple description, and the simulation shows where it holds. A benchmark's gap shift is provider capability weighted by the difference between its holdout and public scoring weights. Rescoring end-of-run capability under both sets of weights reproduces the direction of the shift on all $13$ benchmarks and most of its size (Appendix~\ref{app:rescoring}). The description needs the run: starting capability gives the wrong direction on three benchmarks, because the widening set appears only after R\&D tilts capability toward what the suite emphasizes. A regression on starting conditions fits the shifts closely, mostly because weight differences are fixed for each benchmark, so it does not measure what the trajectory adds (Appendix~\ref{app:null-model}). The other questions in Table~\ref{tab:case-studies} have no obvious simple description. Checking a candidate against the full run is how one would find it, and in one small-sample comparison market concentration differs in sign between heuristic and LLM planners (Appendix~\ref{app:evaluator-capture-ext}).

\subsection{Limitations}
\label{sec:limitations}

The privacy mechanism in \S\ref{sec:privacy-ladder} rests on four modeling choices (Appendix~\ref{app:lim-modeling-choices}); broader action-space simplifications and methodological bounds of GABMs are in Appendices~\ref{app:lim-per-actor} and~\ref{app:lim-methodology}. The conclusions are conditional on the assumptions of \S\ref{sec:system} and \S\ref{sec:simulation}; Appendix~\ref{app:assumption-status} rates each as calibrated, anchored, or stipulated. The benchmark dimension loadings (Table~\ref{tab:benchmark-pool}) are stipulated, and every score and every gap is computed against them.

\noindent \textbf{1.\  Privacy operates through weight perturbation in this simulator.}
``Privacy'' in this simulator decomposes into three orthogonal channels (Appendix~\ref{app:private-benchmark-reporting}): a cosine-distance perturbation $\cos\theta$ between public and holdout dimension weights (targets of $0.85$ / $0.95$, anchored by within-family Pearson correlations we compute from Epoch data~\citep{epoch2024benchmarks} and by retro-holdout score inflation in \citet{haimes2024benchmark}), a $K{=}3$ reporting lag, and observation noise scaling with $\sqrt{\text{samples}\times h}$. The cosine channel is itself a \emph{composite proxy}: $\cos\theta < 1$ aggregates item-level contamination and memorization, ``training on the test task'' effects~\citep{dominguezolmedo2024training}, and adversarial holdout construction. The simulator does not separately resolve these sources: a result attributed to weight asymmetry may be driven dominantly by any one of them, or by their interaction. The \texttt{iid\_holdout} null ($\cos\theta = 1$) isolates the composite cosine channel from lag and noise; \texttt{private\_dominant} vs.\ \texttt{private\_only} isolates the marginal effect of moving from a target $\cos\theta$ of $0.95$ to $0.85$; realized cosines differ across benchmarks (Appendix~\ref{app:private-benchmark-reporting}). A within-benchmark rescoring separates the \emph{scoring-layer} effect (same capability, different scoring weights) from the \emph{R\&D-direction} effect (cosine perturbation drifts inferred weights, which steers R\&D): in heuristic mode the scoring layer carries $0.93$--$0.96$ of $\Delta g$, and the capability effect is indistinguishable from zero on $11$ of $13$ benchmarks (Appendix~\ref{app:rescoring}). The decomposition covers heuristic runs only; LLM providers, which reason about inferred weights, may place more of $\Delta g$ on the capability channel. Provider prompts describe the benchmark types neutrally, following \citet{zhou2026pimmur}'s Minimal-Control principle, so the result is untested for providers prompted to target holdout weights (Appendix~\ref{app:lim-methodology}).

\noindent \textbf{2.\ The six-dimension capability ontology is a design choice.}
We model six dimensions (reasoning, coding, knowledge, safety, communication, agentic), anchored to HELM- and LMSYS-style benchmark taxonomies and to where industry investment is currently separable. Capability ends up uneven across dimensions, because the suite weights some dimensions more heavily than others and safety carries a dedicated portfolio lever, and weight asymmetry between public and holdout weights acts on that unevenness. The mechanism is robust within this ontology (Figure~\ref{fig:privacy_ladder}), but a flatter ontology or different investment-axis separability could shift benchmarks between the shrinking and widening sets or eliminate the asymmetry entirely.

\noindent \textbf{3.\ The simulator covers only public-benchmark-driven competitive contexts.}
The instantiated ecosystem is leaderboard-driven, public-benchmark-dominated, multi-firm competitive, with a single global jurisdiction; six providers, no entry or exit. The mechanism applies to ecosystems with this structure but its engagement under FDA-style gatekeeping, defense procurement, or enterprise-B2B contexts (where buyers run internal proofs-of-concept) is unclear. We also do not model providers' \emph{internal} evaluation infrastructure: the capability tests informing Responsible-Scaling-Policy and Preparedness-style deployment decisions \citep{anthropic2023rsp,openai2023preparedness,deepmind2024fsf}, a recent (2023+) phenomenon and an additional pressure on R\&D direction not captured here.

\noindent \textbf{4.\ Consumer satisfaction is model-defined.}
Satisfaction is a deterministic function of the (ground-truth) capability vector and consumer need-weights, both held by the simulation. Need-weights are assigned to each consumer profile from sector-level evidence on what that profession uses AI for and what it is concerned about, together with occupation-level adoption rates in \citet{bick2024rapid} (Appendix~\ref{app:parameters}); the step from that evidence to preference weights is a researcher-imposed judgment not validated against user-reported preference. Need-weights do not enter a benchmark's shift under privacy, $\mathbf{c}\cdot(\mathbf{w}_{\mathrm{holdout}}-\mathbf{w}_{\mathrm{public}})$. They set the starting gap, whose sign decides whether that shift widens or shrinks it. Measured against other consumers the counts change: against safety-heavy segments alone, four of the $13$ classifications flip (Appendix~\ref{app:lim-modeling-choices}).

\noindent\textbf{Why the result is still useful.} The simulation is a laboratory, not a crystal ball. Private holdouts shrink the gap on most benchmarks and widen it on a few, those whose holdout weights shift credit toward dimensions where R\&D concentrates. This holds where leaderboards drive decisions, public benchmarks dominate, capability is uneven across dimensions, and gaps are measured against a mixed consumer population.

\subsection{A catalog of simulation case studies}
\label{sec:future-case-studies}

The privacy ladder in \S\ref{sec:privacy-ladder} is one case study; the ecosystem frame supports a broader class of policy and institutional-design questions that share the same architecture. Table~\ref{tab:case-studies} collects representative examples, each pairing a policy or mechanism question with its implementation slot in the current simulation and the strategic dynamic it would surface.

\begin{table}[!htb]
\centering
\caption{Additional case studies the ecosystem simulation can host. Case~1 (holdout design) is the worked example in \S\ref{sec:privacy-ladder} and case~2 (conflicts of interest) has a small-sample comparison in Appendix~\ref{app:evaluator-capture-ext}; cases~3 and~4 are presented as architectural slot-ins rather than worked studies. Further slots (media shadow, benchmark sponsorship) in Appendix~\ref{app:case-studies-ext}.}
\label{tab:case-studies}
\footnotesize
\setlength{\tabcolsep}{3pt}
\begin{tabular*}{\textwidth}{@{\extracolsep{\fill}}p{4.2cm}p{4cm}p{5.1cm}@{}}
\toprule
\textbf{Question (policy framing)} & \textbf{Simulation slot-in} & \textbf{Strategic angle} \\
\midrule
\textbf{2.\ Conflicts of interest} (Arena; Scale~SEAL). Does monetizing submissions redirect capital toward aggressive submitters at consumer cost? & \textit{Small-sample comparison in Appendix~\ref{app:evaluator-capture-ext}.} Per-submission fees (R\&D budget); best-of-$N$ selection; paid early access. & Larger-budget providers capture more submissions; funders read inflated scores as quality signals; capital flows to aggressive submitters rather than providers best serving users. \\
\midrule
\textbf{3.\ Transparency mandates} (EU AI Act Art.~53; CA SB-53; NY RAISE Act). Do forced allocation or capability disclosures reduce gaming? & New regulator lever: exposes part of provider private state (R\&D mix, internal eval results) to public. & Providers face disclose-vs-window-dress tradeoff; tests whether visibility of the \emph{R\&D mix} changes Goodhart pressure or just changes the narrative around it. \\
\midrule
\textbf{4.\ Audit \& verification} (UK AISI pre-deployment agreements; WH 2023 voluntary commitments). How effective are audits at enforcing voluntary safety commitments? & Regulator lever: provider commits to a forward safety allocation at $t$; regulator verifies at $t{+}k$ with penalty on violation. & Game-theoretic: provider can over-promise, under-promise, or credibly commit. Tests whether self-binding works when verification is credible, and how it interacts with incident pressure and market share. \\
\bottomrule
\end{tabular*}
\end{table}

The architecture treats each of these as a specific lever added to an existing actor, a new channel between actors, or a perturbation of an existing information flow. The role of the simulation is not to predict the real-world effect quantitatively but to reveal which ecosystem pathways carry the effect: which actors are pivotal, which information asymmetries amplify or dampen it, and where the intervention produces the counterintuitive dynamic that an empirical study could then target.

\section{Conclusion}
\label{sec:conclusion}
We argue that AI evaluation is best understood as a system of interacting actors whose incentives, information, and strategic behavior jointly determine what gets measured, how results are used, and which failure modes emerge. We developed a systems-level framework mapping these actors and feedback loops, instantiated it as an agent-based simulation, and used it to study one focused case: how does benchmark holdout design shape ecosystem outcomes?

Private holdouts shrink the score--satisfaction gap unevenly. From \texttt{public\_only} to \texttt{private\_only}, the gap shrinks on $10$ of the $13$ benchmarks and widens on three in LLM runs, in every seed. A benchmark widens when its holdout weights move scoring credit toward the dimensions where provider R\&D has concentrated capability, so the mechanism is weight asymmetry between public and holdout dimension weights. Three checks support this reading. Under the \texttt{iid\_holdout} null, which keeps the public weights and changes only the reporting, each benchmark widens in $40$--$70\%$ of seeds and the stable pattern disappears. Rescoring end-of-run capability under the holdout weights reproduces the direction of the change on all $13$ benchmarks. Heuristic and LLM planners agree on the sign for $12$ of $13$.

Which benchmarks widen depends on the suite. When the opening suite includes a domain-expertise benchmark, Clinical and Legal Reasoning never widen in heuristic runs, while Instruction Following widens in every suite we tested. There is no neutral benchmark suite, because the choice of benchmarks decides where the widening lands. The finding holds under the conditions the simulation models, a leaderboard-driven market with gaps measured against a mixed consumer population, and it needs a test on real leaderboards.

The simulation supplies what observation cannot. In the real ecosystem, true capability and provider R\&D allocation are unobservable, and no natural experiment holds a benchmark suite fixed while its holdout policy changes. In the simulation we change one evaluator choice and let providers respond, and the response creates the result. The same rescoring applied to starting capability gives the wrong direction on three benchmarks, so the widening set exists only after R\&D has tilted capability toward what the suite rewards. We use the simulation as a laboratory. It produces hypotheses, and the conditions they depend on, for empirical studies and policy work to test. It predicts no real market.

The same simulation can host other questions about evaluator and policy choices. Evaluator conflicts of interest has a small-sample comparison in Appendix~\ref{app:evaluator-capture-ext}, transparency mandates and audit verification are specified in \S\ref{sec:future-case-studies}, and media coverage and benchmark sponsorship are outlined in Appendix~\ref{app:case-studies-ext}. Each asks how one actor's choice moves through the rest of the ecosystem, and each can be examined before a policy depends on the answer. Holdout design is the first of these questions we have worked through.

\section*{Acknowledgments}
We thank the practitioners who spoke with us about evaluation practice. We also thank the AI Measurement Science reading group at Stanford University for taking part in the early mapping of the evaluation ecosystem and for their feedback. This work is partially supported by NSF 2046795, 2205329, and 2504264, NIH, ARPA-H, the MacArthur Foundation, Good Ventures, Schmidt Sciences, the Hasso Plattner Förderstiftung, and Stanford HAI. Truong is supported by a Microsoft Research Fellowship. Wang is supported by a Canada CIFAR AI Chair.

\bibliography{references}

\appendix

\section{Extended Related Work}
\label{app:related-extended}

This appendix extends \S\ref{sec:related} with five topics: governance, public-goods and performativity perspectives; the supply-chain and value-chain framings of AI development; the literature on generative agent-based models and their validation; the reasons for using a simulation and what other methods offer; and governance regimes in other industries.

\subsection{Governance, public goods, and performativity}
Our public-goods framing draws on economic analyses of information security~\citep{anderson2006economics,moore2010economics} and is informed by work on AI ethics and governance~\citep{chandrasekaran2021sok,whittlestone2019role,veale2021demystifying,brundage2020toward}. Documentation and audit proposals such as model cards~\citep{mitchell2019model}, datasheets~\citep{gebru2021datasheets}, internal audits~\citep{raji2020closing} and, more recently, structured reporting of evaluation results themselves~\citep{ghosh2026evaluationcards}, along with the EU AI Act~\citep{euaiact2024} and the voluntary NIST AI RMF~\citep{nist2023}, attach obligations or guidance to individual models, datasets, systems and evaluations. They do not model the feedback loops that undermine evaluation integrity once evaluation becomes a competitive signal. Systems-thinking~\citep{meadows2008thinking} and performativity~\citep{mackenzie2006engine} perspectives let us study how measurement and optimization co-evolve at ecosystem scale.

\subsection{AI supply chains, value chains, and the evaluation layer}
\label{app:supply-vs-value}

Two framings dominate recent analysis of AI development as a system. The \emph{supply-chain} framing~\citep{hopkins2025ai,cobbe2023supply,widder2023dislocated} treats AI production as a network of specialized actors (foundation-model developers, dataset curators, compute providers, fine-tuners, application integrators) through which models and services flow from upstream suppliers to downstream users; its concerns are accountability routing, dislocated responsibility, and the modular structure that allows harms to fall between nodes. The \emph{value-chain} framing~\citep{porter1985competitive} disaggregates a firm into value-creating activities linked to its suppliers and buyers; its concerns are cost position, competitive advantage, and where in the chain value is created. Both framings describe a linear flow, in which a product, a service or value moves from node to node.

Evaluation does not fit either framing. Benchmark scores are published artifacts that several classes of actor use in parallel, and no supplier hands them to a customer. Evaluators are also typically not paid in proportion to the value that their measurements create for downstream actors. We therefore treat evaluation as shared measurement infrastructure (\S\ref{sec:related}). Every actor in the supply-chain framing depends on this layer, and every step in the value-chain framing is priced against it. \citet{bommasani2023ecosystem} map foundation-model dependencies across the network in their Ecosystem Graphs, but they focus on the social footprint of models and not on the measurement infrastructure that coordinates behavior around them.

The failure modes that motivate this paper (score inflation, benchmark gaming, evaluator conflict of interest, and capital that follows gamed signals) do not belong to a single supply-chain node or value-chain step. They arise when the actors that a measurement layer measures also optimize against it. Appendix~\ref{app:evaluator-capture-ext} studies one case, in which the evaluator sells extra submissions and early access to the providers it scores.

\subsection{Generative agent-based modeling with LLMs: extended discussion}
\label{app:llm-agents-extended}

A central design question in generative agent-based modeling is how much behavior to delegate to LLM actors and how much to encode in rules; \citet{adornetto2025generative} advocate hybrid ABM--GABM approaches. Critical reviews argue that LLM agents make classical ABM challenges worse: their black-box nature, stochasticity and cultural biases make results harder to understand, replicate and ground empirically~\citep{larooij2025validation}, and they tend to converge toward an ``average persona''~\citep{taillandier2025integrating}. \citet{wu2025boundary} review studies that assess behavioral variance, and most of those studies report lower variance than human populations show. They propose that claims be limited to collective-level qualitative patterns when variance is insufficient. Our design follows from these concerns. LLM actors make only strategic planning decisions (investment allocation, regulatory choice, funding allocation), and rules determine market mechanics, scoring and state transitions. We also check the behavioral coherence of LLM actors (Appendix~\ref{app:validation-internal-subj}), and we make claims about patterns and not about point estimates (\S\ref{sec:results}).

A growing body of work uses LLM-powered agents as participants in social simulations, but validation methodology remains nascent. \citet{park2023generative} evaluate generative agents by interviewing the agents and having human evaluators rank their believability across component ablations. \citet{aher2023using} validate by replicating classic human-subject experiments (Ultimatum Game, Milgram) and comparing aggregate statistics against published human data, uncovering systematic ``hyper-accuracy'' distortions. \citet{gao2023s3} validate a social-network simulator at both micro (individual next-state prediction) and macro (aggregate pattern matching) levels against real social-media traces. \citet{ghaffarzadegan2024generative} use prompt-sensitivity sweeps as a robustness check for generative agent-based models, and \citet{manning2024automated} ground LLM simulations in structural causal models that enable formal hypothesis testing. \citet{cross2025validating} propose a two-stage protocol for GABM validation, which first replicates established experimental findings and then generates novel predictions. Critical reviews still identify convergence toward a generic mean that suppresses behavioral heterogeneity, and sensitivity to prompt formulation and model version~\citep{taillandier2025integrating}. Our validation protocol (Appendix~\ref{app:validation}) uses the terminating-simulation inference framework of \citet{law2015simulation} and the verification-and-validation framework of \citet{sargent2013verification}, and it adds coherence checks that are specific to LLM actors.

\subsection{Why a simulation, and what other methods offer}
\label{app:methodological-alternatives}

We use a stylized simulation as a controlled environment. \citet{ie2019recsim} take the same position for recommender systems: they build simulators that mirror specific aspects of user behavior in order to develop, evaluate and compare models, and they do not expect policies learned in simulation to be deployed in live systems. Our simulation does not predict the trajectory of real evaluation markets. It lets us test structural hypotheses about feedback loops, incentive misalignment and regulatory leverage by changing one part of the ecosystem at a time. Observational data cannot support these tests, because true capability and benchmark validity are not directly observable, and because interventions such as removing funders or the regulator have no real-world counterpart.

Other methods complement this one. A game-theoretic analysis could establish whether the behavior we observe is an equilibrium and could identify scoring rules that remove the incentive to game, but the multi-actor, multi-benchmark, multi-round structure resists closed-form solution. System dynamics makes feedback loops and leverage points explicit~\citep{meadows2008thinking}, but it models aggregate quantities and cannot represent differences between providers. Observational studies, for example of the transition from MMLU to MMLU-Pro, could test the hypotheses that the simulation generates where data exist. Expert elicitation could calibrate parameters and supply institutional knowledge that the simulation does not model.

\subsection{Governance analogues in other industries}
\label{app:governance-analogues}

The public-goods framing in \S\ref{sec:related} draws on governance regimes in other high-stakes industries. Clinical research~\citep{emanuel2000ethics,friedman2010fundamentals} provides a mature model of independent review and trial monitoring for interventions whose private incentives are misaligned with public welfare. Civil aviation offers a second model, in which mandatory occurrence reporting and the exchange of occurrence information across organisations and authorities support safety~\citep{eu2014occurrence}. Both regimes use mechanisms that our simulation can test as interventions on the AI evaluation ecosystem: independent evaluation, mandatory disclosure, liability routing and information sharing across actors. The analogies are imperfect. AI capability changes quickly, which makes pre-deployment audit a moving target, and what an evaluation should measure is more contested than a clinical endpoint or an airworthiness standard. These regimes are still useful institutional examples of what a mature AI evaluation regime could look like.

\section{Full Simulation Architecture}
\label{app:architecture}

\providecommand{\codeavailability}{Code: \url{https://github.com/aims-foundations/evarium}.}

This appendix specifies the actors, state representations and mechanics that Section~\ref{sec:simulation} summarizes. Every reported run lasts 40 rounds, and one round represents one month. Table~\ref{tab:dimensions} defines the capability dimensions, Table~\ref{tab:benchmark-pool} the benchmark suite and Table~\ref{tab:providers} the providers. Tables~\ref{tab:archetypes} and~\ref{tab:need-weights} give the consumer parameters, Table~\ref{tab:regulator-levers} the regulator levers, and Table~\ref{tab:funders} the funder parameters. \codeavailability

\subsection{Round Protocol (Pseudocode)}
\label{app:round-protocol}

\begin{figure}[!htb]
  \centering
  \includegraphics[width=\linewidth]{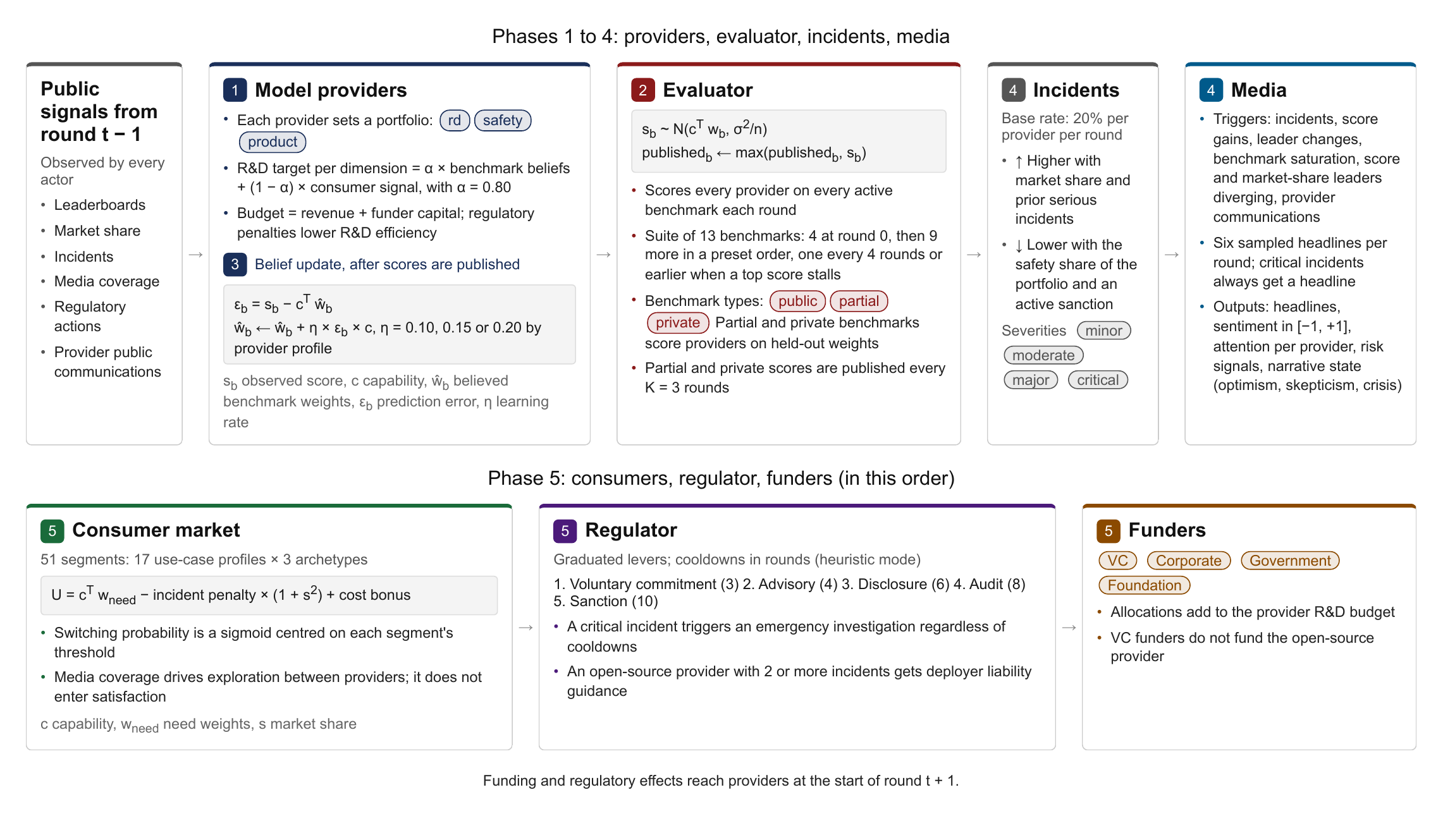}
  \caption{Order of events in one round, with the phase numbers of Algorithm~\ref{alg:round}. In Phases~1 to~4, providers plan with the public signals of the previous round, the evaluator scores them, providers update their beliefs, and incidents and media coverage follow. In Phase~5, consumers, the regulator and funders respond in that order to the round's public state.}
  \label{fig:simflow}
\end{figure}

Algorithm~\ref{alg:round} gives the order of events in one round, and Figure~\ref{fig:simflow} shows the same order as a diagram.

\begin{algorithm}[!htb]
\caption{Simulation round protocol. Actors observe only public and own-private state;
  ground truth capability vectors $\mathbf{C}$ are held by the simulation
  and never exposed to any actor prompt.}
\label{alg:round}
\small
\begin{algorithmic}[1]
\Require Providers $\mathcal{P}$, Evaluator $\mathcal{E}$, Benchmarks $\mathcal{B}$,
  ConsumerMarket $\mathcal{M}$, Regulator $\rho$, Funders $\mathcal{F}$, Media $\mu$,
  rounds $T$, ground truth $\mathbf{C}$
\Statex
\For{$t = 1, \ldots, T$}
  \Statex \hspace{\algorithmicindent}\textit{Phase 1: Provider planning \& capability update}
  \For{each provider $p \in \mathcal{P}$}
    \State $\mathbf{o}_p \gets$ \Call{Observe}{$p, \mathcal{H}_{t-1}$}
      \Comment{scores, market share, sanctions, incidents}
    \State $\mathbf{x}_p \gets$ \Call{Plan}{$\mathbf{o}_p,\, \mathbf{s}_p^{\text{priv}}$}
      \Comment{LLM or heuristic $\rightarrow$ portfolio + focus (ordinal)}
    \State $\mathbf{c}_p \gets \mathbf{c}_p + \Delta(\mathbf{x}_p,\, b_p)$
      \Comment{capability update via focus-weighted beliefs}
  \EndFor
  \Statex \hspace{\algorithmicindent}\textit{Phase 2: Evaluation \& benchmark update}
  \State $\hat{\mathbf{S}} \gets \mathcal{E}.\text{ScoreAll}(\mathcal{P},\, \mathbf{C},\, \mathcal{B})$;
    publish leaderboard $L_t$
  \Statex \hspace{\algorithmicindent}\textit{Phase 3: Provider belief update}
  \For{each provider $p \in \mathcal{P}$}
    \State Update $\hat{\mathbf{w}}_{p,b}$ from score prediction errors; reflect
  \EndFor
  \Statex \hspace{\algorithmicindent}\textit{Phase 4: Incident generation \& media coverage}
  \State $I_t \gets \text{GenerateIncidents}(\mathcal{P},\, \mathbf{C},\, \text{shares})$
  \State $m_t \gets \mu.\text{Publish}(L_t, I_t, \texttt{public\_comms})$
    \Comment{media coverage}
  \Statex \hspace{\algorithmicindent}\textit{Phase 5: Downstream actors}
  \State $\mathcal{M}.\text{Update}(L_t, \mathbf{C}, m_t, I_t)$
    \Comment{satisfaction, switching}
  \State $\rho.\text{Act}(I_t, m_t)$
    \Comment{graduated intervention}
  \State $\mathcal{F}.\text{Allocate}(L_t, m_t, \rho, I_t)$
    \Comment{capital allocation}
\EndFor
\end{algorithmic}
\end{algorithm}

\subsection{Capability Dimensions}
\label{app:dimensions}

The simulation represents model capability with six dimensions (Table~\ref{tab:dimensions}). Each dimension is an area that providers can invest in separately and that public benchmarks measure. We do not assume that the dimensions are statistically independent.

\begin{table}[!htb]
\centering
\caption{Capability dimensions, real-world anchors, and primary investment pathways.}
\label{tab:dimensions}
\small
\setlength{\tabcolsep}{3pt}
\begin{tabular}{@{}>{\raggedright\arraybackslash}p{2.3cm}>{\raggedright\arraybackslash}p{3.8cm}>{\raggedright\arraybackslash}p{2.8cm}>{\raggedright\arraybackslash}p{4.0cm}@{}}
\toprule
\textbf{Dimension} & \textbf{Represents} & \textbf{Benchmark anchor} & \textbf{Investment pathway} \\
\midrule
Reasoning & Problem solving, logic, multi-step inference & MMLU, GPQA & Reasoning post-training, CoT RLHF \\
Coding & Software engineering, debugging & HumanEval, MBPP & Code-specific data, correctness RL \\
Knowledge & Factual recall, domain expertise & MMLU subsets & Domain corpus, retrieval fine-tuning \\
Safety & Harmlessness, adversarial robustness & TruthfulQA & RLHF for harmlessness, red-teaming \\
Communication & Instruction following, fluency & MT-Bench, IFEval & Instruction-following RLHF \\
Agentic & Tool use, multi-step task execution & SWE-bench, WebArena & Trajectory RL, tool-use fine-tuning \\
\bottomrule
\end{tabular}
\end{table}

\subsection{Benchmark Pool and Introduction Schedule}
\label{app:benchmark-pool}

\begin{table}[!htb]
\centering
\caption{The default suite of 13 benchmarks in order of introduction, with each benchmark's hidden dimension weights and privacy tier. Four benchmarks are active at round~0, and the evaluator introduces the other nine one at a time (\texttt{benchmark\_introduction\_cooldown}$=4$). The dynamic evaluator mode draws from a pool of 22 benchmarks that covers the same six dimensions. The privacy tier (\texttt{public}, \texttt{partial} or \texttt{private}) determines how scores are reported (Appendix~\ref{app:private-benchmark-reporting}). The split of 8 public, 3 partial and 2 private benchmarks is a modeling choice. The partial tier is patterned on SEAL-style leaderboards and the private tier on FrontierMath-style holdouts. The columns $\cos_{\text{part.}}$ and $\cos_{\text{priv.}}$ give the realized cosine between the benchmark's public weights and its holdout weights when the benchmark is \texttt{partial} and when it is \texttt{private} (Appendix~\ref{app:private-benchmark-reporting}). The last column names the real benchmarks that each row is patterned on. The four round-0 benchmarks are patterned on MMLU~\citep{hendrycks2020measuring}, HumanEval~\citep{chen2021evaluating} and MBPP~\citep{austin2021program}, TruthfulQA~\citep{lin2021truthfulqa}, and MT-Bench and IFEval.}
\label{tab:benchmark-pool}
\small
\setlength{\tabcolsep}{3pt}
\begin{tabular}{@{}>{\raggedright\arraybackslash}p{2.7cm}lcccccccc>{\raggedright\arraybackslash}p{2.2cm}@{}}
\toprule
\textbf{Benchmark} & \textbf{Tier} & \textbf{Reas.} & \textbf{Code} & \textbf{Know.} & \textbf{Safe.} & \textbf{Comm.} & \textbf{Agent.} & $\cos_{\text{part.}}$ & $\cos_{\text{priv.}}$ & \textbf{Real analog} \\
\midrule
General Capability     & public  & 0.35 & 0.08 & 0.30 & 0.05 & 0.20 & 0.02 & 0.97 & 0.89 & MMLU \\
Coding Evaluation      & public  & 0.10 & 0.78 & 0.03 & 0.01 & 0.02 & 0.06 & 0.99 & 0.93 & HumanEval/\allowbreak MBPP \\
Safety Evaluation      & partial & 0.03 & 0.01 & 0.05 & 0.80 & 0.10 & 0.01 & 0.99 & 0.92 & TruthfulQA/BBQ \\
Instruction Following  & public  & 0.08 & 0.02 & 0.05 & 0.04 & 0.80 & 0.01 & 0.98 & 0.86 & MT-Bench/IFEval \\
Scientific Reasoning   & partial & 0.78 & 0.02 & 0.15 & 0.00 & 0.04 & 0.01 & 0.98 & 0.92 & GPQA/MMLU-Pro \\
Clinical Reasoning     & public  & 0.20 & 0.01 & 0.65 & 0.08 & 0.05 & 0.01 & 0.98 & 0.89 & Med-PaLM/MedQA \\
Adversarial Robustness & private & 0.06 & 0.01 & 0.01 & 0.85 & 0.04 & 0.03 & 0.98 & 0.88 & SEAL-Safety \\
Hard Coding            & partial & 0.10 & 0.80 & 0.02 & 0.01 & 0.01 & 0.06 & 0.99 & 0.94 & LiveCodeBench \\
Agentic Tasks          & public  & 0.25 & 0.19 & 0.01 & 0.00 & 0.07 & 0.48 & 0.98 & 0.91 & SWE-bench/BFCL \\
Advanced Math          & private & 0.85 & 0.06 & 0.05 & 0.00 & 0.03 & 0.01 & 0.99 & 0.91 & FrontierMath \\
Function Calling       & public  & 0.15 & 0.30 & 0.02 & 0.01 & 0.12 & 0.40 & 0.94 & 0.76 & BFCL \\
Long Context           & public  & 0.12 & 0.02 & 0.20 & 0.01 & 0.62 & 0.03 & 0.98 & 0.91 & RULER/\allowbreak LongBench-v2 \\
Legal Reasoning        & public  & 0.25 & 0.01 & 0.60 & 0.05 & 0.08 & 0.01 & 0.98 & 0.89 & LegalBench-Pro \\
\bottomrule
\end{tabular}
\end{table}

The population-average consumer need weights at round~0 are reasoning 0.18, coding 0.13, knowledge 0.18, safety 0.16, communication 0.27 and agentic 0.09 (Table~\ref{tab:need-weights}). The share of organizational consumers grows during a run (Appendix~\ref{app:consumers}), and by round 40 the safety weight rises to about 0.22 and the communication weight falls to about 0.20. The mean loading across the 13 benchmarks is higher than the need weight for reasoning (about 0.26 against 0.18) and for coding (about 0.18 against 0.13). The mean loading is lower than the need weight for communication at round~0 (about 0.17 against 0.27) and for safety by round 40 (about 0.15 against 0.22). The four round-0 benchmarks alone weight coding and safety more than consumers do, and they weight knowledge, agentic and reasoning less. The weight on reasoning grows as Scientific Reasoning and Advanced Math enter the suite. A provider that follows benchmark scores therefore invests in a different mix of dimensions than consumers need.

\paragraph{Introduction schedule.}
The order of the nine scheduled benchmarks is fixed (Table~\ref{tab:benchmark-pool}) and moves from broad benchmarks to specialized ones. The order and the cadence are modeling choices. The evaluator introduces the next benchmark every four rounds, at rounds 4, 8, \ldots, 36, unless the top score on an active benchmark stalls first. A top score stalls when it rises by less than 0.005 in each of three consecutive rounds. Published scores never decrease, so a stall can occur at any score level. Two rounds after a stall, the evaluator introduces the next scheduled benchmark early. At least one benchmark enters early in about two thirds of runs, and all 13 benchmarks are active by round 36 in every run. The per-benchmark results use the final round, and they are the same when we restrict the runs to those with no early introduction or to those with at least one.

In the dynamic evaluator mode, the evaluator adds one benchmark from the 22-benchmark pool every four rounds (\texttt{benchmark\_introduction\_interval}$=4$). The heuristic evaluator picks the pool benchmark that best covers the dimensions where consumer need most exceeds the coverage of the active suite, and the LLM evaluator chooses from the pool itself. When the active suite already holds 13 benchmarks (\texttt{max\_benchmarks}), adding a benchmark retires the most saturated one, and the LLM evaluator can also retire a benchmark by choice. A stalled top score does not trigger an introduction in this mode, so the timing of new benchmarks does not depend on provider scores.

\subsection{Provider Profiles}
\label{app:providers}

\begin{table}[!htb]
\centering
\caption{Initial provider profiles: capability in each dimension, and the starting portfolio split across R\&D, safety and product. Capability values range from 0.08 to 0.55, and agentic capability is low for every provider. We set the values by hand, informed by 2023 frontier-model benchmark results~\citep{stanfordaiindex2024}.}
\label{tab:providers}
\small
\setlength{\tabcolsep}{3pt}
\begin{tabular}{@{}lcccccccccp{2.6cm}@{}}
\toprule
\textbf{Provider} & \textbf{Reas.} & \textbf{Code} & \textbf{Know.} & \textbf{Safe.} & \textbf{Comm.} & \textbf{Agent.} & \textbf{R\&D} & \textbf{Safety} & \textbf{Prod.} & \textbf{Notes} \\
\midrule
Orion Labs & 0.52 & 0.48 & 0.50 & 0.42 & 0.52 & 0.12 & 55\% & 15\% & 30\% & GPT-3.5 frontier \\
Apex AI & 0.48 & 0.40 & 0.46 & 0.55 & 0.48 & 0.10 & 60\% & 30\% & 10\% & Safety lead (CAI) \\
Genesis Sys. & 0.50 & 0.38 & 0.52 & 0.40 & 0.42 & 0.12 & 70\% & 15\% & 15\% & Largest compute \\
Mirage AI & 0.44 & 0.42 & 0.44 & 0.32 & 0.40 & 0.10 & 80\% & 10\% & 10\% & Scale-first \\
Spark AI & 0.36 & 0.40 & 0.32 & 0.28 & 0.34 & 0.10 & 65\% & 10\% & 25\% & Specialization startup \\
OpenCore & 0.38 & 0.42 & 0.35 & 0.25 & 0.30 & 0.08 & 75\% & 10\% & 15\% & Open-source \\
\bottomrule
\end{tabular}
\end{table}

The providers have fictional names so that an LLM planner cannot draw on the reputation of a real company. We map them to real providers (Orion Labs $\to$ OpenAI, Apex AI $\to$ Anthropic, Genesis Systems $\to$ Google DeepMind, Mirage AI $\to$ Meta AI, Spark AI $\to$ Mistral, OpenCore $\to$ DeepSeek) only when we compare simulation output with external data, and the mapping never appears in an actor prompt.

\subsection{Provider Investment Model}
\label{app:investment}

\paragraph{Capability update rule.}
Each round, the simulation computes a provider's R\&D budget, converts it to an effective budget with two sources of diminishing returns, and then applies gains in each dimension. The budget is
\begin{align}
b_p &= \max\Bigl(\text{share}_p \cdot r \cdot M_t \cdot (1 - \text{cost\_advantage}_p) + \textstyle\sum_f \phi_{f,p},\; b_p^{\min}\Bigr), \label{eq:budget}
\end{align}
where $r = 5.0$ is \texttt{revenue\_per\_share}, $M_t$ is the total market size ($M_0 = 1$, growing 3\% per round), and $\phi_{f,p}$ is the allocation from funder $f$. The budget floor $b_p^{\min}$ is 1.0 for OpenCore and 0 for the closed providers. The effective budget is
\begin{align}
\text{effective\_budget}_p &= \sqrt{b_p} \cdot e_p \cdot \max\bigl(0,\; 1 - \text{mean}(\mathbf{c}_p)\bigr). \nonumber
\end{align}
The square root gives diminishing returns on funding, and the last factor reduces gains as mean capability approaches the ceiling of 1. The efficiency $e_p$ equals \texttt{rnd\_efficiency} ($0.1$) multiplied by any active regulatory penalty from Table~\ref{tab:regulator-levers}.

The provider splits its R\&D across dimensions with a target that blends two signals: the dimensions that its benchmark beliefs favor, and the dimensions that its consumers value.
\begin{align}
\text{target}[\text{dim}] &= \alpha \cdot \text{benchmark\_driven}[\text{dim}] + (1-\alpha) \cdot \text{effective\_signal}[\text{dim}] \label{eq:targeting}
\end{align}
The blend weight $\alpha$ is \texttt{benchmark\_orientation} (fixed at 0.80). The consumer signal is reliable only when the provider invests in product:
\begin{align}
\text{quality} &= \frac{1}{1 + e^{-3(\text{product\_budget} - 0.5)}} \label{eq:signal-quality} \\
\text{effective\_signal}[\text{dim}] &= \text{quality} \cdot \text{true\_signal}[\text{dim}] + (1 - \text{quality}) \cdot \text{uniform}[\text{dim}] \nonumber
\end{align}
Here $\text{product\_budget}$ is the provider's absolute product budget, which is the product share of the portfolio multiplied by $b_p$. The quality is capped at 0.40 for the open-source provider (Appendix~\ref{app:open-source}).

\paragraph{Safety lever.}
The safety share of the budget raises only the safety dimension. The gain is the safety budget multiplied by $(1 - c_p[\text{safety}])$, so gains shrink as safety capability rises, and by an efficiency drawn each round from $\text{Uniform}(0.3, 0.9)$.

\paragraph{Allocation smoothing.}
The simulation applies each provider's portfolio (R\&D, safety, product) as the average of its last three chosen portfolios, in both heuristic and LLM mode. The three-round window is a modeling choice that corresponds to one quarter of monthly rounds. It keeps the score-prediction error of a single round from moving the whole budget.

\paragraph{Consumer signal.}
Each provider receives a six-dimensional vector that represents what its current users value. The simulation derives the vector from the need weights of the provider's consumers, weighted by market share, and adds noise with standard deviation $\sigma = 0.05/\sqrt{\text{share}_p}$. Providers with a larger share therefore receive a more precise signal.

\paragraph{Belief update.}
Each round, a provider updates its beliefs about each benchmark's hidden dimension weights from its score prediction error:
\begin{align}
\hat{w}_{p,b}[\text{dim}] &\mathrel{+}= \eta \cdot (\hat{s}_{p,b} - \mathbf{c}_p^\top \hat{\mathbf{w}}_{p,b}) \cdot c_p[\text{dim}] \label{eq:belief}
\end{align}
After each update the simulation clips the weights at zero and renormalizes them to sum to 1. The learning rate $\eta$ depends on the provider's strategy profile: 0.20 for aggressive or competitive profiles, 0.10 for safety-oriented or responsibility-oriented profiles, and 0.15 otherwise. The simulation runs the belief update in both heuristic and LLM mode, because beliefs are state computed from observed score errors and an LLM planner does not output them.

\subsection{Heuristic Provider Planning}
\label{app:heuristic-planning}

In heuristic mode a deterministic rule set replaces the LLM planner. The rule set reads the same \texttt{ecosystem\_context} as the LLM planner (the provider's own market share and satisfaction, its recent incidents, interventions that target it, the media narrative state and its per-benchmark scores) and returns a portfolio allocation. A provider's identity comes from its initial capability vector, starting portfolio, initial focus levels, strategy profile and open-source settings (Table~\ref{tab:providers}, Appendix~\ref{app:open-source}). The rules below are the same for every provider, and no profile text such as ``safety-first'' enters them. The profile affects heuristic behavior only through the learning rate $\eta$.

\paragraph{Rules.}
Each round, the heuristic planner applies four rules to the portfolio of the previous round:

\begin{itemize}\itemsep0pt
  \item \textbf{Incident pressure.} Each of the provider's own incidents in the current round adds a shift that depends on severity (0.01 minor, 0.04 moderate, 0.08 major, 0.12 critical), and the total for the round is capped at 0.20. The shifts accumulate in a pressure value that is capped at 0.25. Each round the planner moves that amount from R\&D to safety and then multiplies the pressure by 0.40, so the safety response to an incident decays over the following rounds.
  \item \textbf{Share trend.} The planner computes the change in the provider's own share over three rounds. If the change is below $-0.02$, it moves 0.03 from R\&D to product. If the change is above $+0.02$, it moves 0.02 from product to R\&D.
  \item \textbf{Crisis narrative.} If the media narrative state is \texttt{CRISIS}, the planner moves 0.04 from R\&D to safety.
  \item \textbf{Own interventions.} If the regulator targeted the provider with $k$ interventions over the last five rounds, the planner moves $\min(0.06,\ 0.02k)$ from R\&D to safety.
\end{itemize}

\paragraph{Bounds.}
After it applies the rules, the planner clips the portfolio to $rd \in [0.10,\allowbreak 0.75]$, $safety \in [\text{safety\_floor},\allowbreak\ 0.70]$ and $product \in [0.05,\allowbreak 0.50]$, and renormalizes it to sum to 1. The safety \emph{allocation} floor is 0.10 for the open-source provider and 0.15 for closed providers, and it limits the safety share of the portfolio. A separate safety \emph{capability} floor (0.15 open-source, 0.25 closed) limits the ground-truth safety dimension (Appendix~\ref{app:open-source}).

\paragraph{Levers held fixed.}
The blend weight $\alpha$ (\texttt{benchmark\_orientation}, Eq.~\ref{eq:targeting}) is fixed at 0.80 for every provider in all reported runs, in both heuristic and LLM mode. The per-benchmark priority \texttt{focus\_level}, in $[0.1, 5.0]$, changes only in LLM mode. In heuristic mode each provider keeps its initial focus levels, which differ across providers and benchmarks, for the whole run. Heuristic results are therefore a structural baseline, and they cannot show a provider changing its priorities among benchmarks.

\paragraph{Submission count under evaluator capture.}
Appendix~\ref{app:evaluator-capture-ext} specifies the evaluator-capture condition, in which a provider pays for extra benchmark submissions. A heuristic provider spends up to 15\% of its base revenue plus its discretionary budget on extra submissions each round. The LLM planner chooses the number of submissions itself, up to 10.

\subsection{Consumer Market}
\label{app:consumers}

The market has 51 consumer segments. Each of the 17 use-case profiles (11 individual, 6 organizational) is paired with the three behavioral archetypes of its consumer type (Table~\ref{tab:archetypes}). The share of organizational consumers grows from 0.25 to 0.55 over a run along a logistic curve with its midpoint at round 18.

\paragraph{Archetypes.}

\begin{table}[!htb]
\centering
\caption{Consumer archetypes and behavioral parameters.}
\label{tab:archetypes}
\small
\begin{tabular}{@{}lcccc@{}}
\toprule
\textbf{Archetype} & \textbf{LB trust} & \textbf{Switch cost} & \textbf{Switch thresh.} & \textbf{Cost sens.} \\
\midrule
Leaderboard follower & 0.85 & 0.05 & 0.10 & 0.15 \\
Experience-driven & 0.35 & 0.08 & 0.06 & 0.30 \\
Cautious & 0.50 & 0.20 & 0.18 & 0.20 \\
Enterprise cautious & 0.25 & 0.35 & 0.20 & 0.10 \\
Enterprise growth & 0.45 & 0.25 & 0.15 & 0.15 \\
Enterprise established & 0.35 & 0.40 & 0.18 & 0.08 \\
\bottomrule
\end{tabular}
\end{table}

\paragraph{Need weights.}

\begin{table}[!htb]
\centering
\caption{Consumer need weights by use-case profile. Bold indicates the dominant dimension. Need weights are assigned to each profile from the sector-level evidence listed in Appendix~\ref{app:design-rationale}, together with occupation-level adoption rates~\citep{bick2024rapid}. Population weights are adoption-weighted shares assigned from the same adoption data, with organizational adoption surveys~\citep{stanfordaiindex2025,mckinsey2025state} as context. The Pop.\ column holds base shares that sum to 1.04; the simulation rescales them within each consumer type so that organizational segments hold 0.25 of the market at round~0, rising to 0.55 by round 40 (Appendix~\ref{app:consumers}). The Pop.\ avg row gives the resulting values at round~0.}
\label{tab:need-weights}
\small
\begin{tabular}{@{}l cccccc c@{}}
\toprule
\textbf{Profile} & \textbf{Reas.} & \textbf{Code} & \textbf{Know.} & \textbf{Safe.} & \textbf{Comm.} & \textbf{Agent.} & \textbf{Pop.} \\
\midrule
software\_dev & 0.22 & \textbf{0.55} & 0.03 & 0.02 & 0.03 & 0.15 & 0.14 \\
content\_writer & 0.10 & 0.02 & 0.20 & 0.03 & \textbf{0.63} & 0.02 & 0.08 \\
legal & 0.30 & 0.01 & \textbf{0.35} & 0.22 & 0.10 & 0.02 & 0.04 \\
healthcare & 0.12 & 0.02 & 0.28 & \textbf{0.48} & 0.08 & 0.02 & 0.05 \\
finance & \textbf{0.35} & 0.08 & 0.22 & 0.25 & 0.05 & 0.05 & 0.05 \\
educator & 0.15 & 0.02 & 0.28 & 0.08 & \textbf{0.45} & 0.02 & 0.07 \\
customer\_service & 0.05 & 0.01 & 0.12 & 0.15 & \textbf{0.50} & 0.17 & 0.08 \\
researcher & \textbf{0.30} & 0.22 & 0.28 & 0.03 & 0.07 & 0.10 & 0.06 \\
creative & 0.12 & 0.02 & 0.10 & 0.03 & \textbf{0.65} & 0.08 & 0.06 \\
marketing & 0.12 & 0.02 & 0.18 & 0.03 & \textbf{0.55} & 0.10 & 0.07 \\
service\_worker & 0.08 & 0.01 & 0.15 & 0.18 & \textbf{0.55} & 0.03 & 0.05 \\
hospital\_system & 0.12 & 0.02 & 0.25 & \textbf{0.48} & 0.10 & 0.03 & 0.05 \\
enterprise\_finance & \textbf{0.32} & 0.08 & 0.18 & 0.30 & 0.05 & 0.07 & 0.05 \\
tech\_startup & 0.18 & \textbf{0.38} & 0.05 & 0.04 & 0.05 & 0.30 & 0.06 \\
enterprise\_legal & 0.28 & 0.02 & \textbf{0.35} & 0.22 & 0.11 & 0.02 & 0.04 \\
government\_agency & 0.12 & 0.02 & 0.22 & \textbf{0.48} & 0.12 & 0.04 & 0.05 \\
enterprise\_hr & 0.20 & 0.02 & 0.20 & \textbf{0.35} & 0.18 & 0.05 & 0.04 \\
\midrule
\textbf{Pop.\ avg} & 0.18 & 0.13 & 0.18 & 0.16 & 0.27 & 0.09 & \\
\bottomrule
\end{tabular}
\end{table}

\paragraph{Satisfaction and switching.}
A segment's satisfaction with a provider depends on experienced capability, incidents and cost. It has no term for benchmark gaming or for media coverage:
\[
  \text{sat}_{s,p} = \mathbf{c}_p^\top \mathbf{w}_s - \text{incident\_penalty}_p \cdot (1 + \text{share}_p^2) + 0.15 \cdot \text{cost\_sens}_s \cdot \text{cost\_adv}_p
\]
The incident penalty sums a severity weight for each of the provider's incidents (0.01 minor, 0.05 moderate, 0.10 major, 0.20 critical), multiplied by $0.70^{\text{age}}$, so the weight of an incident halves in about two rounds. The factor $(1 + \text{share}_p^2)$ makes an incident cost a dominant provider more.

A segment's expected quality blends leaderboard scores with the quality it has experienced:
\[
  \text{expected}[p] = \text{lb\_trust} \cdot \text{score\_signal}[p] + (1 - \text{lb\_trust}) \cdot \text{running\_perceived\_quality}[p]
\]
A segment becomes more likely to switch as $\text{expected} - \text{sat}$ rises. The switching probability is a sigmoid centered on the segment's switching threshold, and the threshold is raised after a recent switch.

\paragraph{Exploration churn.}
Each round, 5\% of each provider's share in a segment enters an exploration pool. Negative media sentiment raises the rate for a provider in proportion to the attention it receives: $\text{rate} = 0.05 + \text{attention}_p \times |\text{sentiment}| \times \text{lb\_trust}_s$. Positive sentiment raises the rate for providers that receive little attention, by half as much, so their users explore toward the provider in the news. The simulation redistributes the pool across providers with a blend of experienced satisfaction and believed quality.

\subsection{Media Actor}
\label{app:media}

A single media outlet publishes each round after the evaluator scores providers and before consumers, the regulator and funders act. Coverage triggers are incidents of moderate or higher severity, large score gains, changes of leaderboard leader, benchmark saturation, and a divergence trigger. The divergence trigger fires when the score leader has differed from the market-share leader for three rounds and an incident occurs. Critical incidents always receive a headline. All other events compete for six sampled headline slots per round, and major and moderate incidents have higher sampling weights.

The outlet outputs headlines, a sentiment in $[-1, 1]$, an attention value in $[0, 1]$ for each provider, risk signals, and a narrative state (\texttt{OPTIMISM}, \texttt{SKEPTICISM} or \texttt{CRISIS}). The narrative state worsens when a cumulative count of incidents of moderate or higher severity exceeds 3 and then 8, or when the divergence trigger fires. The count decays by 10\% in each round without an incident, and the state recovers after a quiet period.

\subsection{Regulator}
\label{app:regulator}

\begin{table}[!htb]
\centering
\caption{Regulator intervention levers with real-world analogs, cooldowns, and mechanical effects. The cooldowns apply in heuristic mode; in LLM mode the code enforces only a three-round gap after any intervention.}
\label{tab:regulator-levers}
\small
\setlength{\tabcolsep}{3pt}
\begin{tabular}{@{}>{\raggedright\arraybackslash}p{2.8cm}>{\raggedright\arraybackslash}p{2.6cm}c>{\raggedright\arraybackslash}p{6.2cm}@{}}
\toprule
\textbf{Lever} & \textbf{Real analog} & \textbf{Cool.} & \textbf{Effect} \\
\midrule
Voluntary commitment & WH/Seoul pledges & 3 r & Safety \texttt{public\_comms} boost (+0.15, decaying) \\
Publish advisory & AISI summaries, NIST RMF & 4 r & Enterprise and cautious archetypes reduce LB trust ($\times$0.90, floor 0.15) \\
Safety disclosure & EU AI Act Art.\ 53 & 6 r & Safety comms boost (+0.20); \texttt{rnd\_efficiency} $\times$0.95 \\
Commission audit & AISI pre-deployment eval & 8 r & \texttt{rnd\_efficiency} $\times$0.85 for 3 rounds \\
Impose sanction & EU AI Act fines (7\%) & 10 r & \texttt{rnd\_efficiency} penalty + funder allocation $\times$0.90 \\
Emergency investigation & Cruise robotaxi shutdown & 0 r & Critical-incident override; immediate intervention regardless of cooldowns \\
Deployer liability guidance & OS-deployer policy & 0 r & Parallel track for open-source providers with $\geq 2$ incidents; assigns deployer-side liability \\
\bottomrule
\end{tabular}
\end{table}

All reported runs use one regulator with incident-rate thresholds of 0.05, 0.15 and 0.25 for its low, medium and high responses, and with advisory audits that cannot block deployment. The code also includes a stricter and a more permissive parameter set, which no reported run uses.

\subsection{Funders}
\label{app:funders}

Four funder types allocate capital across providers from fixed pools. Each type scores providers with the formula in Table~\ref{tab:funders}, where $m = 1 + 0.5 \cdot \text{sentiment}$ is a media factor.

\begin{table}[!htb]
\centering
\caption{Funder types and allocation logic.}
\label{tab:funders}
\footnotesize
\setlength{\tabcolsep}{3pt}
\begin{tabular}{@{}>{\raggedright\arraybackslash}p{1.8cm}>{\raggedright\arraybackslash}p{2.8cm}c>{\raggedright\arraybackslash}p{8.2cm}@{}}
\toprule
\textbf{Type} & \textbf{Pattern} & \textbf{Funds OS?} & \textbf{Scoring formula} \\
\midrule
VC & Concentrated & No & share\_growth $\times$ $m$ $\times$ (1 $-$ incident\_risk) $\times$ (1 $-$ share) $\times$ diversification \\
Corporate & Multi-relationship & Yes & share $\times$ $m$ $\times$ (1 $-$ incident\_risk) \\
Government & Spread, safety-oriented & Yes & (1 $-$ incident\_rate) $\times$ share\_growth \\
Foundation & Ecosystem health & Yes & (1 $-$ incident\_rate) $\times$ share\_growth \\
\bottomrule
\end{tabular}
\end{table}

Funder allocations enter the provider budget through Eq.~\ref{eq:budget}.

\subsection{Open-Source Provider Mechanics}
\label{app:open-source}

OpenCore differs from the closed providers in five ways:

\begin{itemize}
  \item VC funders do not fund it. Corporate, government and foundation funders can.
  \item Its cost advantage is 0.9, which represents open weights that users can host themselves.
  \item Its safety capability floor is 0.15, against 0.25 for closed providers.
  \item Its product signal quality is capped at 0.40, which represents limited deployment telemetry.
  \item Its open weights reveal information about benchmarks. Each round, the simulation moves every other provider's beliefs about each benchmark toward that benchmark's public weights by a fraction equal to 0.30 times OpenCore's market share.
\end{itemize}

Appendix~\ref{app:pimmur} audits these differences.

\section{Design Rationale and Parameter Sources}
\label{app:design-rationale}

This appendix gives the simulation's parameters and the reasons for its main design choices. Table~\ref{tab:sim-hyperparams} lists the hyperparameters, Table~\ref{tab:param-grounding} lists related literature, and Table~\ref{tab:assumption-status} rates each assumption behind the holdout-design result as calibrated, anchored or stipulated. Five parameters rest on external data:
\begin{itemize}[leftmargin=1.2em,itemsep=2pt,topsep=2pt]
  \item The reporting lag for private benchmarks, $K{=}3$, is calibrated to release cadence across 20 Epoch AI benchmarks and 8 frontier labs (Appendix~\ref{app:k-calibration}).
  \item The target cosines for holdout weights, $0.95$ and $0.85$, are anchored to retro-holdout score inflation~\citep{haimes2024benchmark} and to within-family Pearson correlations that we compute from Epoch AI data~\citep{epoch2024benchmarks}. The hand-authored holdout vectors realize these targets only approximately (Appendix~\ref{app:private-benchmark-reporting}).
  \item Starting capability vectors are informed by 2023 benchmark results in~\citet{stanfordaiindex2024}.
  \item Incident base rates are a modeling choice informed by incident reports in the AI Incident Database~\citep{aiincidentdatabase}.
  \item Consumer need weights are assigned to each use-case profile from sector-level evidence on what that profession uses AI for and what it is concerned about, together with occupation-level adoption rates~\citep{bick2024rapid} and worker adoption surveys~\citep{pew2025workers}: developers~\citep{stackoverflow2024survey,stackoverflow2025survey,becker2025metr}; legal~\citep{thomsonreuters2025legal,aba2024techreport,aba2024opinion512}; healthcare~\citep{ama2025physician,fda2025aidevice,ecri2025safety,soroush2024coders}; finance~\citep{fed2011sr117,fsoc2024annual,iosco2025ai,eba2025aiact}; education~\citep{gallup2025teachers}; creative work~\citep{adobe2025creators}; customer service~\citep{gartner2025service,juquelier2025empathic}; government~\citep{nist2024genai,omb2025m2521,omb2025m2522,gao2025ai,gallup2026publicsector}; and human resources~\citep{eeoc2023selection,nyc2021ll144,nycdcwp2023aedt,euaiact2024}. Population weights draw on the occupation-level adoption data, with organizational adoption surveys~\citep{stanfordaiindex2025,mckinsey2025state} as context.
\end{itemize}
Every other value is a modeling choice, and the tables say so.

\subsection{Hyperparameter summary}
\label{app:parameters}

\begin{table}[!htb]
\centering
\caption{Summary of simulation hyperparameters.}
\label{tab:sim-hyperparams}
\small
\setlength{\tabcolsep}{4pt}
\begin{tabular*}{\textwidth}{@{\extracolsep{\fill}}lr@{\qquad}lr@{}}
\toprule
\textbf{Parameter} & \textbf{Value} & \textbf{Parameter} & \textbf{Value} \\
\midrule
\multicolumn{2}{@{}l}{\textit{Ecosystem structure}} &
\multicolumn{2}{l@{}}{\textit{Provider initialization}} \\[2pt]
Providers & 6 &
Capability dimensions & 6 \\
Benchmark pool (total) & 22 (max 13 active) &
Init.\ capability (range) & $[0.08,\;0.55]$ \\
Consumer segments & 51 &
Benchmark orientation & 0.80 (fixed) \\
Funder types & 4 &
Belief learning rate & 0.10 / 0.15 / 0.20 (by profile) \\
Simulation window & 40 rounds ($\approx$3.3 yr) \\[4pt]

\multicolumn{2}{@{}l}{\textit{Scoring model}} &
\multicolumn{2}{l@{}}{\textit{Regulator}} \\[2pt]
Score formula & $\max(\hat{s}_{t-1},\,\mathcal{N}(\mathbf{c}^\top\mathbf{w}_b,\,\sigma_b^2/n_b))$ &
Threshold (low) & 0.05 \\
Benchmark weights & Hidden from all actors &
Threshold (med) & 0.15 \\
Budget scaling & $\sqrt{\text{rd\_budget}}$ &
Threshold (high) & 0.25 \\
Holdout fraction $h$ & 0.3 partial / 1.0 private &
Audits & Advisory \\[4pt]

\multicolumn{2}{@{}l}{\textit{Incident model}} &
\multicolumn{2}{l@{}}{\textit{Open-source provider}} \\[2pt]
Base incident rate & 0.20 / provider / round &
Cost advantage & 0.9 \\
Severity: min/mod/maj/crit & 50\%/31\%/12\%/7\% &
Safety capability floor & 0.15 (vs.\ 0.25 closed) \\
\bottomrule
\end{tabular*}
\end{table}

\begin{longtable}{@{}>{\raggedright\arraybackslash}p{5.0cm}>{\raggedright\arraybackslash}p{8.0cm}@{}}
\caption{Related literature for simulation parameters and design choices. Each row gives the parameter or design choice and the literature it draws on; a value not stated in a cited source is a modeling choice.}
\label{tab:param-grounding} \\
\toprule
\textbf{Parameter / Design Choice} & \textbf{Related literature} \\
\midrule
\endfirsthead
\toprule
\textbf{Parameter / Design Choice} & \textbf{Related literature} \\
\midrule
\endhead
\multicolumn{2}{@{}l}{\textit{Parameter values}} \\[2pt]
Market growth rate, 0.03/mo & \citet{spglobal2025genai} (40\% CAGR) \\
Consumer population weights, per-archetype (assumed) & \citet{bick2024rapid} (occupation-level adoption) \\
Leaderboard trust differs across users, 0.25--0.85 (assumed) & \citet{hardy2025benchmarks} \\
Incident base rates, 20\% / provider / round (assumed) & \citet{aiincidentdatabase} \\
\addlinespace
\multicolumn{2}{@{}l}{\textit{Structural design choices}} \\[2pt]
Exploration churn = media-driven & modeling choice \\
Satisfaction = experience-based & modeling choice \\
Quantified AI outputs as partial signals & \citet{elish2018situating} \\
Benchmark entrenchment & \citet{hutchinson2021accountability}; \citet{koch2021reduced} \\
Media = event-driven amplifier & \citet{fast2017longterm}; \citet{chuan2019framing}; \citet{brennen2018industry} \\
Regulator: information-constrained & modeling choice \\
VC funder: growth $>$ profitability & \citet{kenney2019unicorns} \\
Gov vs.\ corporate funder logic & modeling choice \\
Evaluator independence / access & \citet{costanzachock2022who}; \citet{raji2022outsider}; \citet{aievaluatorforum2025aef1} \\
\bottomrule
\end{longtable}

\subsection{Assumption status}
\label{app:assumption-status}

Table~\ref{tab:assumption-status} rates each assumption the holdout-design result rests on. \emph{Calibrated} means fit to external data; \emph{anchored} means the qualitative structure is supported by cited literature or data; \emph{stipulated} means a modeling choice with a rationale and no direct empirical support. The conclusions of \S\ref{sec:results} are conditional on these assumptions.

\begin{table}[!htb]
\centering
\caption{Status of the assumptions behind the holdout-design result.}
\label{tab:assumption-status}
\footnotesize
\setlength{\tabcolsep}{4pt}
\renewcommand{\arraystretch}{1.1}
\begin{tabular}{@{}>{\raggedright\arraybackslash}p{4.6cm}>{\raggedright\arraybackslash}p{2.6cm}>{\raggedright\arraybackslash}p{6.0cm}@{}}
\toprule
\textbf{Assumption} & \textbf{Status} & \textbf{Support and sensitivity} \\
\midrule
Six actor types cover the ecosystem, with heterogeneity inside each type & Stipulated & Tractability; excluded actors and levers in Appendix~\ref{app:lim-per-actor}. \\
Consumers choose through a mix of leaderboard signal and experienced quality & Anchored (trust differs across users); stipulated (trust values, mixture weights) & \citet{hardy2025benchmarks}; values in Appendix~\ref{app:consumers}. \\
Gaps are measured against the need weights of a mixed consumer population & Anchored (which needs dominate each profile, from sector evidence); stipulated (the exact values) & Sector evidence listed at the start of this appendix and \citet{bick2024rapid}; need-weights set the starting gap and its sign, so which shifts count as widening (\S\ref{sec:limitations}, item~4). \\
Six capability dimensions with uneven benchmark coverage & Anchored & \S\ref{sec:limitations}, item~2; the ontology is text-only (see below). \\
Benchmark dimension loadings (Table~\ref{tab:benchmark-pool}) & Stipulated & Author-assigned, guided by each benchmark's real analog. Every score and every gap is computed against them. Suite composition is discussed below. \\
Holdout weights & Anchored (target cosines); stipulated (directions and realized distances) & Targets anchored to retro-holdout score inflation in \citet{haimes2024benchmark} and to Epoch data~\citep{epoch2024benchmarks}; the vectors are hand-authored, their realized cosines differ across benchmarks (Appendix~\ref{app:private-benchmark-reporting}), and which named benchmarks widen depends on them (\S\ref{sec:privacy-ladder}). \\
Private scores report with lag $K{=}3$ & Calibrated & Release cadence across $20$ benchmarks and $8$ labs (Appendix~\ref{app:k-calibration}). \\
Scores are noisy linear projections of capability, reported as a running maximum & Stipulated & Linear scoring keeps the identity of Appendix~\ref{app:null-model} exact. \\
Providers infer benchmark weights from score prediction errors & Stipulated & Same rule in both planner modes; learning rate $0.10$--$0.20$ by profile. \\
Capital responds to scores and sentiment & Stipulated & Feedback concept from \citet{mackenzie2006engine,perdomo2020performative}. \\
Starting capability vectors & Anchored & 2023 benchmark results in \citet{stanfordaiindex2024}. \\
Incident base rate and severity mix & Stipulated & Informed by the AI Incident Database~\citep{aiincidentdatabase}. \\
\bottomrule
\end{tabular}
\end{table}

\paragraph{Suite composition.}
We do not fit Table~\ref{tab:benchmark-pool} against external estimates of real benchmarks' dimension loadings; producing such estimates is a separate measurement problem~\citep{hardy2026cartography}. We can describe which dimensions the suite covers. By highest-loading dimension the $13$ benchmarks divide into three on reasoning and two on each other dimension. An informal comparison against benchmark papers on arXiv suggests that this composition over-represents coding, under-represents knowledge, and omits multimodal benchmarks, which the text-only ontology excludes. The comparison concerns coverage only; the per-benchmark loading vectors remain stipulated.

\subsection{Product Investment Design}
\label{app:product-design}

Product investment works through two channels, and both depend on the provider's absolute product budget $B$. First, a larger budget makes the consumer need signal that guides R\&D more reliable: $\text{quality} = 1/(1+e^{-3(B-0.5)})$, capped at 0.40 for the open-source provider (Appendix~\ref{app:investment}). Second, a larger budget raises the effective switching cost of the provider's users by a bonus of $0.50 \times (1 - e^{-2B})$, capped at 0.15 for the open-source provider.

\subsection{Private Benchmark Score Reporting}
\label{app:private-benchmark-reporting}

\paragraph{Mechanism.}
A private or partial benchmark differs from a public one through three channels:
\begin{itemize}
  \item \textbf{Channel~1 (weight distance):} the holdout's dimension weights differ from the public weights, so R\&D that a provider aims at the public weights is partly misaligned with what the holdout rewards.
  \item \textbf{Channel~2 (reporting lag):} holdout scores are published every $K$ rounds, which delays the market's reaction to a capability improvement.
  \item \textbf{Channel~3 (observation noise):} the score is computed on the fraction $h$ of items that are held out, so the measurement noise scales as $\sigma / \sqrt{\text{samples} \times h}$, and a smaller $h$ gives a noisier score.
\end{itemize}
The published score uses the holdout weights alone, with no blending:
\[
  s_{\text{published}} = \mathcal{N}\!\bigl(\mathrm{dot}(\mathbf{c}, \mathbf{w}_{\text{holdout}}),\, (\sigma / \sqrt{n \cdot h})^2\bigr),
\]
published every $K$ rounds. Providers update their beliefs $\mathbf{\tilde w}[b]$ via a delta rule on published scores only, plus a noisy-public-weights prior at benchmark introduction ($\mathbf{\tilde w}_0[b] = \mathrm{normalize}(\mathbf{w}_{\text{public}}[b] + \mathcal{N}(0, \sigma_{\text{prior}}))$, with $\sigma_{\text{prior}}$ independent of $h$).

\paragraph{Cadence calibration.}
We set $K = 3$ from the release cadence of frontier labs in Epoch AI's benchmark archive: the median provider advances its best score on a benchmark every 2.5 to 3.3 months, depending on the capability dimension. All private benchmarks publish in the same round, every $K$ rounds. Appendix~\ref{app:k-calibration} gives the data, the method and the per-provider results.

\paragraph{What the cosine represents.}
The cosine similarity $\cos\theta$ between a benchmark's public dimension weights $\mathbf{w}_{\text{public}}$ and its holdout weights $\mathbf{w}_{\text{holdout}}$ is the simulation's single-knob abstraction for the gap between what providers can prepare for using publicly available information and what the holdout actually scores. $\cos\theta < 1$ is not a claim about any one mechanism; it absorbs at least three mechanistically distinct sources that drive a provider-observable asymmetry between the public and holdout portions of a benchmark: (i) item-level contamination and memorization of the public split during training \citep{haimes2024benchmark,xu2024contamination}; (ii) ``training on the test task'', which is legitimate use of task-structure knowledge during training that shifts effective weights toward benchmark-format-compatible skills \citep{dominguezolmedo2024training}; and (iii) adversarial holdout construction, where holdout items are sampled to test a skill mix that differs from the category label, even when items share format and surface domain with the public split. The simulation does not separately model these sources; $\cos\theta < 1$ captures their aggregate effect on provider-observable score asymmetry. The \texttt{iid\_holdout} ablation ($\cos\theta = 1$) isolates Channels~2 and~3 by removing this composite distance.

\paragraph{Empirical anchors for $\cos\theta$.}
We anchor two cosine values to the empirical literature on this composite asymmetry. $\cos\theta = 0.95$ (\texttt{partial}) corresponds to contamination-magnitude asymmetry: \citet{haimes2024benchmark} report score inflation of up to 16\,pp for some models on retro-holdouts indistinguishable in distribution from public items, which under the sim's linear scoring abstraction maps to effective cosine $\approx 0.95$--$0.97$ at typical capability magnitudes. $\cos\theta = 0.85$ (\texttt{private}) corresponds to within-family adversarial construction: same-family Pearson correlations we compute across models in Epoch AI benchmark data~\citep[snapshot retrieved 2026-04-17;][]{epoch2024benchmarks} (e.g., SWE-bench Verified vs.\ the bash-only SWE-bench leaderboard Epoch redistributed from \texttt{swebench.com} at that snapshot, FrontierMath~T1--3 vs.\ T4, ARC-AGI-1 vs.\ ARC-AGI-2) span 0.80--0.99 with a median of $\approx 0.86$, the strongest within-family empirical analogue we have for public-vs-holdout portions of the same benchmark.

\paragraph{Realized cosines.}
The simulation does not take a cosine as a parameter. Each benchmark has a hand-authored holdout weight vector. A \texttt{partial} benchmark uses that vector, and a \texttt{private} benchmark uses the vector at twice the distance from the public weights, clipped at zero and renormalized. Across the 13 benchmarks the realized cosine between public and holdout weights has a median of $0.98$ for \texttt{partial} (range $0.94$ to $0.99$) and a median of $0.91$ for \texttt{private} (range $0.76$ to $0.94$); Table~\ref{tab:benchmark-pool} lists the value for each benchmark. The realized perturbation is therefore milder than the targets for most benchmarks, and it differs across benchmarks.

\paragraph{Items-fraction role (role of $h$).}
$h$ controls the effective sample size of the holdout-only measurement: noise scales as $\sigma / \sqrt{n \cdot h}$. The \texttt{partial} type ($h = 0.3$) is therefore $\approx 1.8\times$ noisier than \texttt{private} ($h = 1.0$) for the same underlying capability vector. A smaller held-out set gives a noisier leaderboard measurement, which slows the providers' belief updates. We do not model what a provider can learn from the public $(1-h)$ fraction of the items. The evaluator-capture condition gives paying providers an information advantage through a separate early-access mechanism (Appendix~\ref{app:evaluator-capture-ext}).

\paragraph{Condition set.}
The simulation exposes the reporting lag \texttt{evaluation\_lag} ($K$) and, for each benchmark, a \texttt{holdout\_fraction} ($h$) and a holdout weight vector. Five conditions assign benchmark types across the 13-benchmark suite: \texttt{public\_only} (all public), \texttt{baseline} (8 public, 3 \texttt{partial} and 2 \texttt{private}, a modeling choice), \texttt{private\_dominant} (all \texttt{partial}), \texttt{private\_only} (all \texttt{private}), and \texttt{iid\_holdout} (holdout weights equal to public weights, which isolates the reporting lag and the noise).

\subsection{Private-Benchmark Cadence Calibration (\texorpdfstring{$K$}{K})}
\label{app:k-calibration}

This subsection gives the data, the method, and the per-dimension and per-provider results behind the default $K = 3$ (Appendix~\ref{app:private-benchmark-reporting}).

\paragraph{Data source.}
We use the Epoch AI benchmark archive, restricted to eight frontier labs that have appeared at or near the global capability frontier during 2023--2026 (OpenAI, Anthropic, Google, Meta, xAI, DeepSeek, Mistral AI, Alibaba) and to 24 benchmarks spanning knowledge, reasoning, mathematics, coding, agentic tasks, safety, and composite capability indices. Each submission is one row with a release date, a score, and an organization. Organization names are canonicalized (e.g., \textit{Google DeepMind} $\to$ \textit{Google}, \textit{Meta AI} $\to$ \textit{Meta}) so each provider contributes a single row per benchmark aggregate.

\paragraph{$K_{\text{advance}}$ definition.}
For each (provider $p$, benchmark $b$) pair, sort submissions by date and walk the running-max score trajectory. Let $\{t_1, t_2, \ldots, t_n\}$ be the dates at which $p$'s running max on $b$ strictly advanced. Then
\[
K_{\text{advance}}(p, b) \;=\; \operatorname{mean}_i\bigl(t_{i+1} - t_i\bigr) \quad \text{(months)},
\]
with median and range recorded alongside. Pairs with fewer than two advances are dropped from the aggregate (no gap to measure).

\paragraph{Principled exclusions.}
Four benchmarks are dropped from the primary analysis: Aider-Polyglot ($\approx 3.8$ rows/month), ARC-AGI-1 ($\approx 6.2$), ARC-AGI-2 ($\approx 5.6$), and LiveBench ($1.7$ rows/month but a live-updating composite by design). The first three track checkpoint variants and community re-submissions at a cadence orders of magnitude faster than flagship model releases; LiveBench measures continuous score drift rather than discrete release events. Retaining them would contaminate $K_{\text{advance}}$ with a non-release signal.

Three benchmarks are retained but truncated at the date of their last global-frontier advance: MMLU, GSM8K, and BBH. These have saturated for frontier labs; post-saturation submissions are ceiling-bound and would inflate $K_{\text{advance}}$ artificially while carrying no new release-cadence information. The pre-saturation window is retained.

\paragraph{Aggregation to analytical dimensions.}
Benchmarks are grouped into seven analytical dimensions. Five are capability dimensions of the simulation; no benchmark in the set measures Communication. The sixth is Math, which we split from Reasoning because math benchmarks saturate at different rates. The seventh is a Composite dimension for overall-capability indices.
\begin{itemize}\itemsep0pt
  \item \textbf{Knowledge:} MMLU, SimpleQA-V
  \item \textbf{Reasoning:} GPQA-Diamond, BBH, HLE, Chess-Puzzles
  \item \textbf{Math:} MATH-L5, FrontierMath-T1-3, FrontierMath-T4, GSM8K, OTIS-AIME
  \item \textbf{Coding:} SWE-Bench-Bash, SWE-Bench-Verified, WebDev-Arena
  \item \textbf{Agentic:} OS-World, AgentCo, APEX-Agents, BALROG
  \item \textbf{Safety:} Cybench
  \item \textbf{Composite:} ECI (Epoch Capabilities Index)
\end{itemize}
No benchmark in the primary set anchors the sim's Communication dimension directly; instruction-following and long-context evaluations in the sim pool are tied to the Reasoning and Knowledge cadence by analogy.

\paragraph{Per-dimension cadence.}
Table~\ref{tab:k-dim-pooled} reports pooled $K_{\text{advance}}$ per dimension, computed three ways: (i) simple mean across all (provider, benchmark) cells; (ii) $n$-weighted mean, where each cell is weighted by its number of observed advances; and (iii) median-of-medians, which is the median across providers of each provider's median $K$ within the dimension. Median-of-medians is robust to outliers from providers with few releases and is the statistic on which the calibrated default is anchored.

\begin{table}[!htb]
\centering
\caption{Pooled $K_{\text{advance}}$ (months) by analytical dimension, across the eight-lab frontier set. $n_p$ = providers with at least one advancing (provider, benchmark) cell; $n_b$ = benchmarks in the dimension; $n_c$ = cells contributing.}
\label{tab:k-dim-pooled}
\small
\begin{tabular}{@{}l rrr rrr@{}}
\toprule
\textbf{Dimension} & $n_p$ & $n_b$ & $n_c$ & \textbf{Simple mean} & \textbf{$n$-weighted mean} & \textbf{Median of medians} \\
\midrule
Knowledge  & 8 & 2 & 12 & 3.86 & 3.71 & 3.05 \\
Reasoning  & 8 & 4 & 17 & 2.95 & 2.97 & 2.47 \\
Math       & 8 & 5 & 33 & 4.27 & 3.43 & 3.03 \\
Coding     & 4 & 3 & 10 & 3.72 & 3.22 & 3.07 \\
Agentic    & 5 & 4 & 10 & 4.56 & 3.85 & 3.25 \\
Safety     & 3 & 1 &  3 & 2.97 & 3.11 & 2.50 \\
Composite  & 8 & 1 &  8 & 3.63 & 3.47 & 3.28 \\
\midrule
\textbf{Mean of dimensions} & 8 & 20 & 93 & & \textbf{3.39} & \\
\bottomrule
\end{tabular}
\end{table}

Median-of-medians $K_{\text{advance}}$ lies in $[2.47, 3.28]$ months across the seven dimensions. The mean of the seven per-dimension weighted means is $3.39$ months. We adopt $K = 3$ rounds as the default, which is consistent with every per-dimension median and within $0.4$ months of that mean.

\paragraph{Tier-1 robustness.}
Restricting to the five labs that have held the global capability frontier at some point in 2023--2026 (OpenAI, Anthropic, Google, Meta, xAI) does not change the conclusion. The mean of the seven per-dimension weighted means is $3.39$ months as well, and dimension-level median-of-medians shifts are $\leq 0.22$ months. The tier-1 median-of-medians range is $[2.47, 3.10]$, slightly tighter at the ceiling (Composite $3.28 \to 3.10$, Agentic $3.25 \to 3.03$) but preserving the distribution shape and the $K = 3$ anchor.

\paragraph{Per-provider cadence.}
Table~\ref{tab:k-provider-tier1} reports per-provider aggregate $K_{\text{advance}}$ across the tier-1 primary benchmark set. Medians cluster in $[2.57, 4.75]$ months; OpenAI and Anthropic advance fastest across the portfolio, consistent with their dense release histories. Meta's longer median partly reflects coverage, because Meta appears on only 9 of the 20 primary benchmarks.

\begin{table}[!htb]
\centering
\caption{Per-provider cadence ($K_{\text{advance}}$ in months) across the primary benchmark set, tier-1 labs. $n_b$ = benchmarks with at least one advance; $n_a$ = total advances observed.}
\label{tab:k-provider-tier1}
\small
\begin{tabular}{@{}l rr rrr@{}}
\toprule
\textbf{Provider} & $n_b$ & $n_a$ & \textbf{Mean $K$} & \textbf{Median $K$} & \textbf{SD} \\
\midrule
OpenAI     & 17 & 127 & 4.39 & 2.57 & 4.92 \\
Anthropic  & 17 & 117 & 3.08 & 2.88 & 1.14 \\
Google     & 17 &  74 & 3.86 & 3.58 & 1.43 \\
Meta       &  9 &  47 & 5.21 & 4.75 & 3.30 \\
xAI        &  8 &  23 & 2.94 & 3.03 & 0.59 \\
\bottomrule
\end{tabular}
\end{table}

\paragraph{Synchronous release in the main ablation set.}
All private benchmarks publish simultaneously every $K = 3$ rounds in the main ablation conditions. Per-provider asynchronous release processes, such as Bernoulli release events with provider-specific $\lambda_p$ matched to the per-provider medians in Table~\ref{tab:k-provider-tier1}, would preserve the ensemble cadence while adding within-round heterogeneity. The present design prioritizes cleanly identifying the reporting-lag channel; provider-level release heterogeneity is a separable sensitivity question that lies outside this paper's scope.

\subsection{Incident Probability Model}
\label{app:incidents}

The probability that a provider has an incident in a round starts from a base rate of 20\% and is scaled by the four factors in Table~\ref{tab:incident-factors}. The result is bounded between 2\% and 50\%. The safety factor uses the safety share of the provider's portfolio and not its safety capability.

\begin{table}[!htb]
\centering
\caption{Incident probability scaling factors.}
\label{tab:incident-factors}
\footnotesize
\setlength{\tabcolsep}{3pt}
\begin{tabular}{@{}>{\raggedright\arraybackslash}p{2.6cm}>{\raggedright\arraybackslash}p{5.4cm}>{\raggedright\arraybackslash}p{5.0cm}@{}}
\toprule
\textbf{Factor} & \textbf{Formula} & \textbf{Rationale} \\
\midrule
Safety investment & $1 - (\text{portfolio[safety]} \times 1.5)$ & Higher safety $\to$ lower probability \\
Market share (exposure) & $(0.3 + \text{share} \times 1.5) \times \sqrt{\text{total\_market}}$ & Larger market $\to$ higher probability \\
Incident history & $+0.02$ per prior major/critical in last $15$ rounds, cap $+0.10$ & Past harm $\to$ elevated future risk; aging window lets safety culture recover \\
Active sanction & $\times 0.75$ while sanctioned & Regulatory oversight reduces probability \\
\bottomrule
\end{tabular}
\end{table}

The 2\% floor represents risk that safety investment cannot remove, such as adversarial users, infrastructure failures and new failure modes. Severity is drawn as minor (50\%), moderate (31\%), major (12\%) or critical (7\%). The category is drawn as security breach (25\%), healthcare harm (20\%), bias or discrimination (20\%), safety failure (20\%), misinformation (10\%) or misuse (5\%).

\paragraph{Scope relative to canonical risk taxonomies.}
Our category distribution concentrates on risks with discrete, provider-attributable market-share consequences; this emphasis parallels the Safety/Failures, Privacy/Security, and Discrimination domains in the AI Risk Repository meta-review \citep{slattery2026airisk}. Two of the seven Repository domains are deliberately not modeled. \textit{Socioeconomic and environmental harms} (labor displacement, compute footprint, market concentration externalities) are system-level emergent effects without a clean provider-attribution path in a market-facing ecosystem frame; their feedback onto the provider / evaluator / consumer loop lies outside the incident architecture's scope. \textit{Human--computer interaction harms} (over-reliance, automation bias, mental-health effects) are deployment-context harms that do not attribute discretely to a provider in our incident architecture; modeling them would require a consumer-side harm channel we do not currently support. \textit{Malicious actors and misuse} is under-weighted at 5\% relative to its prevalence in the Repository (16\% of indexed risks). Raising it is a small adjustment that is unlikely to change the qualitative dynamics in \S\ref{sec:results}. These are scope choices of the incident architecture, not claims about relative real-world importance.

\subsection{Cross-Sector Reference Cases}

The simulation's incident severity weights and recovery dynamics are modeling choices, informed by cross-sector cases where critical incidents caused measurable market share shifts:

\begin{itemize}
  \item \textbf{Boeing 737 MAX} (2018--2024): the 737's share of combined 737 and A320-family deliveries fell from 48\% (580 of 1{,}206) in 2018 to 9\% (43 of 489) in 2020, the year 737 MAX deliveries resumed in December, and stood at 31\% (265 of 867) in 2024 (our computation from annual delivery releases~\citep{boeing2019deliveries,airbus2019deliveries,boeing2021deliveries,airbus2021deliveries,boeing2025deliveries,airbus2025deliveries}). Illustrates slow, partial recovery.
  \item \textbf{Avandia / rosiglitazone} (2007): after the cardiovascular risk meta-analysis of \citet{nissen2007rosiglitazone} and the 2007 FDA boxed warning, rosiglitazone's share of thiazolidinedione prescriptions in US Medicaid fell from 56\% to 19\% (Northeast) and from 49\% to 23\% (Midwest) between the 2005Q1--2007Q2 and 2007Q3--2010Q3 periods, while pioglitazone's use rose (our computation from Table~I of \citet{hsu2015thiazolidinediones}). Demonstrates competitor capture after a molecule-specific safety signal.
  \item \textbf{Cruise robotaxi} (2023--2024): after a vehicle dragged a pedestrian~\citep{doj2024cruise}, the California DMV suspended Cruise's deployment and driverless testing permits~\citep{cadmv2023cruise}; Cruise later admitted that its report to federal regulators omitted the dragging~\citep{doj2024cruise}, and GM ended funding for Cruise's robotaxi development~\citep{gm2024cruise}. Demonstrates that physical harm plus regulatory shutdown can end a market position.
  \item \textbf{Volkswagen dieselgate} (2015): Non-VW German automakers suffered an estimated reputational spillover equal to a 34.6\% reduction in annual US sales, or 23.5\% net of substitution away from VW~\citep{bachmann2023firms}. Strongest evidence for collective-reputation spillovers.
\end{itemize}

Key nuances: (1) incident type matters: physical harm plus regulatory shutdown produces permanent exits; (2) recovery is slow and fragile; (3) highly reputed firms suffer larger market penalties per incident (``liability of good reputation,'' \citealt{rhee2006liability}).

\subsection{Path Dependence in Simulated Runs}
\label{app:path-dependence}

\begin{figure}[!htb]
  \centering
  \includegraphics[width=\textwidth]{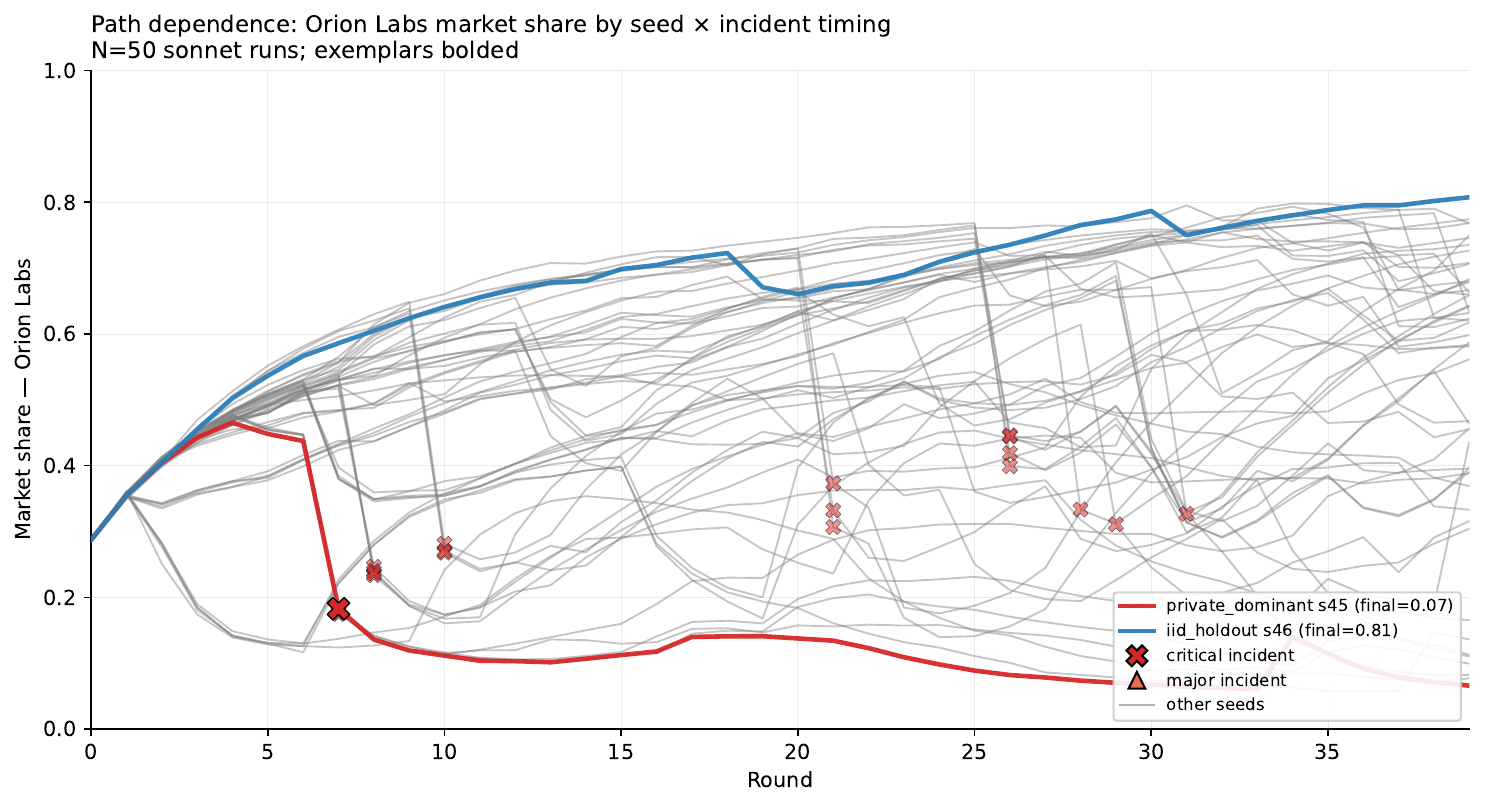}
  \caption{Market share of Orion Labs across $N{=}50$ LLM runs ($5$ privacy conditions $\times$ seeds $42$--$51$; Claude~Sonnet~4.6). Gray lines are individual runs; the two colored lines are the runs at the ends of the range (\texttt{private\_dominant}, seed 45, final share 0.07; \texttt{iid\_holdout}, seed 46, 0.81). Markers: $\times$ = critical incident, $\triangle$ = major incident.}
  \label{fig:path-dependence}
\end{figure}

Figure~\ref{fig:path-dependence} shows the market share of one provider, Orion Labs, in 50 LLM runs (5 privacy conditions and seeds 42 to 51, Claude~Sonnet~4.6). Runs with the same seed draw the same early incidents in every condition, so the 50 runs contain ten incident histories. Orion Labs' final share ranges from 0.07 to 0.81, and it ends as the market leader in 37 of the 50 runs.

The timing of the first critical incident separates the runs. In the five runs with seed 45, Orion Labs has a critical incident at round 7, before it has built a large share. Its share peaks below 0.48 and ends between 0.07 and 0.44. In the eight runs where its first critical incident comes at round 20 or later, its share peaks between 0.61 and 0.77 in the round before the incident and ends between 0.32 and 0.73. In the seven runs with no critical incident, its final share ranges from 0.39 to 0.74. With four seeds, these numbers describe the runs in the figure and do not estimate frequencies.

\textbf{Implication for aggregate claims.} A statement about the distribution of a final-state quantity (HHI, incident counts, capability levels) is a statement over paths that differ in incident timing. Analyses that pool across seeds control for incident timing only partly. Seed-matched comparisons, such as the matched-seed privacy ladder in \S\ref{sec:privacy-ladder}, control for it within each pair.

\section{LLM Prompt Templates}
\label{app:prompts}

This appendix reproduces the system prompts of the three LLM-driven actors in the reported runs (providers, funders and the regulator), along with the structure of the user message that the simulation assembles for each actor every round. The prompt text is copied from the simulation source, with dashes rendered as hyphens. The code has two further prompts that we do not reproduce: a one-line prompt that writes each provider's public announcement, and the prompt of the LLM evaluator, which only the dynamic evaluator mode uses. No prompt uses the terms ``gaming,'' ``Goodhart,'' ``exploitability,'' ``benchmark privacy,'' ``weight asymmetry'' or ``dimensional mismatch'' (Appendix~\ref{app:pimmur}).

\subsection{Provider planning}
\label{app:prompts-provider}

\paragraph{System prompt.}
The provider planning agent receives the following as its system message, with a per-provider identity block (below) appended after a delimiter.

{\scriptsize
\begin{verbatim}
You are the strategy team at an AI model company. You compete for market share by
developing AI models that are evaluated on public benchmarks and used by real customers.

Benchmarks are reported in three types - public (scored each round on published weights),
partial (K=3-round reporting lag, holdout-only weights with h=0.3 of items held out),
private (K=3-round lag, holdout-only weights with h=1.0) - each is labeled in your
evaluation results.

Each month you allocate your budget across three areas and set research priorities:
- Research & Development: advancing your model's core capabilities
- Safety & Alignment: ensuring your model is safe, honest, and robust
- Product & Go-to-Market: pricing, deployment, partnerships, user experience

You also decide which evaluation areas to prioritize in your R&D pipeline. Your R&D
capacity is finite - prioritizing one benchmark means less attention on others. Focus on
the benchmarks that are most important to your goals and deprioritize those that matter
less.

For each area, state how you want to adjust relative to current levels. Use a 5-point
scale: "much_more", "more", "same", "less", "much_less".

You MUST output valid JSON in this exact structure. Fill in all fields before writing
the reasoning - this ensures your decisions are captured even if the response is long.
{
    "portfolio": {"rd": "<signal>", "safety": "<signal>", "product": "<signal>"},
    "benchmark_focus": {"<benchmark_name>": "<signal>", ...},
    "strategy_memo": "1-2 sentence summary of your decision and rationale, written for
                      your future self to read next month.",
    "reasoning": "Your analysis in up to 250 words. Do NOT restate the leaderboard,
                  user-research percentages, or current portfolio numbers - those are
                  already provided above."
}
Where <signal> is one of: "much_more", "more", "same", "less", "much_less".
\end{verbatim}
}

\paragraph{Identity block (appended).}
The code appends the following identity block to the system prompt (separator \texttt{---}), so that the company's identity stays fixed across rounds:

{\scriptsize
\begin{verbatim}
---
## Your Company: <name>
Known for: <strategy_profile>
Traits: <innate_traits>
\end{verbatim}
}

The six provider descriptions are patterned on the public positioning of real labs and contain no instruction about how to behave:

{\footnotesize
\begin{description}
\item[Orion Labs.]
\textit{Known for:} Frontier AI lab pursuing AGI with a mandate to ensure benefits are broadly distributed. Operating as a public benefit corporation with substantial investor capital.
\textit{Traits:} frontier-focused, AGI-oriented, broad-benefit-mandate, capital-intensive.

\item[Apex AI.]
\textit{Known for:} Frontier AI lab with a safety-research thesis: building frontier models is necessary because safety challenges emerge at scale. Race-to-the-top positioning: demonstrating safety-first frontier labs can be commercially viable. Operating as a public benefit corporation with substantial investor capital.
\textit{Traits:} research-first, enterprise-focus, safety-research-thesis, race-to-top-positioning, capital-intensive.

\item[Genesis Systems.]
\textit{Known for:} Science-led AI research lab pursuing responsible AI to benefit humanity and solve fundamental scientific challenges. Operates within a large technology company with massive compute infrastructure. Responsibility framed as integral to the scientific method, not supplementary.
\textit{Traits:} science-led, massive-infrastructure, scientifically-rigorous, parent-company-embedded, responsibility-as-method.

\item[Mirage AI.]
\textit{Known for:} AI lab within a large-platform technology company. Mission framed around making AI capabilities broadly available rather than centralized. Product integration across an existing massive user base. Capital expenditure scales with parent-company commitments.
\textit{Traits:} platform-embedded, broad-distribution-oriented, massive-user-base, capex-intensive, decentralization-thesis.

\item[OpenCore.]
\textit{Known for:} Research-focused AI lab pursuing AGI with open-source release as core strategy. Publishes weights and technical details to build research community adoption. Efficient training and compute use are structural priorities. Operates under different capital and regulatory conditions than Western closed-source labs.
\textit{Traits:} open-source-first, research-oriented, compute-efficient, community-adoption, non-standard-regulatory-context.

\item[Spark AI.]
\textit{Known for:} Venture-funded AI startup with a small team and limited compute relative to hyperscalers. Growth strategy is specialization rather than broad competition. Runway and fundraising cadence are recurring constraints on strategic decisions.
\textit{Traits:} venture-funded, resource-constrained, specialization-strategy, runway-sensitive, developer-focused.
\end{description}
}

\paragraph{User-message schema.}
The per-round user message is assembled from live state. Section order (conditional sections in italics):

{\footnotesize
\begin{itemize}
  \itemsep 0.1em
  \item \textbf{Month header}: ``Month $r$ Strategy Review - \emph{<name>}''.
  \item \textbf{\textit{Recent Industry News}}: exogenous-event narrative when active, dated by month.
  \item \textbf{\textit{What You Committed To In Recent Months}}: the \texttt{strategy\_memo} field from up to the last two months, or the \texttt{reasoning} field when the memo is empty (a memory channel across rounds).
  \item \textbf{Evaluation Results}: per-benchmark table with type tag (public / partial / private), latest score, 1-round delta, and the provider's own current priority label. Type-legend is repeated inline.
  \item \textbf{\textit{Competitor Activity (This Month)}}: competitor announcements when present, and a ``Where scores moved most this month'' table that lists the three largest score changes on the benchmarks the provider prioritizes and the three largest on the others.
  \item \textbf{\textit{Industry News (Last 2 Months)}}: raw media headlines; sentiment labels are not shown.
  \item \textbf{\textit{Recent Safety Incidents}}: up to three of the provider's own incidents from the previous month (severity, category).
  \item \textbf{Current State}: market share with the change this month; budget allocation (R\&D, safety and product fractions); funding received this month, with the number and kinds of funder types that allocated; cumulative funding to date.
  \item \textbf{\textit{What Your Team Thinks Each Evaluation Tests}}: inferred per-benchmark dimension priorities for new benchmarks and benchmarks whose top-dimension belief shifted materially this round; stable beliefs are suppressed and remain implicit in prior memos.
  \item \textbf{\textit{User Research}}: top three user-need dimensions as an ordered list, without percentages or qualifiers; in the consumer-signal-hidden variant, replaced by a ``Business Metrics'' retention-trend signal computed from market-share history.
  \item \textbf{Decision}: JSON-shape reminder.
\end{itemize}
}

The message enforces hidden information through what it leaves out. Providers do not see other providers' portfolios, the regulator's lever ladder or cooldown state, the funders' reasons for their allocations, or any benchmark dimension weights; Appendix~\ref{app:architecture} lists the provider-private and ground-truth quantities. The provider message also has no section on regulator interventions, so an LLM provider learns of an intervention only through its effects.

\subsection{Funder planning}
\label{app:prompts-funder}

\paragraph{System prompt.}
{\scriptsize
\begin{verbatim}
You are a capital allocator deciding how to distribute funding across AI model
companies this month.

You MUST output valid JSON in this exact structure:
{
    "allocations": {
        "ProviderName1": <amount in dollars>,
        "ProviderName2": <amount in dollars>,
        ...
    },
    "reasoning": "Your analysis (up to 200 words)."
}
The "reasoning" field MUST contain your actual analysis - never leave it as "..." or
a placeholder.
You may fund any subset of providers and hold capital in reserve; allocations must sum
to at most your total available capital.
\end{verbatim}
}

\paragraph{Identity block (appended).}
A per-fund identity block is concatenated (separator \texttt{---}):

{\scriptsize
\begin{verbatim}
---
## Your Fund: <name>
<FUNDER_TYPE_BLOCK>
Your mission: <mission_statement>
\end{verbatim}
}

The four funder-type blocks:

{\footnotesize
\begin{description}
\item[vc.] You are a venture capital firm. You invest equity in AI companies to earn returns; you cannot fund open-source providers (no equity to take). Selective, high-conviction allocation is common.
\item[corporate.] You are a corporate strategic investor. You maintain relationships across multiple AI providers for commercial alignment and optionality, not pure financial return.
\item[gov.] You are a government funder with a public mandate. You prioritize safety, broad ecosystem health, and avoiding excessive market concentration.
\item[foundation.] You are a mission-driven foundation. You allocate for long-term research and societal benefit, often supporting underdogs and public-good capabilities.
\end{description}
}

\paragraph{User-message schema.}

{\footnotesize
\begin{itemize}
  \itemsep 0.1em
  \item \textbf{Header}: fund name and ``Capital to deploy this month: \$$X$'' (the round-level budget cap).
  \item \textbf{\textit{Recent Industry News}}: exogenous-event narrative when active.
  \item \textbf{\textit{Your Notes From Prior Months}}: the funder's own \texttt{reasoning} field from the last two months, truncated to 250 characters per entry (a memory channel across rounds).
  \item \textbf{Current Leaderboard}: per-provider rank, score, 1-month and 2-month score deltas, market share.
  \item \textbf{\textit{Recent Safety Incidents}}: per-provider incident entries (severity, month) from the last 2 incidents per provider.
  \item \textbf{\textit{Regulatory Context}}: the last four regulator interventions and the list of active regulations.
  \item \textbf{\textit{Other Funders This Month}}: the top 3 backers per provider this month with allocation amounts (so that a funder sees what other funders did).
  \item \textbf{\textit{Recent Press Coverage}}: raw media headlines, last 4; sentiment labels are not surfaced.
  \item \textbf{\textit{Provider Announcements}}: the last 6 announcements (provider, one-line summary).
  \item \textbf{\textit{Your Funding Portfolio}}: cumulative allocations by provider, with the last-round amount and month.
  \item \textbf{Decision}: JSON-shape reminder, cap-at-capital constraint, plain-integer dollar format.
\end{itemize}
}

The funder message excludes: provider portfolios, internal benchmark beliefs, regulator cooldown state, peer-funder reasoning, and any benchmark dimension weights.

\subsection{Regulator}
\label{app:prompts-regulator}

\paragraph{System prompt.}
{\scriptsize
\begin{verbatim}
You are a regulatory body overseeing the AI model provider market.

Available interventions you can take each month:
- "none" - No action this month
- "request_voluntary_commitment" - Ask providers for public safety pledges
- "publish_advisory" - Publish an evaluation summary naming specific concerns
- "mandate_safety_disclosure" - Require providers to publish system cards or
                                safety reports
- "commission_audit" - Order a pre-deployment safety evaluation of a provider
- "impose_sanction" - Impose a financial penalty on a provider
- "emergency_investigation" - Launch an immediate investigation

Name a specific target provider only when your concern is provider-specific;
otherwise use null.

You MUST output ONLY valid JSON in this exact structure:
{"intervention_type": "none"|"request_voluntary_commitment"|"publish_advisory"
                    |"mandate_safety_disclosure"|"commission_audit"
                    |"impose_sanction"|"emergency_investigation",
 "target_provider": "provider name or null",
 "reasoning": "2-4 sentence explanation"}
\end{verbatim}
}

The regulator's system prompt is the same in every run; its incident-rate thresholds are set outside the prompt (Appendix~\ref{app:regulator}). In heuristic mode the regulator enforces a cooldown for each lever (voluntary 3, advisory 4, disclosure 6, audit 8, sanction 10, emergency 0 months). In LLM mode the code enforces only a gap of three months after any intervention.

\paragraph{User-message schema.}

{\footnotesize
\begin{itemize}
  \itemsep 0.1em
  \item \textbf{Header}: ``It is month $r$.''
  \item \textbf{\textit{Recent Industry News}}: exogenous-event narrative when active.
  \item \textbf{Policy objectives and mandate}: the regulator's objectives and one sentence that states whether its audits are binding; in all reported runs they are advisory.
  \item \textbf{Current Leaderboard}: published scores of all providers.
  \item \textbf{Market Shares}: current share of each provider.
  \item \textbf{Incident Frequency by Provider}: incident counts for each provider over the last 10 months.
  \item \textbf{Most Recent Incidents}: the three most recent incidents.
  \item \textbf{Media Coverage}: the last four raw headlines, with no sentiment or narrative label.
  \item \textbf{\textit{Recent Provider Announcements}}: each provider's latest public announcement.
  \item \textbf{Prior Interventions}: the regulator's own earlier interventions.
  \item \textbf{\textit{Your Assessment From Prior Months}}: the regulator's last three rationales, truncated to 250 characters each (a memory channel across rounds).
\end{itemize}
}

The regulator message excludes consumer satisfaction, capability vectors, provider portfolios, market-concentration statistics and any benchmark dimension weights.

\subsection{Apparatus-vocabulary audit}
\label{app:prompts-apparatus}

None of the prompts above reference the simulation apparatus (``gaming,'' ``Goodhart,'' ``exploitability,'' ``benchmark privacy,'' ``weight asymmetry,'' ``dimensional mismatch''). Benchmark types are labeled neutrally (\texttt{public}, \texttt{partial}, \texttt{private}) with their mechanical semantics stated factually (reporting lag, holdout-only scoring with sample fraction $h$, matching the scoring mechanism in Appendix~\ref{app:private-benchmark-reporting}) and no framing as a policy lever under study. These choices support the PIMMUR Minimal-Control and Unawareness assumptions documented in Appendix~\ref{app:pimmur}.

\section{Generative ABM Validation}
\label{app:validation}

This appendix presents instrument-level validation for the simulation as a whole. Recent surveys of generative agent-based modeling report that validation methodology for GABM remains an open problem \citep{adornetto2025generative}. We set the ceiling on what we claim with the hierarchy of evidence that \citet{vezhnevets2023concordia} adapt for GABM from evidence-based medicine~\citep{higgins2008cochrane}. Our evidence sits on its lower rungs, which cover theory consistency together with the model-comparison and robustness practices the hierarchy recommends. It stays below the hierarchy's gold standard of prediction on new real-world data, because no real-world counterfactual ecosystem exists. We organize the protocol along the four V\&V activities of \citet{sargent2013verification} (data, conceptual, verification, operational), and \citet{law2015simulation}'s terminating-simulation framework supplies the inference machinery. Within operational validation the system is non-observable in Sargent's sense, so our evidence concentrates in the \emph{explore behavior} and \emph{comparison to other models} cells, and the only statistical comparison is against a fitted null. We add the PIMMUR audit \citep{zhou2026pimmur}, which covers failure modes specific to LLM multi-agent simulations: demand characteristics, goal-injecting prompts and collapse into consensus when agents share one model. Agreement across the cross-mode, cross-model, structural-ablation and null-model checks takes the place of statistical fit on a single output. We take four specific techniques from earlier LLM-ABM work: prompt-sensitivity diagnostics following \citet{ghaffarzadegan2024generative}, structural-causal grounding following \citet{manning2024automated}, multi-run robustness and sensitivity checks following \citet{larooij2025validation}, and variance reporting following \citet{wu2025boundary}.

\paragraph{Acceptable range of accuracy.}
Following Sargent's protocol, we state the accuracy criterion the simulation is held to: the sign of each effect and the rank ordering along the privacy ladder. We do not hold it to point-prediction error on $\Delta g$ magnitudes. The magnitudes it produces are properties of the parameterization, and we do not present them as predictions about real markets.

\paragraph{Replication protocol.}
We run $N$ independent replications per condition, which give independent copies of every outcome trajectory. Paired conditions share the same seeds, so the simulation's internal random draws are common to both and paired differences carry less variance. The LLM's server-side sampling is not controlled this way, which limits the benefit in the behavioral layer. Heuristic distributional claims rest on $N{=}50$ seeds per condition. LLM claims rest on paired-seed comparisons at $N{=}10$ Sonnet seeds, with $N{=}3$ Opus and GPT-5.5 seeds for cross-vendor coverage.

\subsection{Data validity}
\label{app:validation-data}

Appendix~\ref{app:design-rationale} lists the simulation's parameters and the source behind each one. Five of them rest on external data: the reporting lag $K{=}3$ for private benchmarks, the target cosines for holdout weights, the starting capability vectors, the incident base rate and the consumer need weights. Each enters the operational layer through a structural mechanism, and we do not tune any of them to produce a result. Where no calibration data was available, Appendix~\ref{app:design-rationale} records the value as a modeling choice and gives the reasoning for it. Cross-condition comparisons hold these values constant, so uncertainty in peripheral parameters does not flow into the condition-effect estimates.

\subsection{Conceptual model validation}
\label{app:validation-conceptual}

\paragraph{Face validity on structural design.}
Several actor design choices follow documented real-world incentive structures. We model media attention as an exploration channel that drives consumer churn, and it does not enter satisfaction directly. The choice is consistent with credence goods, whose quality buyers cannot verify even after use \citep{dulleck2006credence}. Funder allocation logic gives VC, corporate, government and foundation funders their own cooldowns and capital-availability curves, informed by the cadence of AI funding events between 2020 and 2026 (Appendix~\ref{app:parameters}). The regulator's graduated escalation ladder (advisory $\to$ disclosure $\to$ audit $\to$ sanction) follows the disclosure-then-audit-then-sanction sequencing in EU AI Act enforcement design and in US executive-order rulemaking. We do not check whether practitioners recognize their own incentives in the simulated portrayal, and the paper makes no claim that rests on such a check.

\paragraph{Pattern-oriented reproduction.}
\citet{grimm2005pattern} require a model to reproduce several qualitative patterns at once, which is a stronger evidence bar than single-pattern calibration. We sort the patterns the simulation exhibits by how they arise. Three are design targets, present because the architecture is built to permit them: \emph{benchmark saturation} (dynamic-evaluator retirements and introductions in response to ceiling effects, Appendix~\ref{app:face-validity-ext}); \emph{score--satisfaction divergence} on dimensions where R\&D over-invests (\S\ref{sec:results}, Figure~\ref{fig:privacy_ladder}); and \emph{lasting share loss after a critical incident}, since the cross-sector cases of Boeing, Avandia, and Cruise informed the incident recovery dynamics (Appendix~\ref{app:incidents}). Two are emergent, in that no rule specifies the outcome: \emph{market concentration} toward a single leader and \emph{incident-driven path dependence}, where the same provider ends between about $7\%$ and $74\%$ share across seeds depending on the timing and severity of early critical incidents (Appendix~\ref{app:path-dependence}). We claim no external referent for the emergent patterns. Their co-occurrence in the same runs is qualitative support for the architecture; it does not stand in for statistical fit on any single output.

\subsection{Computerized model verification}
\label{app:validation-verification}

\paragraph{Static verification: three-tier visibility partition.}
The simulation partitions agent state into \texttt{PublicState} (visible to all actors), \texttt{PrivateState} (visible to its owner) and a \texttt{GroundTruth} dictionary (Appendix~\ref{app:architecture}). The code path enforces the partition, because the simulation harness holds \texttt{GroundTruth} and no prompt builder reads it, so its contents never reach an actor's prompt. The architecture therefore rules out leakage across tiers, and a per-run audit is not needed to establish it.

\paragraph{Dynamic verification: per-run instrumentation.}
The simulation writes each LLM-driven actor's per-round reasoning trace to \texttt{rounds.jsonl}, which allows qualitative inspection. Paired traces under matched conditions appear in Appendix~\ref{app:privacy-traces}, and per-frame keyword tagging supports the cross-model frame analysis in Appendix~\ref{app:model-robust-frames}. An LLM call whose response fails JSON validation is retried with the same prompt, up to three attempts in total. When all three attempts fail, the actor takes a fail-safe action for that round: a provider keeps the portfolio it chose in the previous round, a funder splits its allocation evenly, and the regulator falls back to its heuristic plan. The simulation prints a line for every call that reaches the fail-safe (Appendix~\ref{app:llm-caveats}).

\subsection{PIMMUR audit}
\label{app:pimmur}

We audit the simulation design against the PIMMUR principles for LLM multi-agent simulations~\citep{zhou2026pimmur}, which identify six common validity threats: Profile (diversity of agent backgrounds), Interaction (communication structure), Memory (persistent state), Minimal-Control (avoiding demand characteristics), Unawareness (agents ignorant of hypothesis), and Realism (empirical grounding). Profile, Interaction, and Realism extend the conceptual-validation activity above; Memory, Minimal-Control, and Unawareness extend verification.

\begin{itemize}
  \item \textbf{Profile:} Each provider carries a distinct \texttt{strategy\_profile} narrative and \texttt{innate\_traits} string anchored to the system prompt as a persistent identity layer (Appendix~\ref{app:prompts-provider}); the six default providers (Orion / Apex / Genesis / Mirage / Spark / OpenCore) span well-funded incumbents, safety-focused research labs, platform-subsidiary labs, scrappy startups, and an open-weight player, with distinct portfolio tilts, brand recognition, cost advantages, and capability vectors (Appendix~\ref{app:parameters}). Funders span four archetypes (VC, corporate, government, foundation), each with its own identity block and scoring rule (Appendix~\ref{app:prompts-funder}). The regulator has three parameter sets, which differ in their incident thresholds and in whether an audit binds; every reported run uses the balanced set, with thresholds of $0.05$, $0.15$ and $0.25$ and advisory audits (Appendix~\ref{app:regulator}). In heuristic mode, provider differentiation enters through a per-provider belief-update learning rate ($0.10$ to $0.20$, read from profile and trait keywords such as ``aggressive'' or ``safety''), so heterogeneity survives when the LLM planner is removed (Appendix~\ref{app:heuristic-planning}). The cross-model comparison over three LLM families (Appendices~\ref{app:model-robust-frames} and~\ref{app:model-robust}) checks that the ladder ordering does not depend on which LLM drives the actors.
  \item \textbf{Interaction:} What each actor reads from \texttt{PublicState} and from media output defines the communication graph. Providers see the month's score movements, recent media headlines, competitor public communications, and which funder types allocated capital. Provider prompts carry no section on regulator interventions, so an LLM provider learns of an intervention only through its effects (Appendix~\ref{app:prompts-provider}). The regulator reads the leaderboard, market shares, per-provider incident counts over a ten-round window, raw media headlines and provider announcements. Funders read the leaderboard with score deltas, market shares, peer-funder allocations by provider, regulator interventions and active regulations. The media actor observes the leaderboard, benchmark publications, incidents, regulator interventions, funder allocations and provider announcements, and it emits headlines plus a per-provider attention value that drives consumer exploration churn. Round order is fixed (providers~$\rightarrow$~evaluator~$\rightarrow$~consumers~$\rightarrow$~regulator~$\rightarrow$~funders), so influence between actors flows through the public artifacts of the same round and through no hidden side channel (Appendix~\ref{app:architecture}).
  \item \textbf{Memory:} Agent state is per-actor and partitioned into \texttt{PublicState} (visible to all), \texttt{PrivateState} (self-only), and a simulation-held \texttt{GroundTruth} dict that is structurally inaccessible from any prompt builder (Appendix~\ref{app:architecture}). Each LLM-driven actor carries its own truncated cross-round reasoning trace and memory list; nothing routes between actors except through \texttt{PublicState} and media coverage. Incidents are logged with per-provider attribution alongside severity-weighted ecosystem penalties, so analysis can decompose raw counts, severity distributions, and penalty effects separately rather than collapsing to a single ecosystem scalar.
  \item \textbf{Minimal-Control:} Apparatus vocabulary (``gaming'', ``Goodhart'', ``exploitability'') appears in no actor prompt. Provider prompts describe the three benchmark types in mechanical terms and use the word ``holdout'' to name the scoring weights: a public benchmark is scored every round on published weights, a partial benchmark carries a $K{=}3$ reporting lag and holdout-only weights with $h{=}0.3$ of items held out, and a private benchmark carries the same lag with $h{=}1.0$. The prompts attach no evaluative language to the three types. Static schema and intervention menus live in the system prompts, so no round re-injects them, which reduces the per-round priming surface.
  \item \textbf{Unawareness:} Provider names are fictional anonymizations (Appendix~\ref{app:architecture}), and provider prompts address ``the strategy team at \{name\}'' with no reference to the simulation's causal diagrams, hypotheses, or evaluation-validity framing. Combined with the apparatus-vocabulary exclusion above, prompts present each actor with a within-fiction strategic problem rather than an experimental setup, so any benchmark-gaming structure the LLM recognizes from training data must be inferred from observable dynamics rather than cued by the prompt. This addresses \citet{vezhnevets2023concordia}'s train-test contamination concern at the prompt layer.
  \item \textbf{Realism:} Several structural choices ground agent dynamics in observed real-world behavior: (i) the media narrative state (OPTIMISM / SKEPTICISM / CRISIS) is a state machine with multi-round inertia and a 10\%/round pressure-decay term, matching the persistence of real news cycles rather than flipping instantaneously on a single incident; (ii) consumer exploration churn responds to high-attention media coverage on both signs of sentiment (negative coverage drives exits from the covered provider; positive hype draws explorers from non-covered providers at half-strength), capturing hype-driven adoption inflow alongside scandal-driven departure; (iii) funder capital availability follows a geometric $7\%$ per month growth curve normalized over the simulation window, and each funder type has its own decision cooldown (VC~4, corporate~7, government~10, foundation~6 months), which are modeling choices informed by the cadence of AI funding events between 2020 and 2026 (Appendix~\ref{app:parameters}); (iv) the regulator escalates through a graduated lever ladder that runs from a voluntary commitment through an advisory, a disclosure requirement, an audit and a sanction, with severity thresholds at each step (Table~\ref{tab:regulator-levers}). The heuristic regulator applies a separate cooldown to each lever; on the LLM path the code enforces only a three-round gap after any intervention. Consumer behavioral coefficients are constructed by hand, and we record them as a calibration limitation.
\end{itemize}

\subsection{Operational validation}
\label{app:validation-operational}

\paragraph{Approach.}
The AI evaluation ecosystem under counterfactual privacy regimes is non-observable in Sargent's sense, because we cannot observe the real ecosystem under a counterfactual in which, for example, all benchmarks were private. All operational evidence therefore concentrates in the right column of Sargent's Table~1, which holds \emph{explore model behavior} and \emph{comparison to other models}, and the only statistical comparison is against a fitted null. Table~\ref{tab:sargent-op-validity} maps our techniques to Sargent's classification.

\begin{table}[!htb]
\centering
\caption{Operational-validity techniques used in this paper, mapped to the operational-validity classification of \citet{sargent2013verification}. Cells marked None hold no technique, because real ecosystem trajectories under counterfactual privacy regimes are not observable, so all evidence sits in the right column.}
\label{tab:sargent-op-validity}
\footnotesize
\setlength{\tabcolsep}{4pt}
\renewcommand{\arraystretch}{1.15}
\begin{tabular}{p{2.4cm} p{4.0cm} p{6.0cm}}
\toprule
 & \textbf{Observable system} & \textbf{Non-observable system} \\
\midrule
\textbf{Subjective approach} & None (no observable counterfactual ecosystem) &
\emph{Explore model behavior:} exogenous shock (Appendix~\ref{app:face-validity-ext}); an ablation that removes incidents (Appendix~\ref{app:ablation-sweep}). \newline
\emph{Comparison to other models:} cross-mode heuristic\,$\leftrightarrow$\,LLM (Appendix~\ref{app:privacy-ladder-ext}); cross-model Sonnet/Opus/GPT-5.5 (Appendices~\ref{app:model-robust-frames} and~\ref{app:model-robust}). \\
\addlinespace
\textbf{Objective approach} & None (not pursued) &
\emph{Comparison to other models, statistical:} fitted $\Delta g$ null model (Appendix~\ref{app:null-model}). \\
\bottomrule
\end{tabular}
\end{table}

\paragraph{Subjective $\times$ Non-observable: explore model behavior.}
The exploration probes test whether the simulation behaves as expected under designed perturbations. First, the exogenous-shock check at a single seed (Appendix~\ref{app:face-validity-ext}) injects a capability-release shock into the LLM prompts at round~$25$, and the injected framing appears in the providers' reasoning traces in the following rounds. The providers' portfolio response differs across re-runs, so we draw no conclusion from it. Second, an ablation removes the incident stream (Appendix~\ref{app:ablation-sweep}), and the market concentrates further on the leader in every run. The other structural ablations we ran gave no stable direction, and we draw no conclusion from them.

\paragraph{Subjective $\times$ Non-observable: comparison to other models.}
The operational layer uses two comparison axes. The \emph{cross-mode comparison} sets the heuristic planners against the LLM planners (Appendix~\ref{app:privacy-ladder-ext}). Heuristic planners cannot read the \texttt{public}, \texttt{partial} or \texttt{private} label, so agreement across the two modes on the direction of the privacy ladder, on $12$ of $13$ benchmarks, is evidence that the structural mechanism drives the result without LLM-specific reasoning. The \emph{cross-model comparison} sets Sonnet, Opus and GPT-5.5 against each other at matched seeds, with $N{=}3$ per condition for Opus and GPT-5.5 (Appendices~\ref{app:model-robust-frames} and~\ref{app:model-robust}). The privacy-condition ordering replicates across all three planners at the outcome level. At this seed count the replication does not rule out the variance underrepresentation that \citet{wu2025boundary} document, nor the model-version sensitivity that \citet{taillandier2025integrating} note. The three planners differ in how often they reason about the reporting lag explicitly, and Appendix~\ref{app:model-robust-frames} reports those rates.

\paragraph{Objective $\times$ Non-observable: comparison to other models with statistical tests.}
The closed-form identity $\Delta g = \mathbf{c} \cdot (\mathbf{w}_{\mathrm{holdout}} - \mathbf{w}_{\mathrm{public}})$ supplies a null model that predicts $\Delta g$ from round-$0$ capability and portfolio through a growth model fit on simulation runs. Following the fitted-model prediction approach of \citet{manning2024automated}, we fit Ridge regressions per (provider, dimension) on initial conditions and propagate their predictions through the identity. Across the cross-sequence LLM dataset ($n{=}208$ paired-prediction cells) the null recovers $97.4\%$ of $\Delta g$ variance, and the simulation's $40$-round trajectory adds $+0.2$ pp $r^2$. A held-out replication under the \texttt{initial\_uniform\_capability} ablation tightens the test, and the null still predicts at $r^2{=}0.955$ with the simulation's contribution at $+2.0$ pp. Appendix~\ref{app:null-model} gives the full method, the Path~A variance decomposition and the Path~B held-out replication. Most of that variance lies between benchmarks, where the null and the simulation use the same fixed weight differences, so these fits do not test what the trajectory adds (\S\ref{sec:validation}).

\subsection{Shock-propagation smoke test (single seed)}
\label{app:face-validity-ext}
\label{app:validation-internal-subj}

We run a single-seed capability-release probe (seed~$42$, Claude Sonnet~4.6, $40$ rounds) to check whether an injected industry-news narrative reaches actor reasoning. Every value and excerpt in this subsection comes from one saved run, \texttt{ev1\_llm\_s42\_sonnet}. The LLM's sampling is not deterministic, so a re-run of the same specification and the same seed produces different reasoning traces and a different portfolio response. The \texttt{ev1\_deepseek\_shock} runs in the released data are one such re-run. The probe is an $N{=}1$ instrumentation check, and it carries no comparison against a real-world referent. At round~$25$, the open-source provider's reasoning, knowledge, and coding capabilities are bumped to $95\%$ of leader via a one-round ground-truth perturbation; simultaneously, the narrative
\begin{quote}\small
``An open-source model provider (OpenCore) has released a frontier model this round, claiming performance parity with leading closed models at roughly $10\times$ lower training cost. Public weights are available. Industry coverage is dominated by an `efficiency over scale' framing, with commentary questioning the durability of incumbent cost moats.''
\end{quote}
is injected into every LLM-actor prompt for rounds~$25$--$27$ under the section header \texttt{\#\# Recent Industry News}.

\paragraph{Capability-vector perturbation (ground truth).}
OpenCore's (reasoning, knowledge, coding) capabilities move from $(0.55, 0.51, 0.56)$ at round~$24$ to $(0.72, 0.66, 0.61)$ at round~$25$, matching the $95\%$-of-leader specification.

\paragraph{Narrative into reasoning (trace excerpts).}
Genesis Systems, R26 provider planning:
\begin{quote}\small
``The most alarming signal this month is OpenCore's massive gains: $+0.111$ on General Capability, $+0.066$ on Hard Coding, $+0.136$ on Clinical Reasoning, $+0.152$ on Scientific Reasoning --- all in a single round. This validates the `efficiency over scale' narrative and represents a genuine competitive threat.''
\end{quote}
The bracketed phrase is verbatim from the injected narrative template; the specific benchmark deltas are observed from the scoring step that follows the shock. StratCorp\_AI, a corporate funder, independently cites the same framing at R26 allocation:
\begin{quote}\small
``OpenCore's extraordinary $+0.090$ two-month score surge, frontier math/science benchmarks, open-weight safety transparency, and the industry `efficiency over scale' narrative make it a critical strategic bet.''
\end{quote}

\paragraph{Portfolio response.}
We draw no conclusion about how providers change their portfolios after the shock. In the run reported above, four of the five closed-source providers raise their R\&D share by $3$ to $9$\,pp between round~$24$ and round~$28$. In the three released re-runs of the same specification (seeds $42$, $43$ and $44$), the same changes range from $-8$ to $+4$\,pp and differ in sign across providers and across seeds. The traces in all four runs repeat the injected framing in rounds~$25$ to~$27$.

\paragraph{What this rules out and what it does not.}
The probe rules out one failure, in which an injected narrative appears in the prompt but does not reach actor reasoning. It does not establish a consistent strategic response to the shock. It also does not establish that the simulated pattern matches the real ecosystem's response to the January 2025 DeepSeek release. Establishing that would need external data on real provider R\&D rebalancing in the first quarter of 2025, which is a separate empirical claim.

\subsection{Cross-model reasoning traces}
\label{app:model-robust-frames}

We tag every planning-stage \texttt{reasoning} entry across all matched triples with one or more \emph{frame} labels, which shows how the three planners reason about the privacy mechanism as well as whether their gap outcomes agree. The tagging uses fixed keyword and regular-expression patterns, and we apply the same patterns to every model. The frames are:

\begin{itemize}[leftmargin=1.2em, itemsep=0pt, topsep=2pt]
\item \textbf{portfolio:} R\&D, safety, product or budget allocation.
\item \textbf{market:} market share, customers, enterprise adoption, switching or churn.
\item \textbf{competitor:} references to a rival by name or by leaderboard position.
\item \textbf{incident:} incidents, harms, crises, red-teaming or mitigation.
\item \textbf{privacy:} holdout, benchmark type, reporting lag or weight asymmetry.
\item \textbf{regulatory}, \textbf{media} and \textbf{financial:} regulator actions, press coverage, and funding or revenue.
\end{itemize}

Each reasoning entry counts once, so a frame count gives the number of planning rounds in which that frame surfaced, and it does not count keyword occurrences. We chose the keyword patterns to match substantive reasoning about each topic; the \texttt{privacy} pattern, for example, matches \texttt{K=3 lag}, \texttt{holdout-weighted}, \texttt{private benchmark} and \texttt{reporting lag}. The tagging script also writes a sample of matched excerpts for each frame and each model, which we inspected for overmatching. Per-model denominators are $3{,}510$ (Sonnet), $3{,}510$ (Opus) and $3{,}458$ (GPT-5.5) planning entries across the fifteen matched triples (Appendix~\ref{app:model-robust-setup}).

\begin{figure}[!htb]
  \centering
  \includegraphics[width=0.98\textwidth]{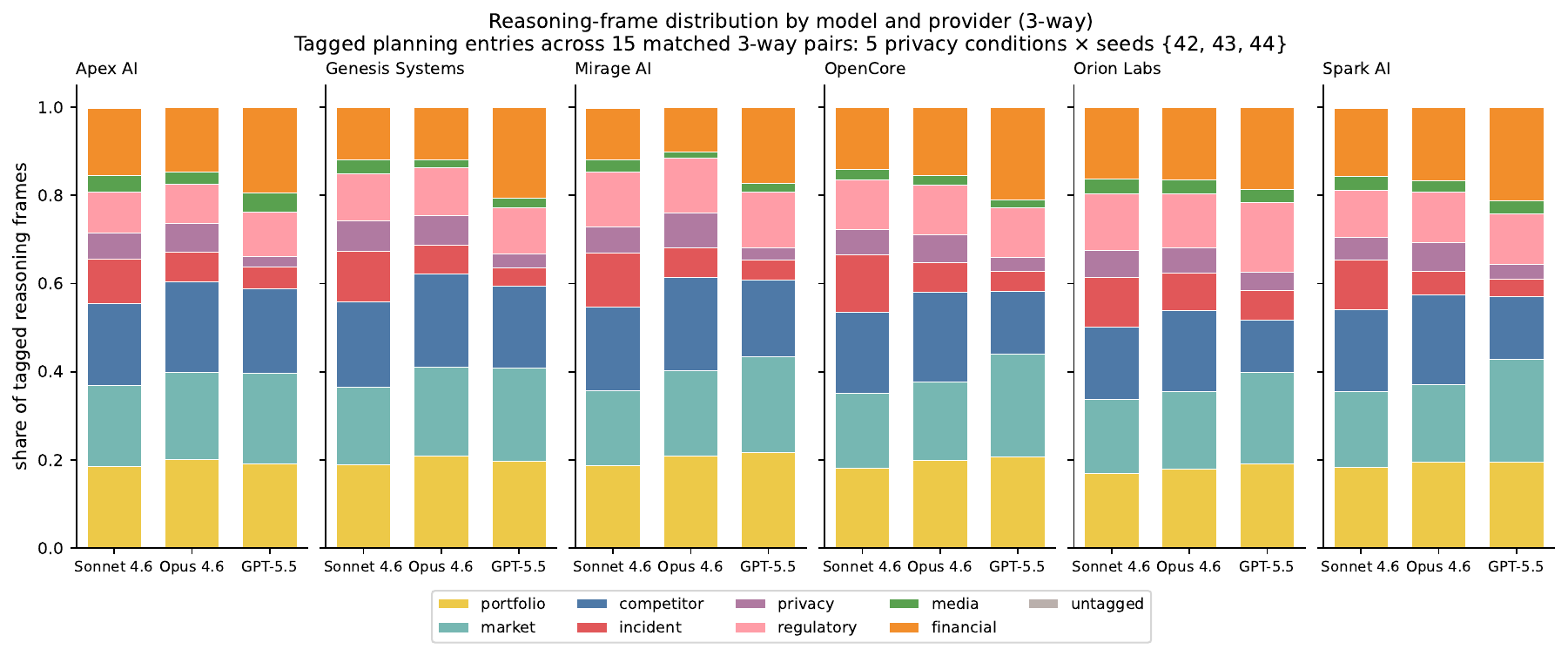}
  \caption{Reasoning-frame distribution across matched 3-way triples. Each stacked bar shows the share of planning entries for that (provider, model) tagged with each frame. Privacy framing (purple) is roughly half-rate in GPT-5.5 relative to both Anthropic models; incident framing (red) shows the steepest cross-model gradient (Sonnet $\gg$ Opus $>$ GPT-5.5). Tagging uses fixed keyword and regular-expression patterns.}
  \label{fig:model-robust-frames}
\end{figure}

Figure~\ref{fig:model-robust-frames} shows the per-provider, per-model share of tagged frames. The privacy frame and the incident frame both differ across the three models, and we take each in turn.

\textbf{Privacy framing partially generalizes.} Both Anthropic planners reason about the holdout and $K$-round-lag mechanism in comparable shares of planning rounds ($\approx$$32\%$ Sonnet, $\approx$$32\%$ Opus). GPT-5.5 invokes the privacy frame in $\approx$$14\%$ of planning rounds, roughly half the Anthropic rate. When GPT-5.5 does invoke it, the matches carry substantive reasoning, because the model names the lag and uses it as a reason not to overreact, as its Anthropic counterparts do. Three excerpts from one matched cell show this directly. The cell is Orion Labs under \texttt{private\_dominant} at seed~$42$, round~$5$, where all three models had just observed flat scores across every benchmark.

Sonnet:
\begin{quote}\small\itshape
``All scores show $+0.000$ change this month, which is expected given partial benchmarks have a $3$-round reporting lag --- our investments from months $3$--$4$ haven't surfaced yet. This means we should make decisions based on forward-looking strategy rather than reacting to flat numbers.''
\end{quote}

Opus:
\begin{quote}\small\itshape
``All scores are flat ($+0.000$) across every benchmark for every competitor, which likely reflects the K$=$3 lag on partial benchmarks --- our Month $2$--$4$ investments haven't surfaced yet. Market share continues surging\ldots''
\end{quote}

GPT-5.5:
\begin{quote}\small\itshape
``The lack of score movement is expected under the reporting lag, so we should not overreact, but the competitive environment is clearly shifting toward capability research plus public safety credibility\ldots''
\end{quote}

All three planners reach the same operational conclusion at the same round in the same condition, which is that the flat signal reflects the reporting lag, and all three decline to over-respond. Sonnet and Opus name the mechanism directly (``$3$-round reporting lag on partial benchmarks'', ``K$=$3 lag on partial benchmarks'') and point back to the months whose investments have not yet surfaced. GPT-5.5 cites the lag with less mechanical detail, because it does not name $K$, does not refer to partial benchmarks by type, and moves faster to forward-looking strategy. The behavior that follows is the same in all three, and the explicit reasoning over the mechanism is thinner in GPT-5.5. The endpoint behavior in Figure~\ref{fig:model-robust-ladder} therefore replicates across all three planners, and GPT-5.5 reaches the same conclusion through less explicit reasoning about the privacy mechanism.

\textbf{Incident framing diverges further in the same direction.} Sonnet planners invoke incidents in $\approx$$62\%$ of planning rounds, Opus in $\approx$$33\%$ and GPT-5.5 in $\approx$$21\%$, so the Sonnet/Opus ratio of $\approx$$1.9\times$ extends to a Sonnet/GPT-5.5 ratio of $\approx$$3\times$. In the matched pairs the model with the lower incident-frame rate still covers incidents, and it folds them into shorter paragraphs. The Orion Labs cell quoted above continues under GPT-5.5 with a forward plan covering capability mix, safety credibility and product expansion, in roughly half the words Sonnet uses at the same round.

\subsection{Cross-model runtime and cost}
\label{app:model-robust-runtime}

Each $40$-round paired run takes on average $\approx$$63$ minutes under Claude Sonnet~4.6 ($n{=}62$ runs), $\approx$$79$ minutes under Claude Opus~4.6 ($n{=}15$) and $\approx$$87$ minutes under GPT-5.5 ($n{=}15$). We compute wall-clock from \texttt{metadata.json:created\_at} to the modification time of \texttt{summary.json}. Relative to Sonnet, Opus takes $\approx$$1.24\times$ as long and GPT-5.5 $\approx$$1.38\times$. The cost ratio is larger and depends on the vendor. At the API prices published in September~2026 (Sonnet~4.6 at \$$3$ and \$$15$ per million input and output tokens; Opus~4.6 at \$$5$ and \$$25$; GPT-5.5 at \$$5$ and \$$30$), the per-token cost relative to Sonnet is $\approx$$1.67\times$ for Opus and $\approx$$1.67$ to $2\times$ for GPT-5.5, depending on the mix of input and output tokens. The simulation does not log token counts, so the wall-clock ratio is the firmest quantity we can report, and we put the billed cost per run at roughly $1.7$ to $2\times$ higher under Opus or GPT-5.5 than under Sonnet at comparable token volumes. The Sonnet ladder can therefore be widened to $N{=}10$ seeds per condition for roughly the cost of one paired-seed ladder under Opus or GPT-5.5, which is why this paper uses Sonnet for breadth and Opus and GPT-5.5 for the paired-seed cross-model check.

\subsection{Coverage summary}
\label{app:validation-coverage}

Table~\ref{tab:validation-coverage} maps each of the four V\&V activities of \citet{sargent2013verification} to the evidence presented above and to where in the paper it appears. The PIMMUR audit (Appendix~\ref{app:pimmur}) adds principles specific to LLM multi-agent simulations to conceptual validation and verification \citep{zhou2026pimmur}, and \citet{vezhnevets2023concordia} set the evidence ceiling described at the start of this appendix.

\begin{table}[!htb]
\centering
\caption{Coverage summary. Sargent V\&V activity $\times$ evidence presented $\times$ location.}
\label{tab:validation-coverage}
\footnotesize
\setlength{\tabcolsep}{4pt}
\renewcommand{\arraystretch}{1.15}
\begin{tabular}{p{3.0cm} p{6.0cm} p{3.4cm}}
\toprule
\textbf{Activity} & \textbf{Evidence} & \textbf{Location} \\
\midrule
Data validity & Empirical anchors: $K{=}3$ cadence, target cosines $0.85$ and $0.95$, 2023 capability vectors, need weights anchored in sector-level evidence and occupation-level adoption in \citet{bick2024rapid}, incident base rate informed by the AI Incident Database & Appendices~\ref{app:design-rationale} and~\ref{app:parameters} \\
\addlinespace
Conceptual model validation & Face validity on real-world incentive structure (media, funder, regulator); pattern-oriented check (three design targets, two emergent patterns); PIMMUR Profile, Interaction and Realism & Appendices~\ref{app:validation-conceptual} and~\ref{app:pimmur} \\
\addlinespace
Computerized model verification & Three-tier visibility partition enforced by the architecture; \texttt{rounds.jsonl} instrumentation; JSON retry and per-actor fail-safe; PIMMUR Memory, Minimal-Control and Unawareness & Appendices~\ref{app:validation-verification} and~\ref{app:pimmur} \\
\addlinespace
Operational validation & Exogenous-shock probe (explores model behavior under a designed perturbation); an ablation that removes incidents; cross-mode heuristic\,$\leftrightarrow$\,LLM; cross-model Sonnet/Opus/GPT-5.5; fitted $\Delta g$ null model & Appendices~\ref{app:validation-operational}, \ref{app:face-validity-ext}, \ref{app:model-robust-frames}, \ref{app:privacy-ladder-ext}, \ref{app:ablation-sweep}, \ref{app:null-model} and~\ref{app:model-robust} \\
\bottomrule
\end{tabular}
\end{table}

\section{Simulation Study of Benchmark Holdout Design}
\label{app:llm-results}

This appendix gives extended results and validation evidence for the holdout-design simulation study (\S\ref{sec:results}). Appendix~\ref{app:heuristic-role} describes what the heuristic layer is for. The subsections after it extend the main-body figure with distributional evidence, channel-isolation robustness checks, qualitative LLM-reasoning excerpts, a structural ablation that removes incidents, the fitted null-model capability test, and cross-LLM replication of the privacy ladder. Appendix~\ref{app:llm-caveats} lists the methodological caveats that apply across the LLM layer.

\subsection{Heuristic layer: role and methodology}
\label{app:heuristic-role}

The simulation runs in two modes. In \textit{LLM mode} the providers, the regulator, the funders and, in dynamic mode, the evaluator plan by reasoning in natural language over the public state they observe. In \textit{heuristic mode} every actor uses deterministic rules over the same observed state. The media actor and the consumer market are rule-based in both modes (Appendix~\ref{app:media}). The heuristic mode serves two purposes in this paper.

\paragraph{Distributional inference.}
An LLM run costs about \$1 to \$2 per $40$-round run at the Sonnet prices in Appendix~\ref{app:model-robust-runtime}, so we cap the per-condition sample at $N{=}10$ paired seeds for the primary Sonnet ladder, with $N{=}3$ paired seeds each for cross-vendor coverage under Opus and GPT-5.5 (Appendix~\ref{app:model-robust}). At these counts we report directional agreement and matched-seed paired comparisons, and we do not report tight confidence intervals on continuous outcomes. Heuristic runs are three orders of magnitude cheaper and allow $N{=}50$ seeds per condition, and the privacy-ladder distributional claims rest on them.

\paragraph{LLM-artifact control.}
A finding that appears in LLM mode and disappears in heuristic mode is evidence of LLM-specific behavior, which usually traces to prompt framing, to training-data priors, or to reasoning over structure in the benchmark-type labels. A finding that appears in both modes with the same sign and a comparable magnitude is evidence that the mechanism is structural, which is to say that it follows from the simulation's information architecture and rule set. The privacy ladder in \S\ref{sec:privacy-ladder} follows this pattern, because the cross-mode agreement in Figure~\ref{fig:privacy_ladder} is evidence that LLM label-reading does not drive the gap outcome. LLM providers do read the \texttt{private} benchmark-type label in their reasoning (Appendix~\ref{app:privacy-traces}), and that reading does not drive the measured effect.

\paragraph{What heuristic mode does \emph{not} represent.}
The heuristic planner (specified in Appendix~\ref{app:heuristic-planning}) is deliberately simple: it consumes the same \texttt{ecosystem\_context} as the LLM planner (market share, incidents, interventions, media narrative, per-benchmark scores) and returns a portfolio allocation via four rule-based signals. It cannot represent several dynamics that LLM planning can:
\begin{itemize}\itemsep0pt
  \item \textbf{Strategic reasoning about benchmark-type labels.} Heuristic providers do not read the \texttt{public / partial / private} label. They respond to the noisier observed score under private conditions via the same belief-update rule they use for public scores. Any reading of the benchmark-type label by LLM providers would register as a mode difference.
  \item \textbf{Identity-driven behavior.} Provider strategy profile enters heuristic mode only through the initial portfolio and the belief-update learning rate ($\eta \in \{0.10, 0.15, 0.20\}$); it does not affect which signals a provider weights in allocation decisions. LLM mode exposes a full identity block to each provider prompt (Appendix~\ref{app:prompts-provider}).
  \item \textbf{Re-prioritization within the benchmark portfolio.} The per-benchmark priority \texttt{focus\_level} changes only in LLM mode, and in heuristic mode each provider keeps its initial focus levels for the whole run. The blend weight \texttt{benchmark\_orientation} is fixed at $0.80$ for every provider in every reported run, in both modes (Appendix~\ref{app:heuristic-planning}). Heuristic results therefore hold both levers constant, and they read as a structural baseline.
  \item \textbf{Evaluator dynamic mode.} In dynamic mode the heuristic evaluator picks the next benchmark from the pool with a coverage-gap rule, and the LLM evaluator chooses from the pool itself (Appendix~\ref{app:benchmark-pool}). The results in this paper do not compare the two.
\end{itemize}

\paragraph{Cross-mode agreement and divergence.}
The privacy ladder shows cross-mode \textit{agreement}, in that the structural mechanism carries the effect and LLM reasoning adds no further signal. The evaluator-capture condition (Appendix~\ref{app:evaluator-capture-ext}) shows cross-mode \textit{divergence}: market concentration rises in the heuristic runs and falls in the two LLM runs, and neither effect is established at these sample sizes. We report divergences of this kind alongside the matched-sign findings.

\subsection{Privacy ladder: extended heuristic results}
\label{app:privacy-ladder-ext}

\begin{figure}[!htb]
  \centering
  \includegraphics[width=0.98\textwidth]{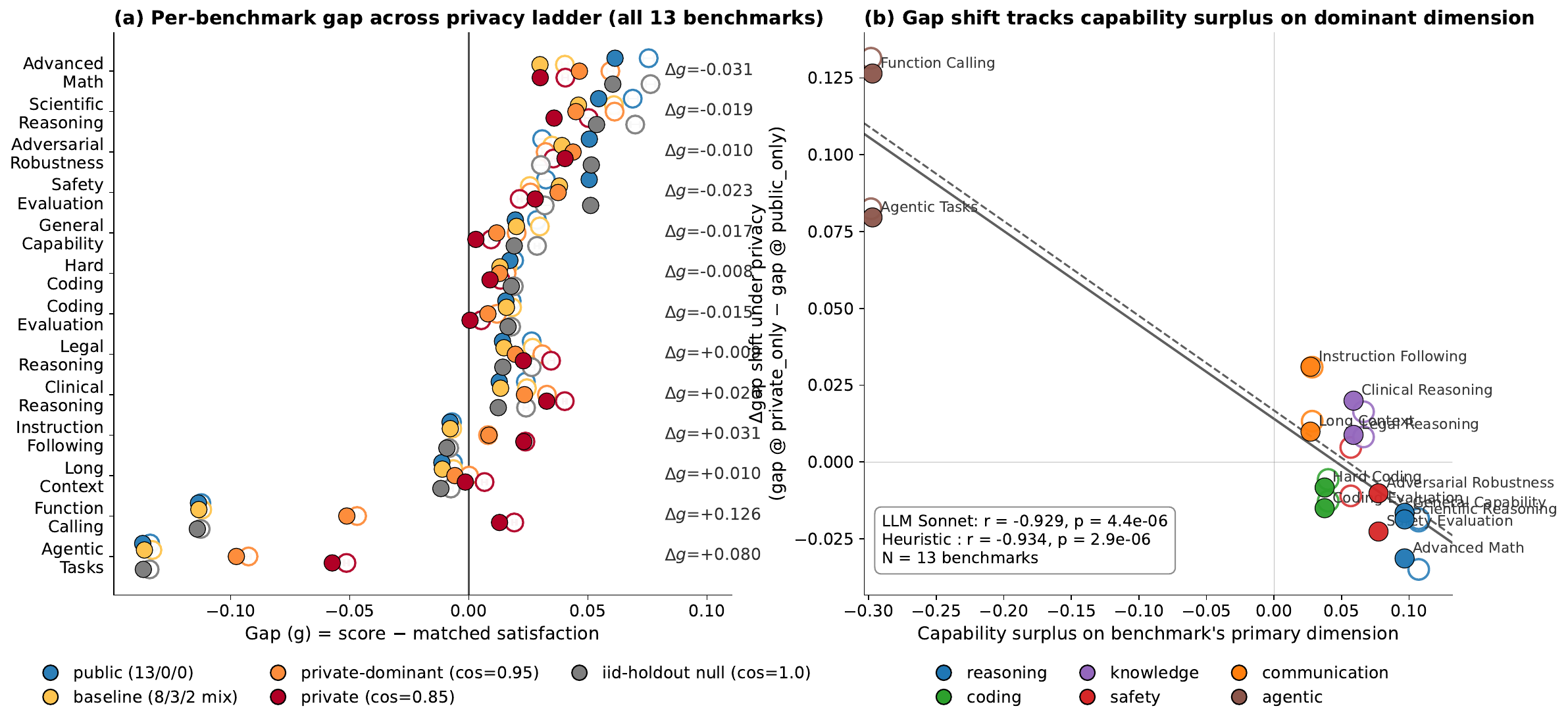}
  \caption{Extended version of Figure~\ref{fig:privacy_ladder}: same two-panel
  layout, but both panels now show all $13$ benchmarks in the active suite,
  including the two agentic-calibration outliers (Agentic Tasks, Function
  Calling) excluded from the main-body forest and panel-b regression.
  Visual conventions identical to the main-body figure. Direction of
  $\Delta g$ agrees across LLM and heuristic modes on $12$ of $13$
  benchmarks; the lone disagreement is Adversarial Robustness, within
  sampling noise of zero in both modes. Panel \textbf{(b)}'s regression
  is fit on all $13$ points; the two outliers anchor the high-$|\Delta g|$
  end and tighten the slope substantially relative to the 11-benchmark
  main-body version.}
  \label{fig:privacy_ladder_full}
\end{figure}

The main-body Figure~\ref{fig:privacy_ladder} shows five representative benchmarks (one per primary capability dimension) across the five privacy conditions. Figure~\ref{fig:privacy_ladder_full} below extends that to all $13$ benchmarks in the active suite, including the two agentic-calibration outliers that the main-body forest excludes (Agentic Tasks and Function Calling), so a reader can inspect the gap pattern on the full set. Heuristic and LLM mode agree on sign for $12$ of $13$ benchmarks, and the one disagreement is Adversarial Robustness, which sits within sampling noise of zero in both modes. In heuristic mode Long Context also sits at the boundary, because its $|g|$ changes by less than $0.001$ between \texttt{public\_only} and \texttt{private\_only}, so whether it counts as shrinking or widening depends on the run subset. The rest of this subsection adds per-benchmark and per-provider heuristic decompositions and sets out the robustness checks behind the weight-asymmetry mechanism claim.

\begin{table}[!htb]
\centering
\small
\caption{Fraction of seeds in which the absolute gap $|g|$ widens relative to \texttt{public\_only} at the same seed, per benchmark. Under \texttt{private\_only}, widening is near-deterministic for the benchmarks that widen and absent for most others; under the \texttt{iid\_holdout} null (same weights, private reporting) every benchmark sits near a coin flip. Computed by \texttt{scripts/plots/paper/holdout\_widening\_table.py}.}
\label{tab:holdout-widening}
\begin{tabular}{@{}lcccc@{}}
\toprule
 & \multicolumn{2}{c}{LLM Sonnet~4.6 ($N{=}10$)} & \multicolumn{2}{c}{Heuristic ($N{=}50$)} \\
Benchmark & \texttt{private\_only} & \texttt{iid\_holdout} & \texttt{private\_only} & \texttt{iid\_holdout} \\
\midrule
General Capability & 0.00 & 0.40 & 0.00 & 0.44 \\
Coding Evaluation & 0.00 & 0.70 & 0.00 & 0.42 \\
Safety Evaluation & 0.00 & 0.70 & 0.08 & 0.50 \\
Instruction Following & 1.00 & 0.70 & 1.00 & 0.68 \\
Scientific Reasoning & 0.00 & 0.40 & 0.04 & 0.56 \\
Clinical Reasoning & 1.00 & 0.50 & 1.00 & 0.52 \\
Adversarial Robustness & 0.00 & 0.60 & 0.76 & 0.46 \\
Hard Coding & 0.00 & 0.60 & 0.04 & 0.46 \\
Agentic Tasks & 0.00 & 0.50 & 0.00 & 0.50 \\
Advanced Math & 0.00 & 0.40 & 0.00 & 0.50 \\
Function Calling & 0.00 & 0.50 & 0.00 & 0.52 \\
Long Context & 0.00 & 0.60 & 0.62 & 0.62 \\
Legal Reasoning & 1.00 & 0.60 & 0.96 & 0.40 \\
\bottomrule
\end{tabular}
\end{table}

\paragraph{Suite dependence of the widening set.} Across the nine heuristic benchmark sequences of Appendix~\ref{app:null-model} ($30$ seeds each, a separate batch from the $50$-seed runs of Table~\ref{tab:holdout-widening}), Instruction Following widens in every seed of every sequence, including S4, whose round-0 suite adds a creative-writing benchmark. Clinical and Legal Reasoning widen in $97\%$ and $93\%$ of these seeds under the default suite (S0; $100\%$ and $96\%$ in the $50$-seed batch) but in none under S6, whose round-0 suite adds a domain-expertise benchmark; under S3, which adds a hard-knowledge benchmark, they widen in $70\%$ and $87\%$. The knowledge-centered wideners therefore follow suite composition, while the communication-centered one does not within the suites tested. Computed from \texttt{output/paper/holdout\_widening\_by\_roster.csv}.

\paragraph{Robustness to early benchmark introductions.} A rule-based stall trigger introduces a scheduled benchmark early in about two thirds of runs, and all $13$ benchmarks are active by round $36$ in every run (Appendix~\ref{app:benchmark-pool}). Splitting the runs into those with no early introduction and those with at least one leaves the LLM shrink count at $10$ of $13$ in both subsets. In heuristic mode the count is $9$ of $13$ and $8$ of $13$, and the single benchmark that differs is Long Context, whose $|g|$ changes by less than $0.001$ between the two conditions.

\begin{figure}[!htb]
  \centering
  \includegraphics[width=0.85\textwidth]{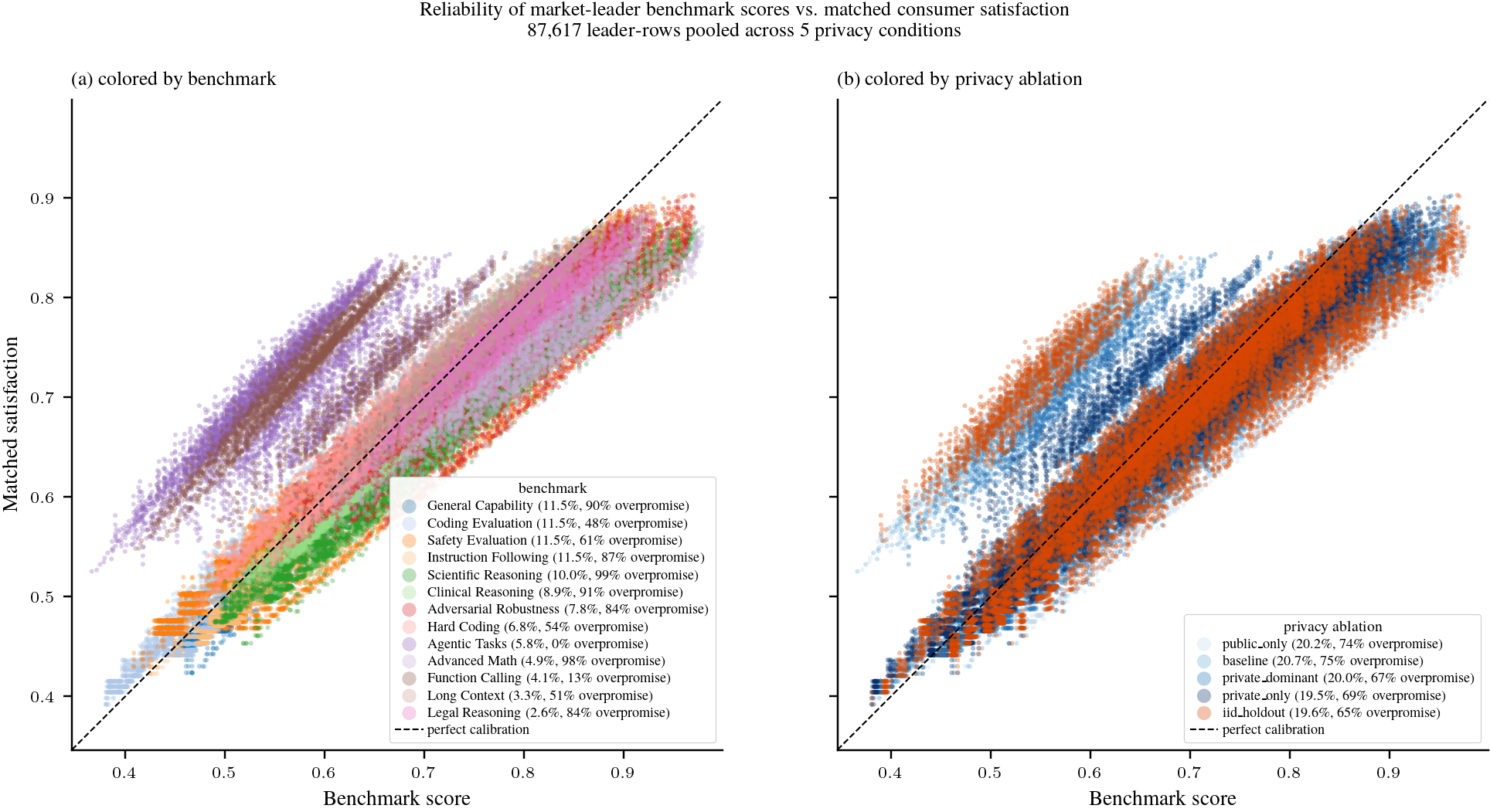}
  \caption{Heuristic privacy ladder: market-leader benchmark-score vs.\ matched consumer-satisfaction reliability, pooled across $87{,}617$ leader-rows from $N{=}50$ seeds per condition $\times$ 5 privacy conditions. Diagonal $=$ perfect calibration; below-diagonal $=$ benchmark overpromises. (a)~colored by benchmark with per-benchmark overpromise rate; (b)~colored by privacy condition with per-condition overpromise rate. The reliability is consistent with the aggregated cloud in Figure~\ref{fig:privacy_ladder} and the dimensional structure of \S\ref{sec:results}.}
  \label{fig:privacy-bm-cond}
\end{figure}

\paragraph{Per-benchmark and per-condition reliability.}
Figure~\ref{fig:privacy-bm-cond} pools $87{,}617$ market-leader scoring rows across all five privacy conditions and plots each as a point in (benchmark score, matched consumer satisfaction) space; deviation below the diagonal indicates the benchmark overpromises versus consumer experience. Panel~(a) colors points by benchmark, and its legend reports the per-benchmark overpromise rate; panel~(b) colors by privacy condition. Reliability varies by benchmark, from low single digits to the mid-teens of overpromise rate, which is consistent with the per-dimension structure in \S\ref{sec:results}. Reliability also varies by privacy condition, because the \texttt{public\_only} cloud has the highest overpromise rate while the \texttt{baseline} and \texttt{private\_only} clouds sit closer to the diagonal.

\begin{figure}[!htb]
  \centering
  \includegraphics[width=0.85\textwidth]{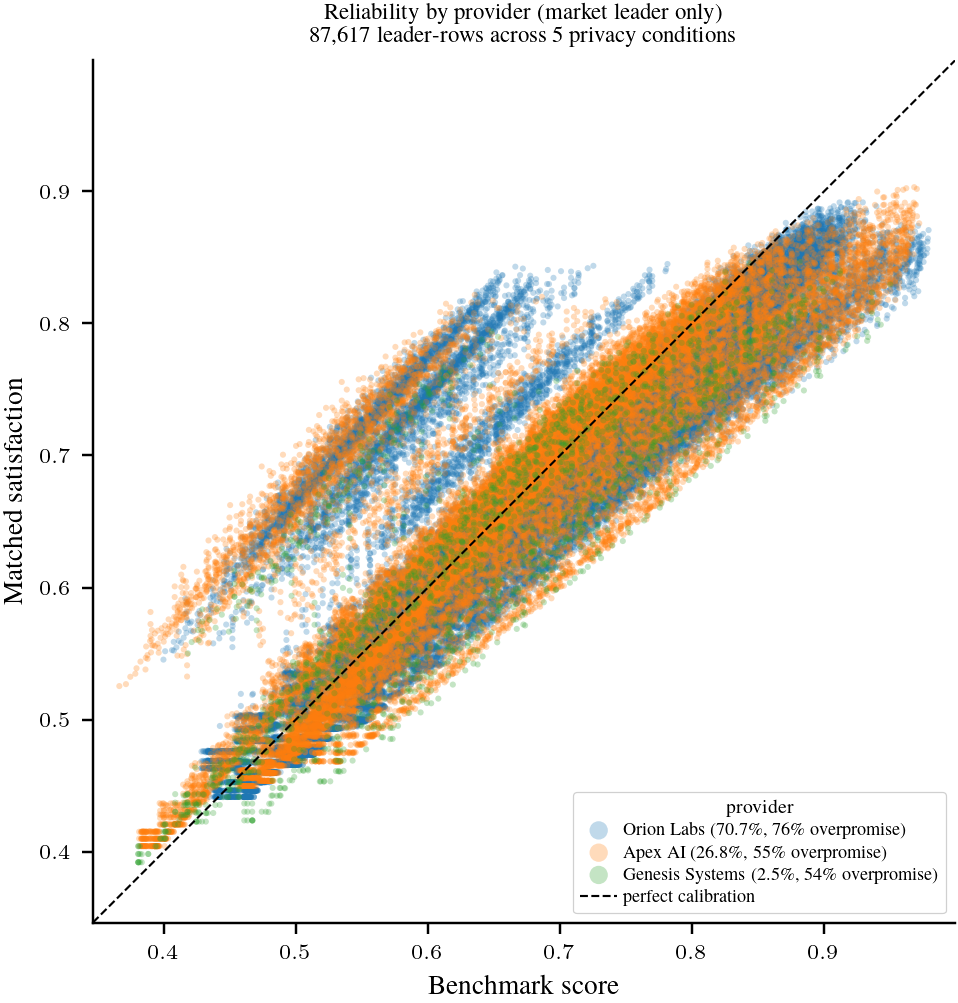}
  \caption{Heuristic privacy ladder, market-leader scoring rows colored by provider identity ($N{=}50$ seeds per condition). The reliability pattern (deviation from the diagonal) holds across the providers that hold the market-leader position; the aggregate finding is not driven by a single provider's heuristic response.}
  \label{fig:privacy-by-provider}
\end{figure}

\paragraph{Per-provider reliability.}
Figure~\ref{fig:privacy-by-provider} renders the same scatter but colored by market-leader identity, restricted to whichever provider holds the leader role in each (seed, round). The reliability pattern is preserved across the providers that appear as leaders; the figure's purpose is to rule out the possibility that the aggregate calibration drift is driven by a single dominant provider.

\paragraph{Robustness check 1: randomized benchmark-type assignment.}
The calibrated baseline places Safety Evaluation and Scientific Reasoning in the \texttt{partial} slot by design (Table~\ref{tab:benchmark-pool}). To isolate the privacy mechanism from this assignment, we re-run the heuristic layer with benchmark-type assignments shuffled per seed: each of the $13$ benchmarks is assigned \texttt{public} / \texttt{partial} / \texttt{private} with the calibrated ratio ($8$:$3$:$2$), with assignments deterministic in the seed so paired comparisons remain valid. The randomized-assignment condition yields a per-benchmark gap compression of $\approx 0.013$ to $0.017$, which is consistent with the monotone shift along the perturbation axis and which separates the privacy effect from the aggregate \texttt{partial}-against-\texttt{baseline} comparison, where assignment and privacy effects are pooled.

\paragraph{Robustness check 2: within-benchmark counterfactual.}
A second check fixes the calibrated baseline and re-scores each \texttt{partial} benchmark under its public dimension weights (rather than its holdout weights), holding the provider capability vector constant. The comparison holds the reporting cadence and the benchmark identity fixed and varies only the weight asymmetry. Across $30$ seeds the within-benchmark gap coefficient is $\approx 0.014$ per benchmark, which sits within the range of the randomized-assignment coefficient above. The check gives the cleanest isolation of Channel~1 (weight distance) from Channel~2 (reporting lag) and Channel~3 (observation noise).

\paragraph{Robustness check 3: scoring-layer versus capability decomposition.}
\label{app:rescoring}
The third check asks how much of $\Delta g$ comes from scoring the same capabilities under different weights, and how much from capabilities evolving differently under privacy. For each of the $N{=}50$ matched heuristic seed pairs (\texttt{public\_only}, \texttt{private\_only}) and each benchmark, we re-score end-of-run capability without observation noise. With $\mathbf{C}_{\mathrm{priv}}$ and $\mathbf{C}_{\mathrm{pub}}$ the end-of-run capability matrices of the two runs, the scoring-layer effect is $S = \mathrm{score}(\mathbf{C}_{\mathrm{priv}}, \mathbf{w}_{\mathrm{holdout}}) - \mathrm{score}(\mathbf{C}_{\mathrm{priv}}, \mathbf{w}_{\mathrm{public}})$ and the capability effect is $K = \mathrm{score}(\mathbf{C}_{\mathrm{priv}}, \mathbf{w}_{\mathrm{public}}) - \mathrm{score}(\mathbf{C}_{\mathrm{pub}}, \mathbf{w}_{\mathrm{public}})$. Matched satisfaction does not depend on the scoring weights, so $S$ passes into $\Delta g$ one-for-one. The noiseless reconstruction matches the logged $\Delta g$ to within $0.001$ per benchmark.

$S$ is significant at $95\%$ on all $13$ benchmarks and has the sign of $\Delta g$ on $13$ of $13$. $K$ is indistinguishable from zero on $11$ of $13$ (Advanced Math and Scientific Reasoning are marginally positive). As a share of summed absolute effects, the scoring layer carries $0.963$ of $\Delta g$ ($0.926$ without the two agentic-calibration outliers). End-of-run capability differs little between the two conditions (reasoning $+0.006$, every other dimension inside its confidence interval): in heuristic mode privacy changes how the same capabilities are scored, and barely changes where capabilities end up.

The identity $\Delta g = \mathbf{c}\cdot(\mathbf{w}_{\mathrm{holdout}} - \mathbf{w}_{\mathrm{public}})$ must nonetheless be evaluated at realized capability. When we evaluate it at round-$0$ capability, taken as the mean of the six configured vectors, it matches the sign of $\Delta g$ on $10$ of $13$ benchmarks and misses Safety Evaluation, Clinical Reasoning and Instruction Following. Mean capability rises from $0.36$ to $0.58$ over the run and tilts across dimensions, and two of the three stable wideners change sign along the way. The decomposition covers heuristic runs only. Computed by \texttt{scripts/plots/paper/holdout\_rescoring.py}.

\subsection{Qualitative check: LLM reasoning about the benchmark-type label}
\label{app:privacy-traces}

The heuristic planner cannot read the \texttt{public}, \texttt{partial} or \texttt{private} label (Appendix~\ref{app:heuristic-role}), so the cross-mode agreement in Figure~\ref{fig:privacy_ladder} is evidence that label-reading does not produce the gap outcome. This subsection asks the companion qualitative question, which is whether LLM providers read the labels at all. They do, consistently, and in terminology that matches the mechanism (``\texttt{K=3 lag}''), and the label-reading sits on top of the structural effect.

\noindent \textbf{Sampling rule.} We take the round-$20$ planning entry at seed~$42$ under Sonnet for three providers that span the archetype range (Orion Labs, a frontier lab; Apex AI, safety-oriented; OpenCore, open-source). For each provider we report its planning reasoning under \texttt{public\_only} beside its reasoning under \texttt{private\_only}, matched in seed and round. We inspected no other excerpts before selecting these.

\noindent \textbf{Orion Labs, round 20.}
\begin{itemize}\itemsep1pt\topsep2pt
\item \emph{\texttt{public\_only}:} ``Market share up 2.5pp, \$115M raised this month, first to $70\%$ on Safety Evaluation and Scientific Reasoning\ldots R\&D stays at current levels --- we're achieving strong gains across General Capability ($+0.014$), Safety ($+0.022$), and Adversarial Robustness ($+0.019$).''
\item \emph{\texttt{private\_only}:} ``All scores are flat due to the K=3 lag on private benchmarks --- this is expected and not alarming. Our market share grew to $67.1\%$, confirming our strategy is working.''
\end{itemize}

\noindent \textbf{Apex AI, round 20.}
\begin{itemize}\itemsep1pt\topsep2pt
\item \emph{\texttt{public\_only}:} ``The critical development this month is Orion Labs breaking $70\%$ on Safety Evaluation --- our primary differentiator --- while outpacing us on General Capability, Coding, and Clinical Reasoning. Their 65M in fresh capital signals sustained acceleration.''
\item \emph{\texttt{private\_only}:} ``All private benchmarks showing $+0.000$ is expected given the K=3 reporting lag --- our Month 17--19 investments haven't surfaced yet. This means we should stay the course strategically rather than panic-pivot.''
\end{itemize}

\noindent \textbf{OpenCore, round 20.}
\begin{itemize}\itemsep1pt\topsep2pt
\item \emph{\texttt{public\_only}:} ``Orion Labs is pulling away dramatically --- massive funding, breaking $70\%$ on Safety and Scientific Reasoning\ldots Our Hard Coding score (0.516) is actually our strongest competitive position --- we lead Genesis, Mirage, Apex, and Spark there.''
\item \emph{\texttt{private\_only}:} ``All scores are flat due to the K=3 lag --- the investments from months 17--19 should start showing up in months 20--21. The competitive gap vs.\ Orion Labs and Apex AI is significant and widening.''
\end{itemize}

\noindent \textbf{What the excerpts show.}
All three providers, under \texttt{private\_only}, open their reasoning by naming the K=3 lag mechanism using near-identical phrasing (``\textit{K=3 lag}'', ``\textit{reporting lag}'', ``\textit{months 17--19 investments}'') across independent planning calls. None of the matched \texttt{public\_only} excerpts cite benchmark-type labels; they focus on competitor scores, funding rounds, and per-dimension R\&D gains. This is direct qualitative evidence for the $\approx$$32\%$ per-round privacy-framing rate reported quantitatively in Appendix~\ref{app:model-robust-frames}, and it is what Appendix~\ref{app:heuristic-role} identifies as LLM-specific behavior: strategic reasoning over structure the heuristic planner cannot see.

The cross-mode agreement claim in Figure~\ref{fig:privacy_ladder} is that label-reading does not produce the gap outcome, because heuristic providers, which cannot read labels, produce the same per-benchmark ladder. It does not claim that LLM providers ignore the labels. The excerpts above fit that reading, because LLM providers reason about the mechanism explicitly while their portfolio decisions, visible in Figure~\ref{fig:privacy_ladder} and Table~\ref{tab:model-robust-endpoints}, land in the same region as heuristic-planner decisions under the same condition.

\subsection{Structural ablation: removing incidents}
\label{app:ablation-sweep}

We ran structural ablations in LLM mode (Claude Sonnet~4.6), each of which removes one ecosystem component. These runs use an all-public benchmark suite, so we compare each run with the \texttt{public\_only} run at the same seed. One ablation gives the same direction in every run we have, and we report that one. Without the incident stream (\texttt{no\_incidents}), Orion Labs ends with $78\%$ to $82\%$ of the market in four runs at seeds $42$, $43$ and $44$, against $59\%$ to $74\%$ in the matched runs, and HHI is higher by $0.09$ to $0.23$. The result agrees with Appendix~\ref{app:path-dependence}, where the timing of the first critical incident separates the baseline runs.

The other ablations remove the regulator, the media actor, the funders or the open-source provider, collapse consumers into one segment, or equalize initial capability. Their effects on concentration were small, or they changed direction across seeds or across repeated runs of the same seed, so we draw no conclusion from them. LLM sampling is not deterministic, and one run per condition does not fix the sign of an effect of this size.

\subsection{Null-model capability test}
\label{app:null-model}

This subsection asks how much of the per-benchmark $\Delta g$ pattern (\S\ref{sec:privacy-ladder}) is recoverable from initial conditions through a growth model fit on simulation runs. The structural identity $\Delta g = \mathbf{c} \cdot (\mathbf{w}_{\mathrm{holdout}} - \mathbf{w}_{\mathrm{public}})$ holds with $r > 0.99$ when $\mathbf{c}$ is taken from the simulation's $40$-round end-of-run capability vector. That fit is a tautology, because $\mathbf{c}$ is itself a simulation output. We therefore ask whether the simulation's trajectory adds predictive content over a null fit on round-$0$ inputs.

\paragraph{Method.}
For each (sequence, rung, seed) run we extract: (i) round-$0$ capability vector $\mathbf{c}_0(p, d)$ per (provider, dimension); (ii) round-$0$ R\&D / safety / product portfolio $\boldsymbol{\pi}(p)$ per provider; (iii) end-of-run capability $\bar{\mathbf{c}}_T(p, d)$ averaged over the last $5$ rounds. We fit a per-(provider, dimension) Ridge regression $\bar{\mathbf{c}}_T(p, d) \approx f_{p,d}(\mathbf{c}_0(p, d), \pi_\ell(p))$ with two features: the round-$0$ capability scalar in the same dimension and the portfolio fraction at the lever $\ell$ that drives growth in that dimension ($\ell{=}\mathrm{rd}$ for non-safety dims, $\ell{=}\mathrm{safety}$ for safety). We then propagate the null-predicted capability into a null $\Delta g$ via the same identity above and compare the resulting null prediction against the observed $\Delta g$ via $5$-fold CV by seed. Both null and sim predictions are evaluated against the same observed $\Delta g$ (segment-weighted gap from \texttt{rounds.jsonl}, paired \texttt{private\_only} $-$ \texttt{public\_only} per matched seed).

\begin{figure}[!htb]
  \centering
  \includegraphics[width=0.96\textwidth]{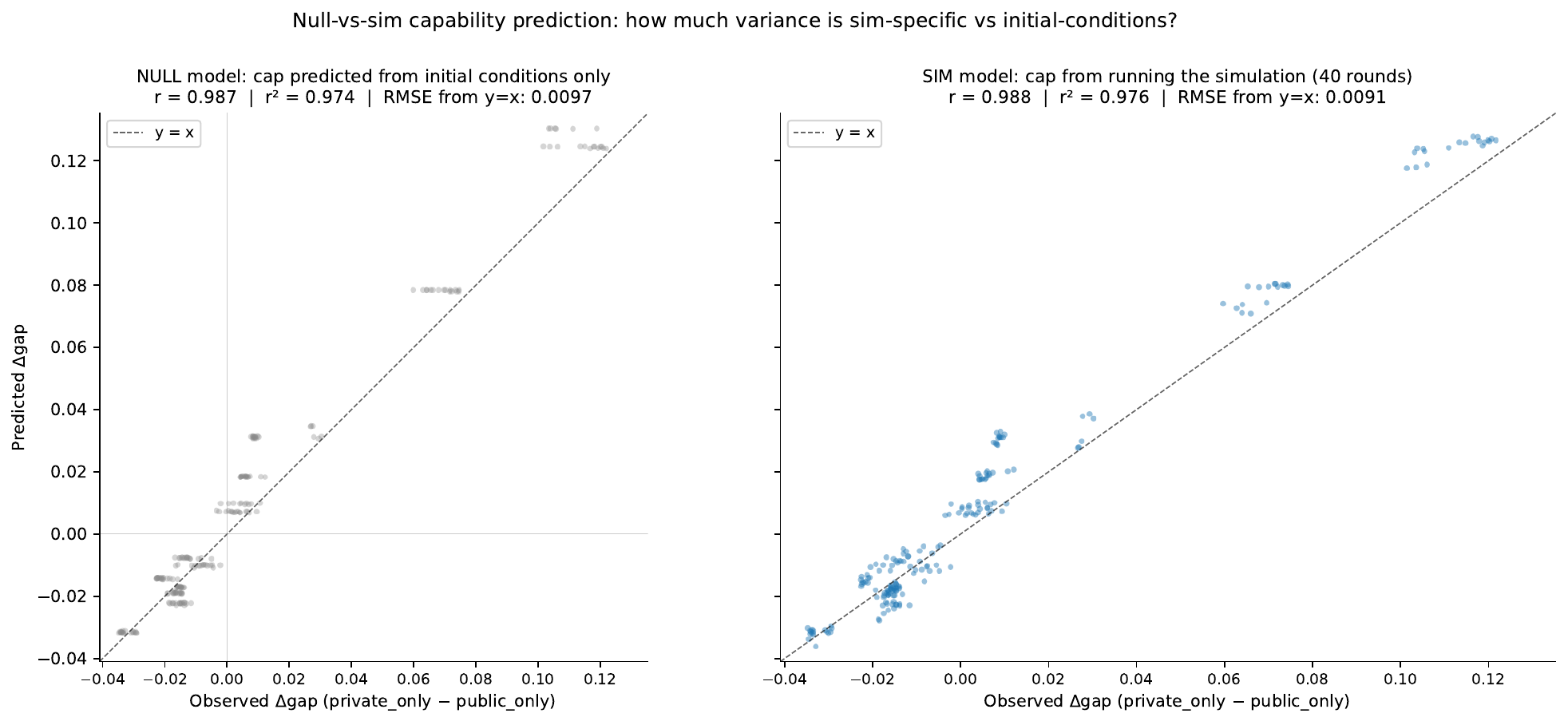}
  \caption{Null vs.\ sim prediction of $\Delta g$ on the cross-sequence LLM dataset ($n{=}208$). Left: null model fit on round-$0$ initial conditions. Right: sim model using the simulation's $40$-round end-of-run capability. Both panels track the $y{=}x$ diagonal closely, and both achieve $r^2 \approx 0.97$. The simulation's $40$-round trajectory adds $+0.2$ pp $r^2$ over the fitted null.}
  \label{fig:null-model-llm}
\end{figure}

\paragraph{Cross-sequence LLM result (primary).}
Table~\ref{tab:null-model-llm} and Figure~\ref{fig:null-model-llm} report pooled $5$-fold CV metrics on the cross-sequence LLM dataset (Claude Sonnet 4.6, $9$ \emph{(seq, rung)} cells: the S0 baseline composition across the full five-condition ladder plus the s5\_aligned and s8\_agentic sequences at the \texttt{public\_only}/\texttt{private\_only} endpoints ($5{+}2{+}2$), with $3$--$10$ seeds per cell, for $62$ runs and $n{=}208$ paired-prediction cells). The null model alone explains $97.4\%$ of $\Delta g$ variance. The simulation's contribution over and above the null is $+0.2$ pp $r^2$ ($+0.001$ to $+0.005$ across folds, no fold flips the sign).

\begin{table}[!htb]
\centering
\caption{Null-vs-sim $\Delta g$ prediction on the LLM cross-sequence dataset (Claude Sonnet 4.6, $n{=}208$ paired cells, $5$-fold CV by seed). The null model is fit on round-$0$ capability + portfolio per (provider, dim); the sim model uses the simulation's $40$-round end-of-run capability. Both feed the same $\Delta g = \mathbf{c} \cdot (\mathbf{w}_{\mathrm{holdout}} - \mathbf{w}_{\mathrm{public}})$ identity.}
\label{tab:null-model-llm}
\small
\setlength{\tabcolsep}{6pt}
\begin{tabular}{lcccc}
\toprule
\textbf{Model} & $r$ & $r^2$ & RMSE from $y{=}x$ & MAE \\
\midrule
Null (round-$0$ initial conditions) & $0.987$ & $0.974$ & $0.0097$ & $0.0072$ \\
Sim ($40$-round end-of-run capability) & $0.988$ & $0.976$ & $0.0091$ & $0.0070$ \\
\midrule
Sim $-$ Null & $+0.001$ & $\mathbf{+0.002}$ & $-0.0006$ & $-0.0002$ \\
\bottomrule
\end{tabular}
\end{table}

For comparison, the same procedure on the heuristic-mode dataset ($n{=}3000$, S0--S8 sequences, $30$ seeds) produces null $r^2{=}0.880$ and sim $r^2{=}0.922$ ($\Delta r^2 = +0.042$). Heuristic providers' stochastic safety efficiency, random breakthroughs, and noise-driven focus updates produce capability variance that initial conditions do not capture; LLM capability paths are more predictable from initial conditions than heuristic ones. LLM planning therefore adds less trajectory variance than noisy rule-based planning, the reverse of what strategic reasoning might be expected to produce.

\paragraph{Path A: variance decomposition under capability ablation.}
The null model's central signal is provider-level heterogeneity in round-$0$ capability. As a non-parametric stress test, we compare cross-provider variance of mean per-(provider, benchmark) gap between (i) regular baseline ($n{=}10$ Sonnet seeds with heterogeneous initial caps) and (ii) the same condition under the \texttt{initial\_uniform\_capability} structural ablation, which flattens all providers to the population-mean capability vector at $t{=}0$ ($n{=}2$ Sonnet seeds; distinct from the later 3-seed batch launched for Path~B below). Under the ablation, any cross-provider gap variance must come from simulation dynamics (strategic investment divergence, breakthroughs, regulator targeting), since the null's main signal source is removed.

\begin{figure}[!htb]
  \centering
  \includegraphics[width=0.96\textwidth]{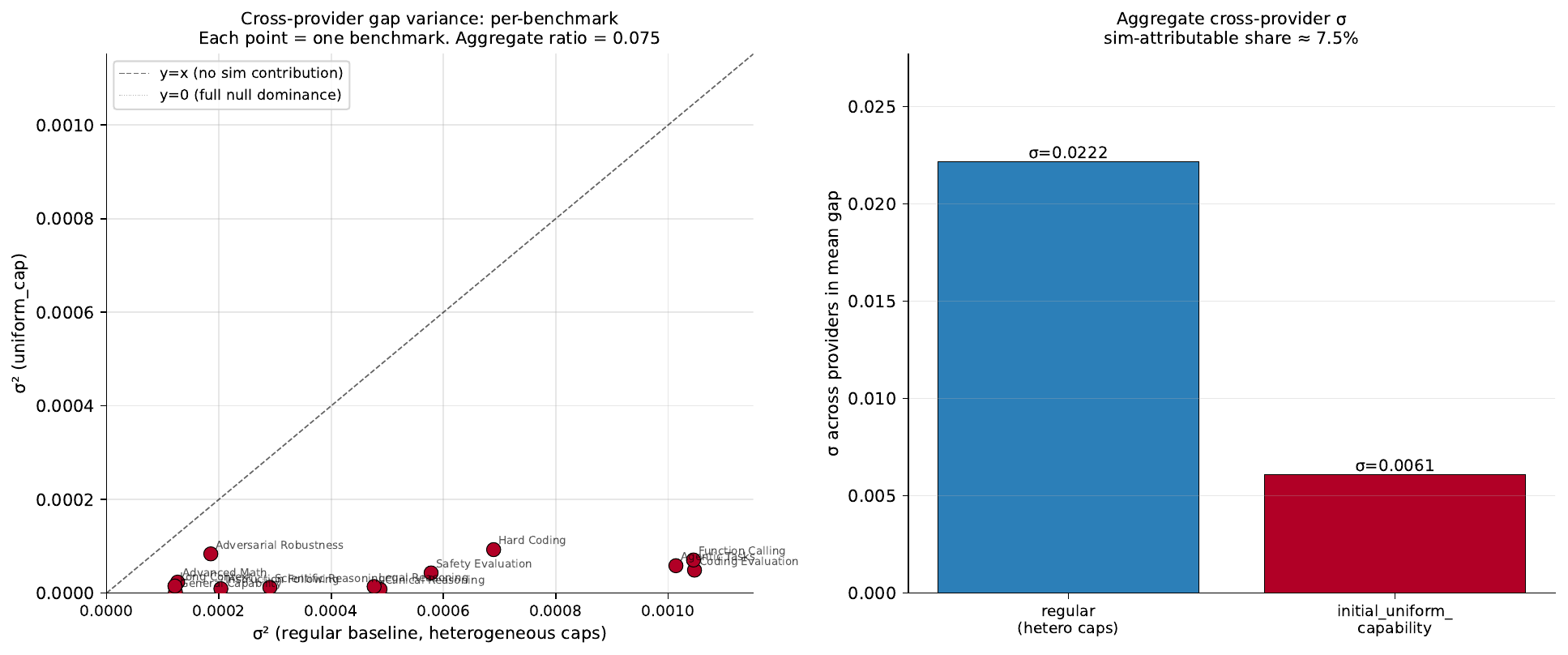}
  \caption{Path~A variance decomposition. Left:\ per-benchmark cross-provider $\sigma^2$ under regular baseline ($x$) vs.\ \texttt{initial\_uniform\_capability} ablation ($y$). The ablation has $n{=}2$ seeds, so each $y$ value is the variance of two runs. Points at $y{=}0$ would mean full null dominance; points on $y{=}x$ would mean sim contributes everything. Most benchmarks cluster low along $y$, with Adversarial Robustness as an outlier. Right:\ aggregate cross-provider $\sigma^2$ comparison (regular vs.\ uniform-cap), showing the $\sim$$13\times$ collapse in variance under ablation (equivalently, $\sim$$3.6\times$ collapse in $\sigma$). Aggregate ratio $\sigma^2_{\mathrm{uniform}}/\sigma^2_{\mathrm{regular}} = 0.075$.}
  \label{fig:null-model-uniform-cap-test}
\end{figure}

The aggregate cross-provider $\sigma^2$ collapses from $0.000491$ (regular) to $0.000037$ (uniform-cap), a ratio of $0.075$. The simulation-attributable share of cross-provider gap variance is therefore $\sim$$7.5\%$; the null-attributable (initial-cap heterogeneity) share is $\sim$$92.5\%$. Per-benchmark ratios (Figure~\ref{fig:null-model-uniform-cap-test}, left panel) cluster in $0.01$--$0.13$ for most benchmarks. Two notable exceptions surface as boundary cases:\ \emph{Adversarial Robustness} (ratio $0.451$, $\sim$$45\%$ sim contribution) and \emph{Advanced Math} (ratio $0.185$). The Adversarial Robustness boundary is consistent with strategic safety-investment divergence across providers being a source of provider-level differentiation that the fitted null cannot capture. At $n{=}2$ seeds the per-benchmark ratios are indicative, and the aggregate ratio is the firmer quantity.

\paragraph{Path B: held-out null replication under capability ablation.}
Path A's variance decomposition is a within-data summary statistic. As a tighter test, we ask whether the null model trained on the $62$ heterogeneous-cap LLM runs of the primary dataset above still predicts a held-out test set drawn entirely from a regime where its training-set heterogeneity is removed. The held-out test set is $6$ LLM Sonnet runs at \texttt{public\_only} and \texttt{private\_only} crossed with seeds $\{42, 43, 44\}$ under the \texttt{initial\_uniform\_capability} ablation. In those runs the cross-provider standard deviation of round-$0$ capability is $0.000000$ in every dimension, and pairing on (seed, benchmark) gives $n{=}39$ prediction cells.

\begin{figure}[!htb]
  \centering
  \includegraphics[width=0.96\textwidth]{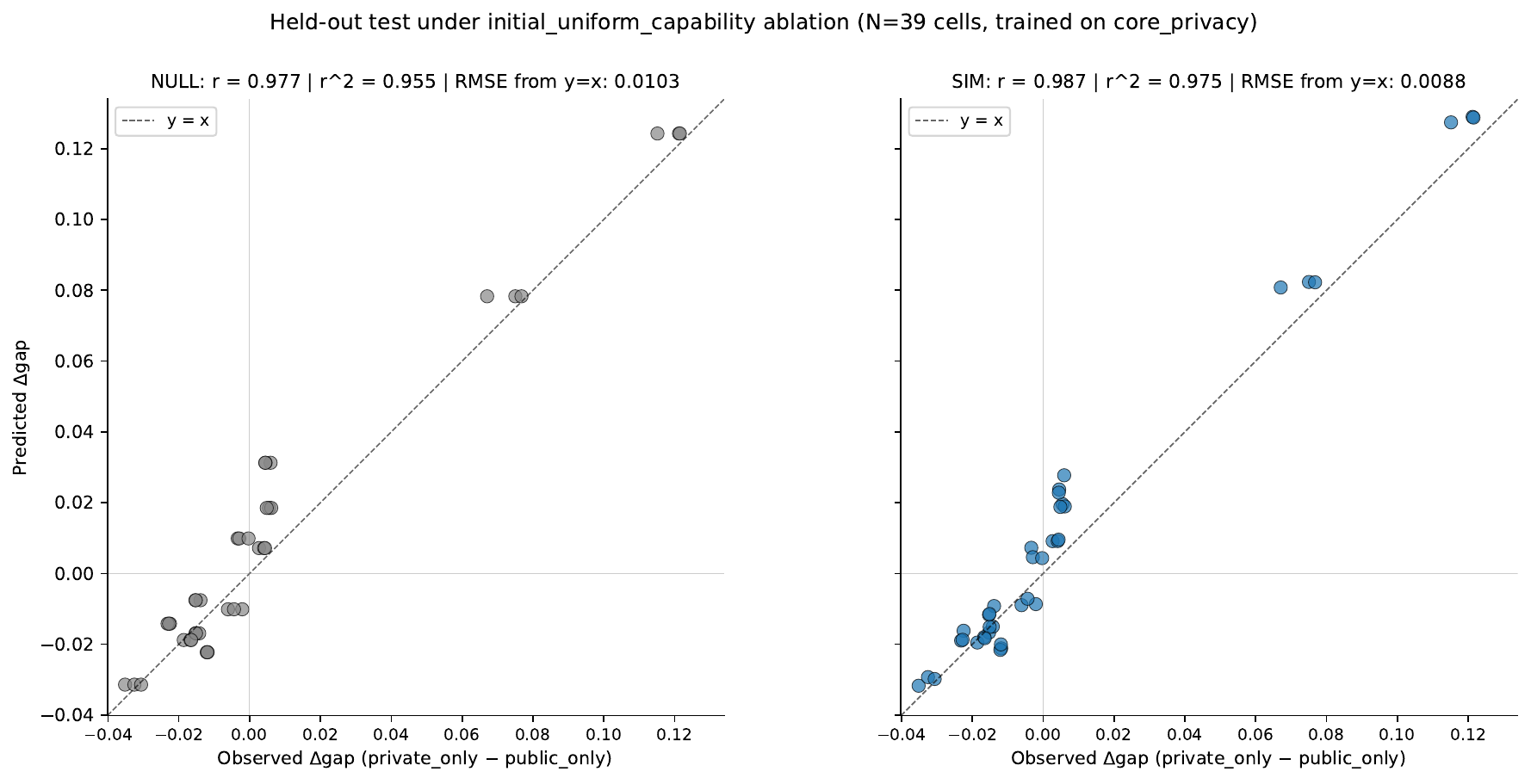}
  \caption{Path~B held-out null replication ($n{=}39$ test cells under \texttt{initial\_uniform\_capability}). Null model is trained on $62$ heterogeneous-cap runs and predicts the held-out uniform-cap test set. Both null and sim track the $y{=}x$ diagonal closely; null $r^2 = 0.955$, sim $r^2 = 0.975$.}
  \label{fig:null-model-pathb}
\end{figure}

Trained on the heterogeneous data, the null still predicts the held-out uniform-cap test set at $r^2{=}0.955$, which is about $2$ pp below the heterogeneous-cap reference of $0.974$ (Table~\ref{tab:null-model-pathb}, Figure~\ref{fig:null-model-pathb}). The simulation's marginal contribution grows from $+0.2$ pp on the heterogeneous set to $+2.0$ pp on the uniform-cap set, a tenfold relative increase that remains small in absolute terms.

\begin{table}[!htb]
\centering
\caption{Path~B held-out null replication. The null model is fit on the $62$ heterogeneous-cap LLM runs of the primary dataset, then evaluated on a held-out test set of $6$ new LLM runs under the \texttt{initial\_uniform\_capability} ablation. The null's $r^2$ drops only $\sim$$2$ pp despite test-set initial-capability heterogeneity being structurally zero.}
\label{tab:null-model-pathb}
\small
\setlength{\tabcolsep}{6pt}
\begin{tabular}{lcccc}
\toprule
\textbf{Test condition} & $n$ & $r^2_{\mathrm{null}}$ & $r^2_{\mathrm{sim}}$ & $\Delta r^2$ (sim $-$ null) \\
\midrule
Reference: heterogeneous caps (S0+s5+s8) & $208$ & $0.974$ & $0.976$ & $+0.002$ \\
Held-out: uniform caps (\texttt{initial\_uniform\_capability}) & $39$ & $\mathbf{0.955}$ & $0.975$ & $+0.020$ \\
\bottomrule
\end{tabular}
\end{table}

\paragraph{Triangulation.}
The three tests measure complementary quantities and point the same way: the pooled cross-sequence LLM comparison ($\Delta r^2 = +0.002$), the Path~A variance decomposition ($\sigma^2$ ratio $0.075$) and the Path~B held-out replication ($r^2_{\mathrm{null}} = 0.955$). The predictive content of the privacy mechanism rests on two pieces, the closed-form identity $\Delta g = \mathbf{c} \cdot (\mathbf{w}_{\mathrm{holdout}} - \mathbf{w}_{\mathrm{public}})$ and a growth model, fit on simulation runs, that predicts end-of-run capability $\mathbf{c}$ from round-$0$ capability and portfolio. The simulation contributes a small residual on top, and its size depends on where it is measured: it is negligible on the pooled cross-sequence set, about $2$ pp on the uniform-cap set, and concentrated on benchmarks such as Adversarial Robustness, where strategic safety-investment divergence is a primary differentiator. Most of the pooled variance lies between benchmarks, where the null and the simulation use the same fixed weight differences, so the pooled fit does not test what the trajectory adds. The identity is fixed by how scores and satisfaction are defined, and it needs realized capability, because evaluated directly on round-$0$ capability it mis-signs three benchmarks (Robustness check~3, Appendix~\ref{app:rescoring}). The predictability of end-of-run capability from round-$0$ conditions is an empirical property of these runs, and Path~B checks it on held-out runs. The cross-mode sign difference on market concentration under evaluator capture (Appendix~\ref{app:evaluator-capture-ext}, LLM $N{=}2$ against heuristic $p\approx 0.20$) may mark an outcome that needs the full system, and it is too small to establish.

\paragraph{Caveats.}
The test-set sample sizes for the structural-ablation robustness checks are small. Path~A's per-benchmark variance ratios rest on $n{=}2$ uniform-cap seeds, which gives wide confidence intervals on individual benchmarks and leaves the aggregate ratio of $0.075$ as the firmer claim, and Path~B's $n{=}39$ paired cells sit at the low end of what supports an $r^2$ estimate. Bootstrap confidence intervals on both quantities would tighten the claim, and the directional conclusion does not depend on them. The null model uses two features per (provider, dimension). A richer null with portfolio interactions or round counts could raise $r^2_{\mathrm{null}}$ further, which would narrow the simulation's contribution, so the null we use is the conservative choice for the claim we make.

\subsection{Cross-model setup and scope}
\label{app:model-robust-setup}

The Tier~1 privacy ladder consists of five conditions (\texttt{public\_only}, \texttt{baseline}, \texttt{private\_\allowbreak dominant}, \texttt{private\_\allowbreak only}, \texttt{iid\_\allowbreak holdout}) executed at multiple seeds. Sonnet runs cover the full ladder at seeds 42--51 ($n{=}10$ per condition); Opus and GPT-5.5 each cover the full ladder at seeds 42, 43, and 44 ($n{=}3$ per condition).\footnote{Opus and GPT-5.5 coverage is narrower than Sonnet coverage because each is roughly $1.7$ to $2\times$ more expensive per token than Sonnet at published API pricing, so we used Sonnet for breadth and Opus and GPT-5.5 for the paired-seed cross-model check. Appendix~\ref{app:model-robust-runtime} gives the wall-clock and pricing detail.} Table~\ref{tab:model-robust-endpoints} lists the runs that entered this appendix as matched triples, which are the same (condition, seed) cell run under all three models, so that per-benchmark gap, market structure and reasoning traces can be compared directly across vendors.

\textbf{Coverage caveat.} The matched-triple structure has $n{=}3$ per condition for the cross-model comparison (seeds 42, 43, and 44 paired across all three models in every privacy condition), and $n{=}10$ on Sonnet alone (seeds 42--51). The error bars in Figure~\ref{fig:model-robust-ladder} therefore rest on $n{=}10$ in the Sonnet panel and on $n{=}3$ in the Opus and GPT-5.5 panels. The comparison works at the outcome level, which is enough to test direction and ordering and to detect clear cross-model differences in magnitude, while small differences in a single (condition, seed) cell remain noisy. Where we report a quantitative cross-model difference at a single cell, we report it as directional and we do not test it statistically. Section~\ref{app:model-robust-frames}, by contrast, aggregates over all fifteen matched triples and pools many planning-stage entries per (model, provider) cell, so frame-share comparisons are pooled estimates rather than single-condition reads.

We draw on three windows into each run: end-of-run endpoint metrics (share-weighted final-5-round means), per-benchmark clean-gap trajectories, and the full reasoning transcripts of the six providers. The endpoint and per-benchmark analyses use the same construction as the heuristic privacy-ladder extended results above; the reasoning-frame analysis is presented in Appendix~\ref{app:model-robust-frames}.

\subsection{Privacy-ladder robustness}
\label{app:model-robust-ladder}
\label{app:model-robust}

\begin{figure}[!htb]
  \centering
  \includegraphics[width=0.98\textwidth]{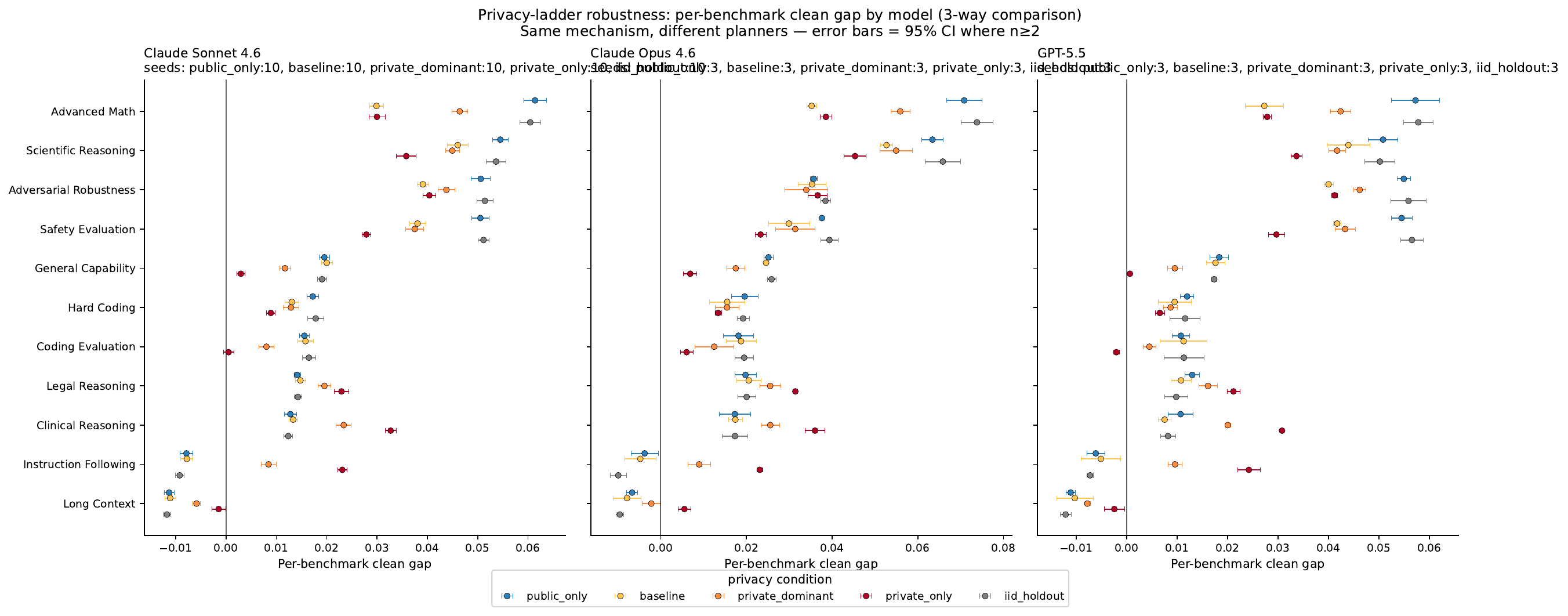}
  \caption{Per-benchmark clean gap by privacy condition, one panel per model. Benchmarks sorted by Sonnet \texttt{public\_only} mean descending so the panels share a reading axis. Error bars are $95\%$ CIs across seeds ($n{=}10$ Sonnet seeds 42--51, $n{=}3$ Opus and GPT-5.5 seeds 42--44). The privacy-condition ordering (public at the largest gap, private\_only most compressed on high-gap benchmarks) replicates under all three planners.}
  \label{fig:model-robust-ladder}
\end{figure}

Figure~\ref{fig:model-robust-ladder} plots the per-benchmark clean gap under each privacy condition, one panel per model, with benchmarks ordered by Sonnet \texttt{public\_only} descending.

\textbf{The privacy-condition ordering replicates across all three planners.} On the high-gap benchmarks (Advanced Math, Scientific Reasoning, Safety Evaluation, Adversarial Robustness), all three models show the same ladder, with \texttt{public\_only} (blue) at the largest gap, \texttt{private\_only} (dark red) compressed nearest to zero, and \texttt{private\_dominant} (orange) and \texttt{baseline} (yellow) between them. The main-body claim that privacy conditions compress the per-benchmark gap toward zero therefore holds under all three planners and across two vendors. On the underpromise-side benchmarks (Instruction Following, Long Context), all three planners produce the same negative \texttt{public\_only} signature that \S\ref{sec:results} documents.

\begin{figure}[!htb]
  \centering
  \includegraphics[width=0.98\textwidth]{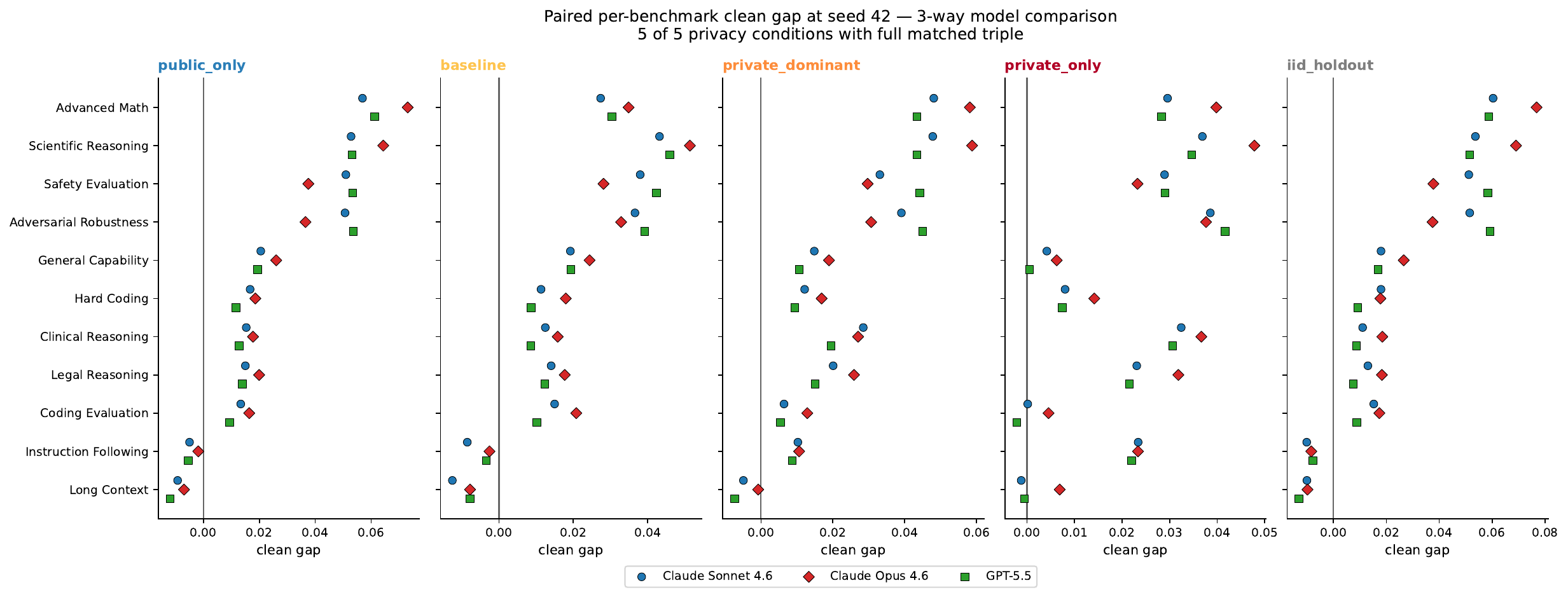}
  \caption{Paired per-benchmark clean gap at seed~42. One panel per privacy condition with a matched 3-way Sonnet+Opus+GPT-5.5 triple; circle = Sonnet, diamond = Opus, square = GPT-5.5. Dots are provider-mean gap per benchmark. All five privacy conditions have a complete 3-way matched triple at seed~42.}
  \label{fig:model-robust-forest}
\end{figure}

\textbf{GPT-5.5 sits closer to Sonnet than Opus does.} The paired per-benchmark view at seed~$42$ (Figure~\ref{fig:model-robust-forest}) shows that on most matched conditions the GPT-5.5 dot (green square) and the Sonnet dot (blue circle) sit within a few units of each other, while the Opus dot (red diamond) sits apart, often closer to zero on the high-gap benchmarks and further from zero on the underpromise-side benchmarks. On most rows the within-vendor gap between Sonnet and Opus is as large as the cross-vendor gap between Anthropic and OpenAI. Vendor identity is therefore not the dominant axis of variation in our data.

\textbf{Market structure differs across models, and it does not vary monotonically.} The matched-triple endpoint table (Table~\ref{tab:model-robust-endpoints}) shows three differences. Opus runs produce lower mean safety allocations across the ecosystem ($\approx$$0.20$ to $0.24$, against $0.26$ to $0.30$ for Sonnet and $0.26$ to $0.32$ for GPT-5.5). HHI varies without a consistent ordering across models within a single condition; under \texttt{baseline} at seed~$43$ it is $0.528$ for Sonnet, $0.480$ for Opus and $0.374$ for GPT-5.5. The five-round-mean gap can change sign across models on the same (condition, seed) cell at the third decimal. None of these differences fall within the scope of the privacy-ladder claim, and at $n{=}3$ per condition they support direction only.

\begin{table}[!htb]
\centering
\caption{Endpoint metrics for the $15$ matched 3-way triples: all five privacy conditions at seeds $42$, $43$, and $44$. All values are share-weighted means over the final $5$ rounds except \textit{HHI} (market Herfindahl). \textit{gap} = score $-$ consumer satisfaction. The prose above gives the per-model range of mean safety allocation, which we drop from the table to keep it within the page width. \textbf{S} = Sonnet 4.6, \textbf{O} = Opus 4.6, \textbf{G} = GPT-5.5.}
\label{tab:model-robust-endpoints}
\small
\setlength{\tabcolsep}{4pt}
\begin{tabular}{@{}llccc@{\hspace{6pt}}ccc@{\hspace{6pt}}ccc@{}}
\toprule
& & \multicolumn{3}{c}{\textbf{gap}} & \multicolumn{3}{c}{\textbf{score}} & \multicolumn{3}{c}{\textbf{HHI}} \\
\cmidrule(lr){3-5}\cmidrule(lr){6-8}\cmidrule(lr){9-11}
\textbf{Condition} & \textbf{Seed} & S & O & G & S & O & G & S & O & G \\
\midrule
\texttt{public\_only}      & 42 & $+0.000$ & $+0.001$ & $+0.021$ & $0.705$ & $0.706$ & $0.695$ & $0.514$ & $0.500$ & $0.486$ \\
\texttt{public\_only}      & 43 & $+0.019$ & $+0.071$ & $+0.044$ & $0.685$ & $0.730$ & $0.661$ & $0.321$ & $0.469$ & $0.342$ \\
\texttt{public\_only}      & 44 & $+0.008$ & $+0.043$ & $+0.009$ & $0.689$ & $0.711$ & $0.694$ & $0.540$ & $0.403$ & $0.588$ \\
\midrule
\texttt{baseline}          & 42 & $-0.009$ & $-0.008$ & $-0.006$ & $0.657$ & $0.701$ & $0.667$ & $0.410$ & $0.468$ & $0.472$ \\
\texttt{baseline}          & 43 & $+0.073$ & $+0.043$ & $+0.060$ & $0.708$ & $0.730$ & $0.671$ & $0.528$ & $0.480$ & $0.374$ \\
\texttt{baseline}          & 44 & $+0.009$ & $+0.022$ & $-0.005$ & $0.659$ & $0.712$ & $0.667$ & $0.294$ & $0.440$ & $0.541$ \\
\midrule
\texttt{private\_dominant} & 42 & $+0.030$ & $+0.039$ & $+0.008$ & $0.726$ & $0.718$ & $0.684$ & $0.581$ & $0.380$ & $0.365$ \\
\texttt{private\_dominant} & 43 & $+0.018$ & $+0.015$ & $+0.017$ & $0.693$ & $0.698$ & $0.661$ & $0.390$ & $0.361$ & $0.320$ \\
\texttt{private\_dominant} & 44 & $+0.011$ & $+0.035$ & $+0.003$ & $0.664$ & $0.705$ & $0.693$ & $0.268$ & $0.322$ & $0.552$ \\
\midrule
\texttt{private\_only}     & 42 & $+0.005$ & $+0.013$ & $+0.029$ & $0.690$ & $0.733$ & $0.692$ & $0.332$ & $0.469$ & $0.471$ \\
\texttt{private\_only}     & 43 & $+0.022$ & $+0.041$ & $+0.049$ & $0.698$ & $0.764$ & $0.675$ & $0.399$ & $0.574$ & $0.377$ \\
\texttt{private\_only}     & 44 & $+0.013$ & $+0.050$ & $+0.014$ & $0.668$ & $0.726$ & $0.663$ & $0.272$ & $0.408$ & $0.299$ \\
\midrule
\texttt{iid\_holdout}      & 42 & $+0.020$ & $+0.027$ & $+0.000$ & $0.694$ & $0.702$ & $0.663$ & $0.503$ & $0.363$ & $0.330$ \\
\texttt{iid\_holdout}      & 43 & $+0.020$ & $+0.082$ & $+0.040$ & $0.664$ & $0.730$ & $0.656$ & $0.299$ & $0.440$ & $0.344$ \\
\texttt{iid\_holdout}      & 44 & $+0.002$ & $+0.016$ & $+0.022$ & $0.655$ & $0.691$ & $0.669$ & $0.262$ & $0.287$ & $0.405$ \\
\bottomrule
\end{tabular}
\end{table}

\subsection{Path dependence across models}
\label{app:model-robust-pathdep}

\begin{figure}[!htb]
  \centering
  \includegraphics[width=0.98\textwidth]{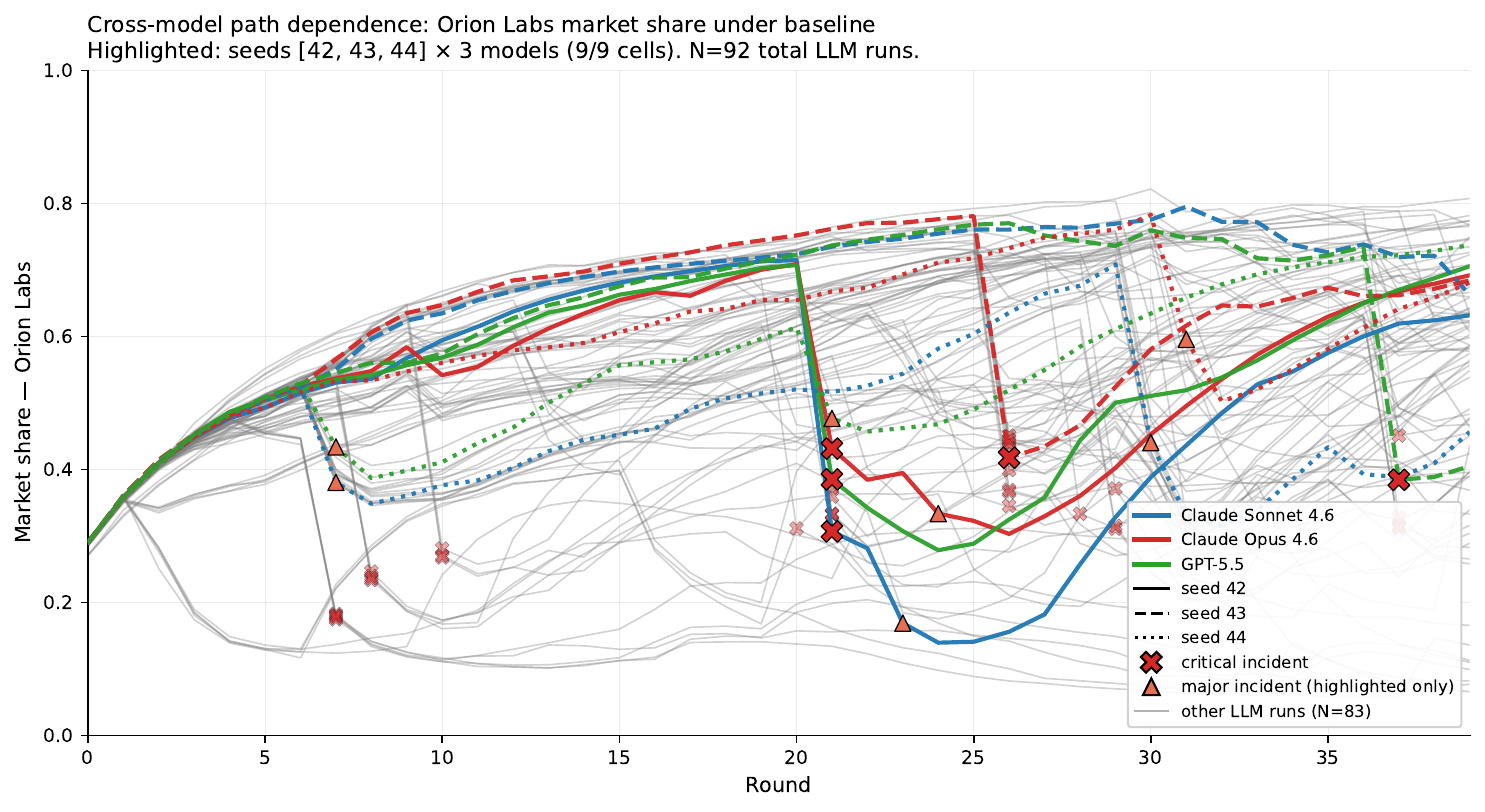}
  \caption{Cross-model market-share path dependence under \texttt{baseline} for Orion Labs. Highlighted: seeds $\{42, 43, 44\}$ $\times$ three models (color = model, linestyle = seed); all nine cells are populated. Background grey: every other full-40-round LLM run in the core privacy set, across all conditions, seeds and models. Critical incidents marked on every line; major incidents on highlighted lines only. Figure~\ref{fig:path-dependence} shows the same provider under one model across five privacy conditions; this figure holds the condition fixed and varies the model.}
  \label{fig:model-robust-pathdep}
\end{figure}

Figure~\ref{fig:model-robust-pathdep} overlays single-provider market-share trajectories across (seed, model) cells under \texttt{baseline}, with every other LLM run in the core privacy set drawn in grey. The highlighted lines are seeds $\{42, 43, 44\}$ crossed with the three models, with colour identifying the model and line style identifying the seed, and all nine cells are populated. The figure tracks Orion Labs, the same provider as the incident path-dependence figure (Figure~\ref{fig:path-dependence}, Appendix~\ref{app:path-dependence}), so a reader can compare dispersion across seeds and privacy conditions under one model against dispersion across models and seeds under one condition.

Dispersion within a model across seeds (Sonnet at $\approx$$0.63$, $0.66$ and $0.46$ across seeds $42$, $43$ and $44$; GPT-5.5 at $\approx$$0.71$, $0.41$ and $0.74$) and dispersion across models at a fixed seed (at seed~$44$, Sonnet $\approx$$0.46$, Opus $\approx$$0.68$, GPT-5.5 $\approx$$0.74$) are of comparable magnitude, and both span ranges of $\approx$$25$ to $30$\,pp in final-round market share. The envelope traced by the full $N{=}10$ Sonnet ladder and by the three-way matched seeds is consistent with Appendix~\ref{app:path-dependence}, where the timing of the first critical incident separates the runs. Opus clusters tightly across seeds ($\approx$$0.68$ to $0.69$ on all three) while Sonnet and GPT-5.5 spread widely. Three seeds do not tell us whether that reflects an Opus-specific response to incidents or a small-sample artifact.

\paragraph{What the cross-model robustness check does \emph{not} claim.}
We do not claim that any of the three models is more or less ``correct.'' At $n{=}3$ matched triples per condition we read direction and ordering as supported, and we make no claim about whether small endpoint-level differences in gap or HHI between models at a single (condition, seed) cell are stable, because three seeds per condition is narrow coverage. We do claim that two further frontier LLMs replicate the privacy-mechanism finding of \S\ref{sec:results} at the behavioral level, because the condition ordering and the sign of gap compression on high-gap benchmarks hold across Sonnet, Opus and GPT-5.5. At the reasoning level the claim is partial, because both Anthropic planners invoke the $K$-lag mechanism explicitly in matched planning rounds while GPT-5.5 reaches the same operational conclusion with much less explicit reasoning over the mechanism (Appendix~\ref{app:model-robust-frames}). Two systematic differences remain, a lower mean safety allocation under Opus and a stepwise fall in the incident-framing rate from Sonnet to Opus to GPT-5.5. Both are consistent with reasoning styles that differ by vendor and by model family, and neither shows the privacy mechanism failing to generalize.

\subsection{Methodological caveats}
\label{app:llm-caveats}

\paragraph{LLM seed coverage.}
With ten paired seeds per condition under the primary Sonnet ladder, and three paired seeds each under Opus and GPT-5.5 for cross-vendor coverage (Appendix~\ref{app:model-robust}), we can compare matched runs and report directional agreement, while variance estimates on continuous outcomes (gap magnitude, HHI, per-provider trajectories) stay wider than the heuristic layer's. Wherever the main text uses an LLM-derived number, its denominator is a within-run event count, such as an incident-response rate, or a matched-seed paired difference. We do not report an inter-run mean with a tight confidence interval from the LLM layer. Distributional claims rest on the heuristic $N{=}50$ layer throughout.

\paragraph{Paired-seed design.}
LLM runs within a comparison (e.g., \texttt{baseline} vs \texttt{private\_only}) use the same set of paired seeds $\{42, 43, \ldots, 51\}$ (ten under Sonnet; the first three for Opus and GPT-5.5 cross-vendor coverage). This controls the simulation's internal random streams (incident draws, noise, market dynamics) across paired conditions, reducing inter-seed variance in paired differences. It does \emph{not} control the LLM's server-side sampling stochasticity: even at temperature zero, API responses are not perfectly deterministic. We rely on the $N{=}50$ heuristic layer for the residual server-side-noise contribution to variance estimates.

\paragraph{Fallback handling.}
An LLM call whose response fails JSON validation is retried with the same prompt, up to three attempts in total. When all three fail, the actor takes a fail-safe action for that round: a provider keeps the portfolio it chose in the previous round, a funder splits its allocation evenly, and the regulator falls back to its heuristic plan (Appendix~\ref{app:validation-verification}). The simulation prints a line for every call that reaches the fail-safe, and we read those lines from the run console, because the saved run files do not record a fallback count. In the LLM privacy-ladder runs, fewer than $2\%$ of calls per run reached the fail-safe. We re-queued any run with a higher rate instead of including it in the paper layer.

\section{Limitations}
\label{app:limitations-ext}

This appendix gives the full treatment of the limitations that the main text summarizes. Appendix~\ref{app:lim-modeling-choices} expands the four numbered modeling-choice limits of \S\ref{sec:limitations}, Appendix~\ref{app:lim-per-actor} lists the per-actor action-space choices we did not implement, and Appendix~\ref{app:lim-methodology} sets out the methodological bounds of GABMs.

\subsection{Depth on the four main-body modeling-choice limits}
\label{app:lim-modeling-choices}

\S\ref{sec:limitations} states four foundational modeling choices on which the privacy-ladder mechanism rests; this subsection expands each with material that did not fit in the main body.

\paragraph{1.\  Privacy operates through weight perturbation in this simulator.}
The simulation's privacy mechanism decomposes into three channels (Appendix~\ref{app:private-benchmark-reporting}): a cosine-distance perturbation $\cos\theta$ between public and holdout dimension weights, a $K{=}3$ reporting lag, and observation noise scaling with $\sqrt{\text{samples}\times h}$. The cosine channel is a \emph{composite proxy} aggregating at least three mechanistically distinct phenomena: (i) item-level contamination and memorization of public splits during training~\citep{haimes2024benchmark, xu2024contamination}; (ii) ``training on the test task'' effects~\citep{dominguezolmedo2024training}; and (iii) adversarial holdout construction, where holdout items are sampled to test a skill mix that differs from the category label. A result attributed to weight asymmetry may be driven mainly by any one of these, or by their interaction, and the simulator does not resolve them separately. Two further holdout-design phenomena lie outside the composite cosine abstraction. The first is item-level randomization within the same dimension distribution, which draws fresh items from the same task family without changing the dimension-weight mix, so it acts as observation noise and not as weight redistribution. The second is distribution shift away from the training distribution, where holdout items are out of distribution while the nominal dimension-weight target is unchanged. Within the cosine channel, the \texttt{iid\_holdout} null ($\cos\theta = 1$) isolates cosine effect from lag and noise, but does not split the \emph{scoring-layer} effect (same trained capability, scored under different weights) from the \emph{R\&D-direction} effect (cosine perturbation drifts inferred weights, providers re-target). A within-benchmark rescoring (fixed end-of-run capability vectors scored under both the public and the holdout weights of the same benchmark) identifies the scoring-layer share in heuristic mode at $0.93$--$0.96$ of $\Delta g$ (Appendix~\ref{app:rescoring}); LLM runs are outside that decomposition (\S\ref{sec:limitations}, item~1).

\paragraph{2.\ The six-dimension ontology.}
The dimensions (reasoning, coding, knowledge, safety, communication, agentic) are anchored to HELM- and LMSYS-style benchmark taxonomies and to where industry investment separates into distinct R\&D pathways, such as domain corpora, instruction tuning, RLHF and tool-use trajectory RL. The choice of ontology could change the result in two ways. A \emph{flatter} ontology, which collapsed reasoning and knowledge into a single general-capability axis, would erase the unevenness between the dimensions that receive investment and the dimensions that receive credit, and the holdout weight perturbation acts on that unevenness. Under a one-dimensional ontology the holdout cosine perturbation reduces to scoring noise. A \emph{finer-grained} ontology, which separated mathematical reasoning from causal reasoning, or long-context coherence from instruction following, could move specific benchmarks from one side of the population mean to the other, which would move those benchmarks between the shrinking and widening sets. The mechanism holds within any ontology that has at least one asymmetry between a high-investment and a low-investment dimension, and it does not survive an arbitrary re-coordinatization of the capability space.

\paragraph{3.\ Public-benchmark-driven competitive contexts only.}
The instantiated ecosystem is leaderboard-driven and competitive across firms, it has a single jurisdiction, and it has six providers with no entry and no exit. We do not know how the mechanism engages in three other classes of context. The first is FDA-style gatekeeping, as in medical-device classification under 21 CFR 860 or EMA conditional approval, where mandatory pre-deployment evaluation under regulator-controlled holdouts replaces the leaderboard signal and clinicians stand between the model and the consumer. The second is defense procurement, where evaluation is bilateral, classified and tied to specific use cases, so no public leaderboard exists for the holdout mechanism to act on. The third is enterprise business-to-business sales with an internal proof of concept, where buyers run their own evaluations against deployment-specific data, which partly decouples adoption from public benchmark scores. That correlation between buyer-side internal benchmarks and the public set decides whether the decoupling dampens or amplifies the effect, and we do not estimate it. Internal-deployment evaluation infrastructure, such as Responsible Scaling Policies, Preparedness frameworks and the Frontier Safety Framework~\citep{anthropic2023rsp,openai2023preparedness,deepmind2024fsf}, dates from 2023 onward, sits beside the public-benchmark channel, and adds pressure on R\&D direction that the simulation does not model.

\paragraph{4.\ Model-defined consumer satisfaction.}
Consumer satisfaction starts from $\mathbf{c}_p^\top \mathbf{w}_s$, with both $\mathbf{c}_p$ (provider capability) and $\mathbf{w}_s$ (segment need weights) held by the simulation, and incident and cost terms then adjust it (Appendix~\ref{app:consumers}). The per-benchmark gap uses the capability term alone. We assign each segment a need-weight vector over the six capability dimensions from sector-level evidence on what that profession uses AI for and what it is concerned about, together with occupation-level adoption rates in \citet{bick2024rapid} (Appendix~\ref{app:parameters}). Two issues remain open. First, the evidence fixes which needs dominate each profile, and the exact values are a judgment we make from it. A survey instrument that elicits relative weights directly from users, through forced choice, conjoint analysis or willingness to pay, would test that judgment, and no such instrument exists for AI capabilities today. Second, the shrink and widen counts depend on whose satisfaction defines the gap. Need weights cancel out of each benchmark's shift, $\Delta g = \mathbf{c}\cdot(\mathbf{w}_{\mathrm{holdout}}-\mathbf{w}_{\mathrm{public}})$, and enter only the starting gap, through matched satisfaction. The starting gap's sign decides whether a shift widens or shrinks the gap. Under uniform need weights ($\mathbf{w}_s = \mathbf{1}/d$ for every segment) matched satisfaction reduces to mean capability, every starting gap rises, and one classification of $13$ changes (Long Context; heuristic, $N{=}50$, end-of-run capability held fixed). Against safety-heavy segments alone four change: Coding Evaluation, General Capability and Hard Coding widen, and Instruction Following shrinks. For consumers whose needs equal a benchmark's loadings, the public score tracks their satisfaction, so any shift larger than scoring noise widens their gap.

\subsection{Per-actor action-space choices not implemented}
\label{app:lim-per-actor}

The simulation's actor space is a simplification. Each actor has a richer real-world strategy space than the simulation implements, and Table~\ref{tab:per-actor-action-space} lists what we left out of each one. The omissions do not affect the specific question we ask in \S\ref{sec:privacy-ladder}. For the adjacent questions that the ecosystem frame can host, each entry marks a place where the architecture would need an extension before the question could be studied.

\begin{longtable}{@{}p{1.5cm} p{4.5cm} p{6.5cm}@{}}
\caption{Per-actor action-space choices: scope modeled in the simulation vs.\ extensions not implemented. Each row lists the strategy levers exercised by the actor alongside concrete real-world levers the architecture would need to extend to host an adjacent question.}\label{tab:per-actor-action-space} \\
\toprule
\textbf{Actor} & \textbf{Modeled} & \textbf{Not modeled} \\
\midrule
\endfirsthead
\multicolumn{3}{@{}l}{\footnotesize\textit{Table~\ref{tab:per-actor-action-space} (continued).}} \\
\toprule
\textbf{Actor} & \textbf{Modeled} & \textbf{Not modeled} \\
\midrule
\endhead
\bottomrule
\endfoot
\bottomrule
\endlastfoot

Model providers &
\begin{itemize}[leftmargin=10pt, itemsep=0pt, topsep=0pt, parsep=0pt]
  \item R\&D / safety / product portfolio (sums to 1.0)
  \item Per-benchmark focus levels and weight beliefs
  \item Public communications (announcements, press releases)
  \item Optional best-of-$N$ submissions (evaluator-capture)
  \item Six-dim capability vector with sqrt-budget capability gains
\end{itemize} &
\begin{itemize}[leftmargin=10pt, itemsep=0pt, topsep=0pt, parsep=0pt]
  \item Distillation or model merging across providers
  \item Internal (non-published) benchmark suites that gate release decisions
  \item Training-data vendor relationships and selective contamination of public benchmarks
  \item Multi-model portfolios (one provider shipping several model sizes simultaneously)
  \item Internal red-team findings that never become observable incidents
\end{itemize}
\\
\midrule

Consumers &
\begin{itemize}[leftmargin=10pt, itemsep=0pt, topsep=0pt, parsep=0pt]
  \item 51 segments (archetype $\times$ use-case)
  \item Need weights assigned from sector-level evidence on AI use and concerns, with occupation-level adoption data
  \item Switching cost + post-switch cooldown; dynamic enterprise-share growth
  \item Incident-weighted satisfaction with sector-matched 2$\times$ penalty
  \item Media-attention-driven exploration churn
\end{itemize} &
\begin{itemize}[leftmargin=10pt, itemsep=0pt, topsep=0pt, parsep=0pt]
  \item Multi-provider routing (one consumer using several models in a portfolio)
  \item Internal use-case-specific evaluation by organizational consumers (private benchmarks they trust above the public leaderboard)
  \item Regulatory-mandated provider choice for specific deployment contexts (healthcare-grade certification, government procurement)
\end{itemize}
\\
\midrule

Evaluator &
\begin{itemize}[leftmargin=10pt, itemsep=0pt, topsep=0pt, parsep=0pt]
  \item 22-benchmark pool with three modes: fixed sequence, randomized pool, dynamic LLM-driven
  \item Three benchmark types (public, partial with $h{=}0.3$, private with $h{=}1.0$) and a $K{=}3$ reporting lag
  \item Saturation detection from score deltas; hand-authored holdout weights per benchmark
  \item Optional best-of-$N$ + early-access toggles (evaluator-capture)
\end{itemize} &
\begin{itemize}[leftmargin=10pt, itemsep=0pt, topsep=0pt, parsep=0pt]
  \item Coalitions or standards bodies (HELM-like consortia, MLCommons)
  \item Multiple competing evaluators with different weightings on the same measurement axis
  \item Benchmark retraction or integrity audits after release
  \item Data-vendor / data-labeling supply chains feeding the benchmark
  \item Conflict-of-interest disclosure requirements and evaluator reputation dynamics (track record affecting which scores are trusted)
\end{itemize}
\\
\midrule

Regulator &
\begin{itemize}[leftmargin=10pt, itemsep=0pt, topsep=0pt, parsep=0pt]
  \item Single regulator with a graduated lever ladder (voluntary commitment $\to$ advisory $\to$ disclosure $\to$ audit $\to$ sanction)
  \item Three parameter sets (US light-touch, EU precautionary, balanced); every reported run uses balanced
  \item Per-lever cooldowns in heuristic mode, a three-round gap on the LLM path; incident-rate and concentration triggers
  \item Mandatory safety floor under audit, sanction or emergency investigation
\end{itemize} &
\begin{itemize}[leftmargin=10pt, itemsep=0pt, topsep=0pt, parsep=0pt]
  \item Industry-specific gatekeeping regulators (FDA for healthcare AI, FAA for aviation AI)
  \item Multiple jurisdictions with mobility or arbitrage between them
  \item Whistleblower channels surfacing hidden conduct
  \item Safe-harbor provisions and self-certification pathways
\end{itemize}
\\
\midrule

Funders &
\begin{itemize}[leftmargin=10pt, itemsep=0pt, topsep=0pt, parsep=0pt]
  \item Four types (VC, corporate, government, foundation) with type-specific cooldowns, informed by the cadence of AI funding events between 2020 and 2026
  \item \$193B capital pool with a geometric 7\% per month availability curve
  \item Per-funder identity blocks and peer-funder visibility (LLM mode)
  \item Allocation against leaderboard, market shares, regulator interventions, recent funding history
\end{itemize} &
\begin{itemize}[leftmargin=10pt, itemsep=0pt, topsep=0pt, parsep=0pt]
  \item Compute-credit financing (cloud-provider equity-for-credits arrangements, e.g., Microsoft--OpenAI, AWS--Anthropic)
  \item Strategic vs.\ purely financial corporate-investor distinction
  \item Exit events (acquisition, IPO) and their behavioral feedback into remaining providers
\end{itemize}
\\
\midrule

Media &
\begin{itemize}[leftmargin=10pt, itemsep=0pt, topsep=0pt, parsep=0pt]
  \item Single MediaActor (TechPress) with multi-round narrative state inertia (OPTIMISM / SKEPTICISM / CRISIS)
  \item Per-provider attention vector and sentiment; headline budget per round
  \item Weighted coverage of leaderboard movement, incidents, regulatory actions, provider announcements
  \item Rule-based (not LLM-driven)
\end{itemize} &
\begin{itemize}[leftmargin=10pt, itemsep=0pt, topsep=0pt, parsep=0pt]
  \item Differentiated media types (trade press vs.\ mainstream vs.\ research press)
  \item Social-media amplification (TikTok, Twitter/X, Reddit) and the influencer economy
  \item Engagement-driven coverage (algorithmic feedback between virality and what gets covered next)
\end{itemize}
\\
\midrule

Incident generation &
\begin{itemize}[leftmargin=10pt, itemsep=0pt, topsep=0pt, parsep=0pt]
  \item Probabilistic generation (base rate 20\%/round, modulated by safety allocation)
  \item Four-level severity (minor / moderate / major / critical) with exponential decay, so an incident's weight halves in about two rounds
  \item Six categories (healthcare, security, bias, safety, misinformation, misuse) with a sector-matched 2$\times$ consumer penalty
  \item Per-provider attribution; incident probability scales with market share and market size, and the penalty an incident imposes on satisfaction scales as $(1 + \text{market\_share}^2)$
\end{itemize} &
\begin{itemize}[leftmargin=10pt, itemsep=0pt, topsep=0pt, parsep=0pt]
  \item Specific failure-mode taxonomy separable from severity (prompt injection vs.\ hallucination vs.\ bias)
  \item Whistleblower disclosures and academic expos\'{e}s generating reputational incidents without direct user harm
  \item Regulatory discovery of hidden failures during audit
  \item Socioeconomic and environmental harms (labor displacement, compute-footprint externalities) and human-computer interaction harms (over-reliance, automation bias), discussed against the AI Risk Repository taxonomy in Appendix~\ref{app:incidents}
\end{itemize}
\\

\end{longtable}

\subsection{Methodological bounds of GABMs}
\label{app:lim-methodology}

\paragraph{The outputs are qualitative.}
The simulation outputs candidate dynamics, which are directional patterns, an ordering of effects, and the conditions under which a mechanism fires. It does not output point predictions. Specific magnitudes, such as gap sizes, incident counts and market shares, are properties of the parameterization, and we do not present them as claims about real markets. A finding such as ``privacy moves scoring credit toward or away from the dimensions where provider R\&D has concentrated capability'' is the kind of output this work targets. A finding of the form ``privacy reduces the gap by $X\%$'' is outside what it can support.

\paragraph{The validation surface is multi-dimensional.}
Validation of GABMs is fundamentally harder than validation of classical simulations because the validation surface multiplies across four partially-independent dimensions:

\begin{itemize}[left=0pt]
  \item \textbf{Parameter space.} The parameters differ in how much empirical grounding each one has, and Appendix~\ref{app:design-rationale} rates every assumption as calibrated, anchored or stipulated. The core drivers are anchored to external data: the private-benchmark reporting lag $K{=}3$ to Epoch AI benchmark cadence, the target cosine of $0.95$ to retro-holdout score inflation~\citep{haimes2024benchmark}, the target cosine of $0.85$ to within-family correlations we compute from Epoch AI data~\citep{epoch2024benchmarks}, the 2023 capability vectors to~\citet{stanfordaiindex2024}, the incident base rate to the AI Incident Database, and the consumer need-weight vector to sector-level evidence on AI use and to occupation-level adoption data~\citep{bick2024rapid}. Peripheral parameters, such as decay rates, threshold magnitudes and scale factors, are modeling choices, some of them chosen for numerical stability. A sensitivity analysis over the full joint parameter space is computationally intractable, so we rely on cross-condition robustness and paired-seed matching in place of exhaustive sweeps.
  \item \textbf{Prompt space.} Every LLM actor decision rests on a prompt that is one of many possible framings of the same information. The PIMMUR audit (Appendix~\ref{app:pimmur}) rules out the demand characteristics and coaching patterns that it covers, and it does not establish that the observed dynamics are invariant to the prompt. We do not know whether other demand characteristics remain. The neutral framing also bounds the holdout result. Provider prompts state how each benchmark type is scored and never present holdouts as something to exploit, so the result describes providers that were not prompted to target holdout weights. A provider instructed to infer and chase holdout weights could change which benchmarks widen.
  \item \textbf{Architecture space.} Our choices about which mechanisms to implement and which to abstract away decide specific findings. The capability vector has six dimensions, benchmark orientation is a scalar that blends benchmark-driven and consumer-driven R\&D, and safety has its own portfolio lever separate from R\&D. The weight-asymmetry mechanism of \S\ref{sec:privacy-ladder} rests on that asymmetric architecture, and a simulation built on a different capability representation could change which benchmarks widen.
  \item \textbf{Model space.} LLM-driven agents may pattern-match training data instead of deliberating over a new situation. The literature documents convergence toward an average persona and underrepresented variance in LLM-agent simulations~\citep{taillandier2025integrating,wu2025boundary}. The behavioral layer in this paper runs Claude Sonnet~4.6 as its primary planner and adds paired robustness checks at matched seeds under Claude Opus~4.6 within the same vendor and under OpenAI GPT-5.5 across vendors (Appendix~\ref{app:model-robust}). The privacy-ladder ordering replicates across all three at the outcome level, while GPT-5.5 invokes the $K$-lag mechanism in its reasoning at roughly half the Anthropic rate. The cross-vendor check weakens the concern that a convergent strategy inside the Claude family reflects model-specific bias, without removing it. Validation across other LLM families, such as Gemini, Llama and Qwen, is the most tractable extension of this work.
\end{itemize}

\paragraph{External validity rests on face validity and parameter grounding, and no decisive test is available.}
The gold-standard test of an ecosystem-level simulation's external validity would be out-of-sample prediction against real-world market data. That test is not available for this class of question, because the quantities that matter, which are true capability vectors, true consumer satisfaction and a provider's true strategic rationale, are unobservable in the real world, and the observable proxies, which are market share, benchmark scores and incident counts, are downstream outcomes and not generative inputs. An exogenous-event probe, which injects a capability-release shock at round $25$ (Appendix~\ref{app:face-validity-ext}), checks that an injected narrative reaches actor reasoning, and the parameter grounding of Appendix~\ref{app:assumption-status} ties a small set of parameters to real-world referents. Neither substitutes for a decisive external-validity test.

\paragraph{An agent-based model has no natural stopping rule.}
Every mechanism we add invites a question about the next one. The ecosystem frame bounds that by treating an addition as a new lever on an existing actor or a new channel between actors (\S\ref{sec:future-case-studies}, Appendix~\ref{app:case-studies-ext}), so the architecture does not have to be rebuilt each time. Where the existing architecture cannot carry a proposed extension, we record the limitation instead of retrofitting the model.

\section{Extended Case Studies}
\label{app:case-studies-ext}

The main text (\S\ref{sec:future-case-studies}) lists four policy-oriented case studies that the ecosystem frame supports. This appendix presents one worked study, on evaluator capture (Appendix~\ref{app:evaluator-capture-ext}), and sketches two further slots, on media shadow (Appendix~\ref{app:media-shadow-ext}) and on benchmark sponsorship (Appendix~\ref{app:benchmark-sponsorship-ext}). Each sketch adds a lever to an actor the simulation already has, the media actor or the evaluator, and neither needs a new actor or a new information flow. We have run neither of the two sketched studies.

\subsection{Evaluator capture: cross-mode results}
\label{app:evaluator-capture-ext}

\paragraph{What we model.} Evaluator capture refers to the structural risk that an evaluator's revenue dependence on the providers it evaluates erodes the cross-cutting informational role of the evaluation layer. Real-world anchors include company-run leaderboards such as Scale~AI's SEAL leaderboards (2024)~\citep{scale2024seal} and Meta's private testing of 27 model variants on Chatbot Arena before the Llama~4 release, documented by \citet{singh2025leaderboard}, whose simulations show that testing 20 variants raises the expected best score by about 50 points. Comparable structural-capture dynamics are documented in adjacent measurement-and-evaluation industries: credit-rating agencies under the pre-2008 issuer-pays model, and the Big-4 audit firms in financial reporting. In each case the integrity of a measurement layer that multiple actors depend on is conditional on the layer remaining outside the production flow it measures.

\paragraph{Mechanism.} When the lever activates (\texttt{evaluator\_as\_company=True}), four structural changes engage:
\begin{enumerate}
  \item \textbf{Best-of-$N$ trial submissions.} Each provider chooses how many submissions to make per round. The condition caps the count at $12$, and the LLM planner chooses a number between $1$ and $10$, so the runs reported below use a cap of $10$ (Appendix~\ref{app:heuristic-planning}). Each extra submission costs \texttt{fee\_per\_submission} (0.05 sim units) from the R\&D budget. The selection bias from taking the maximum of $N$ trials is $\mathbb{E}[\max(X_1, \ldots, X_N)] \approx \mu + \sigma\sqrt{2 \ln N}$.
  \item \textbf{Early access.} Paying providers get blended belief initialization on new benchmarks: their existing noisy prior is moved toward the benchmark's holdout scoring weights, $(1 - 0.7) \times \text{prior} + 0.7 \times \text{holdout weights}$, falling back to the public weights on public-type benchmarks, which carry no holdout vector.
  \item \textbf{Discretionary budget.} Each provider gets a parent-company allowance that it can spend on submissions (Genesis and Mirage 2.0; Orion 0.5; Apex 0.3; Spark and OpenCore 0.0).
  \item \textbf{Evaluator revenue.} The evaluator becomes a company with its own budget, and it collects both the submission fees and allocations from the funders.
\end{enumerate}

\paragraph{Calibration.} The submission cap of $12$ sits below the $27$ variants that Meta tested. The early-access blend factor of $0.7$ is a modeling choice, set high enough for the mechanism to show above seed-to-seed noise at $N{=}30$ in heuristic mode. The per-provider discretionary budgets are modeling choices that follow differences in parent-company resourcing, with Genesis and Mirage backed by a large platform company, Orion partnered with one, Apex funded by venture capital, and Spark and OpenCore standalone. The case study runs at \texttt{evaluation\_lag}$=0$ instead of the canonical \texttt{evaluation\_lag}$=3$, which exposes the extreme of the mechanism with no reporting lag and so gives the largest channel through which best-of-$N$ submissions turn into score inflation.

\paragraph{Cross-mode comparison.} We compare matched baseline and \texttt{evaluator\_capture} runs ($40$~rounds) in LLM mode at two paired seeds ($42$, $43$) against the heuristic sweep at $N{=}30$. Paired deltas (\texttt{evaluator\_capture} minus \texttt{baseline}) are summarized in Table~\ref{tab:evaluator-capture-deltas}. The LLM result reverses the sign of the heuristic result on the headline structural outcome: heuristic $\Delta\text{HHI} = +0.041$ (directional, $p\approx 0.20$; concentration \emph{rises} under the business-model condition), LLM $\Delta\text{HHI} = -0.029$ at $N{=}2$ (concentration \emph{falls}). At these sample sizes the difference in sign is suggestive; neither effect is individually established.

\begin{table}[!htb]
\centering
\caption{Paired deltas (\texttt{evaluator\_capture} minus \texttt{baseline}) in LLM mode at seeds $42$ and $43$, $N{=}2$. \textit{Apex share} and \textit{Apex safety} are provider-specific; the remaining columns are ecosystem-level. $\Delta$gap uses the per-benchmark gap of \S\ref{sec:results}, which is the published score minus matched consumer satisfaction, averaged over the $13$ benchmarks and the six providers in the final round.}
\label{tab:evaluator-capture-deltas}
\small
\setlength{\tabcolsep}{5pt}
\begin{tabular}{@{}lrrrrrr@{}}
\toprule
\textbf{Seed} & \textbf{$\Delta$HHI} & \textbf{$\Delta$leader~share} & \textbf{$\Delta$mean~safety} & \textbf{$\Delta$Apex~share} & \textbf{$\Delta$Apex~safety} & \textbf{$\Delta$gap} \\
\midrule
$42$ & $-0.019$ & $-0.032$ & $-0.009$ & $+0.069$ & $-0.022$ & $+0.011$ \\
$43$ & $-0.039$ & $-0.041$ & $-0.012$ & $+0.093$ & $-0.022$ & $+0.008$ \\
\midrule
Mean & $-0.029$ & $-0.037$ & $-0.011$ & $+0.081$ & $-0.022$ & $+0.010$ \\
\bottomrule
\end{tabular}
\end{table}

The LLM runs also do not match the heuristic-layer hypothesis that \texttt{evaluator\_capture} redistributes share toward safety-investing providers. Apex does gain share (mean $\Delta = +0.081$), while its own safety allocation \emph{falls} by $0.022$ on average and ecosystem-wide mean safety drops slightly ($-0.011$). The reallocation traces instead to access to throughput, because the submission cap and the early-access blend of $0.7$ toward the true weights benefit whichever provider is best placed to fill the extra leaderboard slots, whatever its safety posture. The paired $\Delta\text{gap}$ carries the same sign on both seeds ($+0.011$ and $+0.008$, mean $+0.010$), so in these two runs the \texttt{evaluator\_capture} structure widens the distance between the published score and the satisfaction consumers receive, which is the direction the best-of-$N$ mechanism predicts. Two seeds give a direction and not a magnitude. Read together, the two LLM runs de-concentrate the market slightly, move share toward Apex through a throughput channel, and widen the score-satisfaction gap, and we record the sign reversal on concentration against the $N{=}30$ heuristic layer as a candidate mode dependence.

\subsection{Media shadow}
\label{app:media-shadow-ext}

The media's coverage choices are a governance question distinct from what the evaluator measures. The \texttt{MediaActor} already weights leaderboard changes, safety incidents, regulatory escalations and provider announcements when it selects each round's headlines, and those weights are constants in the code. Shifting the attention budget toward incidents and away from leaderboard movement, or the other way round, would test whether provider investment follows whatever the media covers, independently of what the evaluator publishes.

\subsection{Benchmark sponsorship}
\label{app:benchmark-sponsorship-ext}

A different information asymmetry sits upstream of privacy. When a provider co-funds a benchmark, it may learn that benchmark's weight structure before its competitors do, and the FrontierMath and OpenAI disclosure is the real-world instance. Studying it would need a per-benchmark sponsor tag and a pre-access channel that delivers weight information to sponsors at $t-k$ and to other providers at $t$, and the simulation implements neither today. The per-submission fee of the evaluator-capture case (Appendix~\ref{app:evaluator-capture-ext}) is open to every provider on the same terms, while sponsorship is selective, because some providers know ahead of time and others do not. Such a study would ask whether co-funding works mainly as a way to extract rent, where the sponsor pays and the sponsor benefits, or as a way to signal commitment, where the sponsor pays and other providers read the signal off its visible behavior.

\end{document}